%% file: main.tex
\documentclass[sigconf]{acmart}
\usepackage{xspace}
\usepackage{accents}
\usepackage{pgf}
\usepackage{multirow}
\usepackage{rotating, multirow}
\usepackage{makecell}
\usepackage{wrapfig}
\usepackage{bm}

\newcommand{\bs}{\bm}
\newcommand{\GCN}{\textsc{GCN}\xspace}

\newcommand{\An}{\bs{{\hat{A}}}}
\newcommand{\A}{\bs{A}}

\newcommand{\Aslice}{\bs{{\dot{A}}}}
\newcommand{\Xslice}{\bs{{\dot{X}}}}
\newcommand{\hiddenhat}{\bs{{\hat{H}}}}
\newcommand{\unperturbed}{{\bs{{\dot{H}}}}^{(2)}}

\newcommand{\X}{\bs{X}}
\newcommand{\yt}{y_t^*}
\newcommand{\hidden}{\bs H}
\newcommand{\Xadv}{\tilde{\X}}

\newcommand{\cvec}{\bs{c}}

\newcommand{\W}{\bs{W}}
\newcommand{\bvec}{\bs b}
\newcommand{\pvec}{\bs p}
\newcommand{\alphavec}{\bs{\Omega}}
\newcommand{\betavec}{\bs{\Psi}}

\newcommand{\V}{\bs \Phi}
\newcommand{\Vhat}{\hat{\bs \Phi}}
\newcommand{\upperb}{\bs S}
\newcommand{\lowerb}{\bs R}

\newcommand{\numclasses}{K}
\newcommand{\numnodes}{N}
\newcommand{\numdim}{D}
\newcommand{\numhidden}{h}
\newcommand{\numlayers}{L}
\newcommand{\iterlayer}{l}
\newcommand{\target}{t}
\newcommand{\Mlocal}{\bs \eta}
\newcommand{\Mglobal}{\rho}
\newcommand{\margin}{M}
\newcommand{\latentsize}{h^{(l)}}

\newcommand{\epsplus}{\varepsilon^+}
\newcommand{\epsminus}{\varepsilon^-}
\newcommand{\gamplus}{\gamma^+}
\newcommand{\gamminus}{\gamma^-}
\newcommand{\tauvar}{\tau}
\newcommand{\muvar}{\mu}
\newcommand{\lambdavar}{\lambda}
\newcommand{\epshat}{\hat{\varepsilon}}

\DeclareMathOperator{\softmax}{softmax}
\DeclareMathOperator{\Tr}{Tr}
\DeclareMathOperator{\relu}{ReLU}

\DeclareMathOperator{\ilargest}{i-th\_largest}

\DeclareMathOperator{\sumtopQ}{sum\_top\_Q}

\newtheorem{problemB}{Problem}
\newenvironment{problem}{\begin{problemB}\begin{itshape}}{\end{itshape}\end{problemB}}

\def\BibTeX{{\rm B\kern-.05em{\sc i\kern-.025em b}\kern-.08emT\kern-.1667em\lower.7ex\hbox{E}\kern-.125emX}}
    
\copyrightyear{2019}
\acmYear{2019}
\setcopyright{acmlicensed}
\acmConference[KDD '19]{The 25th ACM SIGKDD Conference on Knowledge Discovery and Data Mining}{August 4--8, 2019}{Anchorage, AK, USA}
\acmBooktitle{The 25th ACM SIGKDD Conference on Knowledge Discovery and Data Mining (KDD '19), August 4--8, 2019, Anchorage, AK, USA}
\acmPrice{15.00}
\acmDOI{10.1145/3292500.3330905}
\acmISBN{978-1-4503-6201-6/19/08}
\begin{document}

\settopmatter{printacmref=true}
\fancyhead{}

%
\title[Certifiable Robustness and Robust Training for Graph Convolutional Networks]{Certifiable Robustness and Robust Training for\\ Graph Convolutional Networks}

%
\author{Daniel Z\"ugner \quad Stephan G\"unnemann}
\affiliation{%
	\institution{Technical University of Munich, Germany}
}
\email{{zuegnerd,guennemann}@in.tum.de}

%
\renewcommand{\shortauthors}{Daniel Z\"ugner and Stephan G\"unnemann}

%
\begin{abstract}
Recent works show that Graph Neural Networks (GNNs) are highly non-robust with respect to adversarial attacks on both the graph structure and the node attributes, making their outcomes unreliable. We propose the first method for certifiable (non-)robustness of graph convolutional networks with respect to perturbations of the node attributes\footnote{Code available at \url{https://www.kdd.in.tum.de/robust-gcn}}. We consider the case of binary node attributes (e.g. bag-of-words) and perturbations that are $L_0$-bounded.
If a node has been certified with our method, it is guaranteed to be  robust under any possible perturbation given the attack model. Likewise, we can certify non-robustness.
Finally, we propose a robust semi-supervised training procedure that treats the labeled and unlabeled nodes jointly. As shown in our experimental evaluation, our method significantly improves the robustness of the GNN with only minimal effect on the predictive accuracy. 
\end{abstract}

\maketitle

 \input{sections/introduction.tex}
 \input{sections/preliminaries.tex}
\input{sections/model.tex}
\input{sections/experiments.tex}
\input{sections/conclusion.tex}

\bibliographystyle{ACM-Reference-Format}
\bibliography{bibliography}

\input{sections/appendix-primal.tex}
\input{sections/appendix2.tex}

\end{document}

%% file: sections/introduction.tex
\section{Introduction}

Graph data is the core for many high impact applications ranging from the analysis of social networks, over gene  interaction networks, to interlinked document collections. One of the most frequently applied tasks on graph data is \textit{node classification}: given a single large (attributed) graph and the class labels of a few nodes, the goal is to predict the  labels of the remaining nodes. Applications include the classification of proteins in interaction graphs \cite{hamilton2017inductive}, prediction of customer types in e-commerce networks \cite{DBLP:journals/pvldb/EswaranGFMK17}, or the assignment of scientific papers from a citation network into topics~\cite{kipf2016semi}. %
While there exist many classical approaches to node classification \citep{london2014collective, semisupervised}, recently \emph{graph neural networks} (GNNs), also called graph convolutional networks, have gained much attention and improved the state of the art in node classification  \cite{kipf2016semi, cnn_defferrard2016convolutional,DBLP:conf/icml/GilmerSRVD17,klicpera_predict_2019}. 

However, there is one big catch: Recently it has been shown that such approaches are vulnerable to adversarial attacks \citep{zugner2018adversarial, pmlr-v80-dai18b,zugner2019meta}:
Even only slight deliberate perturbations of the nodes' features or the graph structure can lead to completely wrong predictions.
Such negative results significantly hinder the applicability of these models. The results become unreliable and such problems open the door for attackers that can exploit these vulnerabilities.

So far, no effective mechanisms are available, which (i) prevent that small changes to the data lead to completely different predictions in a GNN, or (ii) that can verify whether a given GNN is robust w.r.t.\ specific perturbation model.
This is critical, since especially in domains where graph-based learning is used (e.g.\ the Web) adversaries are omnipresent, e.g., manipulating online reviews and product websites~\cite{DBLP:conf/sdm/HooiSBGAKMF16}.
One of the core challenges is that in a GNN a node's prediction is also affected when perturbing \textit{other} nodes in the graph -- making the space of possible perturbations large. How to make sure that small changes to the input data do not have a dramatic effect to a GNN?

In this work, we shed light on this problem by proposing the first method for \textit{provable robustness} of GNNs.
More precisely, we focus on graph convolutional networks and potential perturbations of the node attributes, where we provide:
\setlength{\leftmargini}{4mm}
\begin{itemize}
\item[1)] \textit{Certificates:} Given a trained GNN, we can give robustness certificates that state that a node is robust w.r.t.\ a certain space of perturbations. If the certificate holds, it is guaranteed that \textit{no} perturbation (in the considered space) exists which will change the node's prediction. 
Furthermore, we also provide non-robustness certificates that, when they hold, state whether a node is not robust; realized by providing an adversarial example. 
\vspace*{2mm}
\item[2)] \textit{Robust Training:} We propose a learning principle that improves the robustness of the GNN (i.e. making it less sensitive to perturbations) while still ensuring high accuracy for node classification. Specifically, we exploit the semi-supervised nature of the GNN learning task, thus, taking also the unlabeled nodes into account.
\end{itemize}

In contrast to existing works on provable robustness for classical neural networks/robust training (e.g.\ \cite{kolterpolytope,semidefinite,DBLP:conf/nips/HeinA17}), we tackle various additional challenges: Being the first work for graphs, we have to deal with perturbations of multiple instances simultaneously. For this, we introduce a novel space of perturbations where the perturbation budget is constrained locally and globally. Moreover, since the considered data domains are often discrete/binary attributes, we tackle challenging $L_0$ constraints on the perturbations. Lastly, we exploit a crucial aspect of semi-supervised learning by taking also the unlabeled nodes into account for robust training.

The key idea we will exploit in our work is to estimate the \textit{worst-case} change in the predictions obtained by the GNN under the space of perturbations. If the worst possible change is small, the GNN is robust. Since, however, this worst-case cannot be computed efficiently, we provide \textit{bounds} on this value, providing conservative estimates. More technically, we derive relaxations of the GNN and the perturbations space, enabling efficient computation.

Besides the two core technical contributions mentioned above, we further perform extensive experiments:
\begin{itemize}
	\item[3)] \textit{Experiments:} We show on various graph datasets that GNNs trained in the traditional way are not robust, i.e.\ only few of the nodes can be certified to be robust, respectively many are certifiably non-robust even with small perturbation budgets. In contrast, using our robust training we can dramatically improve robustness increasing it by in some cases by factor of four.
\end{itemize}
Overall, using our method, significantly improves the reliability of GNNs, thus, being highly beneficial when, e.g., using them in real production systems or scientific applications.

%% file: sections/preliminaries.tex
\section{Related Work}
The sensitivity of machine learning models w.r.t.\ adversarial perturbations has been studied extensively \cite{1412.6572.pdf} .
Only recently, however, researchers have started to investigate adversarial attacks on graph neural networks \citep{zugner2018adversarial, pmlr-v80-dai18b,zugner2019meta} and node embeddings \cite{bojchevskinode}. All of these works focus on generating adversarial examples. In contrast, we provide the first work to certify and improve the robustness of GNNs. As shown in \citep{zugner2018adversarial}, both perturbations to the node attributes as well as the graph structure are harmful. In this work, we focus on perturbations of the node attributes and we leave structure perturbations for future work. 

For `classical' neural networks various heuristic approaches have been proposed to improve the the robustness to adversarial examples \cite{DBLP:conf/sp/PapernotM0JS16}. However, such heuristics are often broken by new attack methods, leading to an arms race. As an alternative, recent works have considered certifiable robustness \cite{kolterpolytope,semidefinite,DBLP:conf/nips/HeinA17,croce2018provable}
 providing guarantees that no perturbation w.r.t.\ a specific perturbation space will change an instance's prediction.

For this work, specifically the class of methods based on convex relaxations are of relevance \cite{kolterpolytope,semidefinite}. They construct a convex relaxation
for computing a lower bound on the worst-case margin achievable over all possible perturbations. This bound serves as a certificate of robustness. Solving such convex optimization problems can often been done efficiently, and by exploiting duality it enables to even train a robust model \cite{kolterpolytope}. As already mentioned, our work differs significantly from the existing methods since (i) it considers the novel GNN domain with its relational dependencies, (ii) it handles a discrete/binary data domain, while existing works have only handled continuous data; thus also leading to very different constraints on the perturbations, and (iii) we propose a novel robust training procedure which specifically exploits the semi-supervised learning setting of GNNs, i.e.\ using the unlabeled nodes as well.

%% file: sections/model.tex
\input{sections/model-intro.tex}

\input{sections/model-linear-program.tex}
\input{sections/model-relaxation.tex}

\input{sections/model-dual-program.tex}
\input{sections/model-bounds.tex}
\input{sections/model-robust-training.tex}

%% file: sections/model-intro.tex
\section{Preliminaries}

We consider the task of (semi-supervised) node classification in a single large graph having binary node features. 
Let $G=({\A}, {\X})$ be an attributed graph, where ${\A} \in\{0,1\}^{N\times N}$ is the adjacency matrix 
and ${\X} \in \{0,1\}^{N\times D}$ represents the nodes' features. 
W.l.o.g.\ we assume the node-ids to be $\mathcal{V}=\{1,\ldots,N\}$. 
Given a subset $\mathcal{V}_L\subseteq \mathcal{V}$ of labeled nodes, with class labels from $\mathcal{C} = \{1, 2, \dots, K\}$, the goal of node classification is to learn a function $f: \mathcal{V} \rightarrow \mathcal{C}$ which maps each node $v\in \mathcal{V}$ to one class in $\mathcal{C}$.
In this work, we focus on node classification employing graph neural networks. In particular, we consider graph convolutional networks where the latent representations $\hidden^{(\iterlayer)}$ at layer $\iterlayer$ are of the form
\begin{align}
    \label{eq:GNN}
    &\hidden^{(\iterlayer)} =\sigma^{(\iterlayer)}\left( \An^{(\iterlayer-1)} \hidden^{(\iterlayer-1)} \W^{(\iterlayer-1)} + \bvec^{(\iterlayer-1)} \right)   \text{~~~for } \iterlayer = 2,...,\numlayers
\end{align}
where $ \hidden^{(1)} = \X $ and with activation functions given by
\begin{align}
    &\sigma^{(\numlayers)} \left(\cdot \right) = \softmax \left(\cdot \right),
    &&\sigma^{(\iterlayer)}\left(\cdot \right) = \relu \left(\cdot \right)  && \text{for } \iterlayer = 2,...,\numlayers-1. \notag 
\end{align}
The output $\hidden^{(\numlayers)}_{vc}$ denotes the probability of assigning node $v$ to class $c$. The $\An^{(\iterlayer)}$ are the message passing matrices that define how the activations are propagated in the network. In \GCN \cite{kipf2016semi}, for example, $\An^{(1)}=...=\An^{(L-1)}=\tilde{\bs D}^{-\frac{1}{2}} \bs{{\tilde{A}}} \tilde{\bs D}^{-\frac{1}{2}}$, where $\bs{{\tilde{A}}} = \A + \bs{I}_{\numnodes \times \numnodes}$ and $\tilde{\bs D}_{ii} = \sum_j \bs{{\tilde{A}}}_{ij}$. The  $\W^{(.)}$ and $\bvec^{(.)}$ are the trainable weights of the graph neural network, usually simply learned by minimizing the cross-entropy loss on the given labeled training nodes $\mathcal{V}_L$. 

\textit{Notations:} We denote with $\mathcal{N}_{l}(t)$ the $l$-hop neighborhood of a node $t$, i.e.\ all nodes which are reachable with $l$ hops (or less) from node $t$, including the node $t$ itself. Given a matrix $\X$, we denote its positive part with $[\X]_+=\max(\X,0)$ where the max is applied entry-wise. Similarly, the negative part is $[\X]_-=-\min(\X,0)$, which are non-negative numbers. All matrix norms $||\X||_p$ used in the paper are meant to be entry-wise, i.e.\ flattening $\X$ to a vector and applying the corresponding vector norm. We denote with $\latentsize$ the dimensionality of the latent space in layer $l$, i.e.\ $\hidden^{(l)}\in \mathbb{R}^{\numnodes\times \latentsize}$. $\X_{i:}$ denotes the $i$-th row of a matrix $\X$ and $\X_{:j}$ its $j$-th column.

\section{Certifying Robustness for Graph Convolutional Networks}

Our first goal is to derive an efficient principle for robustness certificates.
That is, given an already trained GNN and a specific node $t$ under consideration (called \textit{target} node), our goal is to provide a certificate which guarantees that the prediction made for node $t$ \textit{will not change} even if the data gets perturbed (given a specific perturbation budget). That is, if the certificate is provided, the prediction for this node is robust under \textit{any} admissible perturbations.
Unlike existing works, we cannot restrict perturbations to the instance itself due to the relational dependencies.

However, we can exploit one key insight: for a GNN with $\numlayers$ layers, the output $\hidden^{(\numlayers)}_{\target:}$ of node $\target$ depends only on the nodes in its $\numlayers-1$ hop neighborhood $\mathcal{N}_{\numlayers-1}(t)$. Therefore, instead of operating with Eq. \eqref{eq:GNN}, we can `slice' the matrices $\X$ and $\An^{(l)}$ at each step to only contain the entries that are required to compute the output for the target node $\target$.\footnote{Note that the shapes of $\W$ and $\bvec$ do not change.}  
 This step drastically improves scalability -- reducing not only the size of the neural network but also the potential perturbations we have to consider later on. We define the matrix slices for a given target $t$ as follows:\footnote{To avoid clutter in the notation, since our method certifies robustness with respect to a specific node $t$, we omit explicitly mentioning the target node $\target$ in the following.} 
\begin{align}
    &\Aslice^{(\iterlayer)} = \An^{(l)}_{\mathcal{N}_{\numlayers - \iterlayer}(\target), \mathcal{N}_{\numlayers - \iterlayer+1}(\target)} \text{ for }\iterlayer=1,...,\numlayers - 1, &&\Xslice = \X_{\mathcal{N}_{\numlayers - 1}(\target):}
\end{align}
where the set indexing corresponds to slicing the rows and columns of a matrix, i.e. $\An_{\mathcal{N}_{2}(\target),\mathcal{N}_{1}(\target)}$ contains the rows corresponding to the two-hop neighbors of node $\target$ and the columns corresponding to its one-hop neighbors.  As it becomes clear, for increasing $\iterlayer$ (i.e. depth in the network), the slices of $\An^{(l)}$ become smaller, and at the final step we only need the target node's one-hop neighbors.

Overall, we only need to consider the following \textit{sliced} GNN:
\begin{align}
	\hiddenhat^{(\iterlayer)} &= \Aslice^{(l-1)} \hidden^{(\iterlayer-1)}\W^{(\iterlayer-1)}  + \bvec^{(\iterlayer-1)} && \text{for } \iterlayer=2,...,\numlayers \label{eq:Hslice} \\
	\hidden^{(\iterlayer)}_{nj} &= \max \left\{\hiddenhat^{(\iterlayer)}_{nj},0\right\} && \text{for } \iterlayer=2,...,\numlayers-1 \label{eq:Hhatslice}
\end{align}
and $\hidden^{(1)}=\Xslice$.
Here, we replaced the $\relu$ activation by its analytical form, and we denoted with $\hiddenhat^{(\iterlayer)}$ the input before applying the $\relu$, and with $\hidden^{(\iterlayer)}$ the corresponding output. Note that the matrices are getting smaller in size -- with $\hiddenhat^{(\numlayers)}$ actually reducing to a vector that represents the predicted log probabilities (logits) for node $t$ only. Note that we also omitted the $\softmax$ activation function in the final layer $L$ since for the final classification decision it is sufficient to consider the largest value of $\hiddenhat^{(\numlayers)}$. Overall, we denote the output of this sliced GNN as
$f_{\theta}^\target(\Xslice, \Aslice ) =\hiddenhat^{(\numlayers)}\in \mathbb{R}^{\numclasses}$. Here $\theta$ is the set of all parameters, i.e. $\theta=\{\W^{(\cdot)},\bvec^{(\cdot)}\}$.

%% file: sections/model-linear-program.tex
\subsection{Robustness Certificates for GNNs}
Given this set-up, we are now ready to define our actual task: We aim to verify whether no admissible perturbation changes the prediction of the target node $t$. Formally we aim to solve:
\begin{problem} \label{problem} Given a graph $G$, a target node $t$, and an GNN with parameters $\theta$. Let $y^*$ denote the class of node $t$ (e.g.\ given by the ground truth or predicted). The worst case margin 
between classes $y^*$ and $y$ achievable under 	some set $\mathcal{X}_{q,Q}(\Xslice)$ of admissible perturbations to the node attributes is given by
\begin{align}
	\label{eq:adv_goal}
	m^t(y^*,y):=~&\underset{\Xadv}{\mathrm{minimize}} \:\: f_\theta^t(\Xadv, \Aslice)_{y^*} - f_\theta^t(\Xadv, \Aslice)_{y} \\
	&\textnormal{subject to }\Xadv \in \mathcal{X}_{q,Q}(\Xslice) \notag
\end{align}
If $m^t(y^*,y)>0$ for all $y\neq y^*$, the GNN is certifiably robust w.r.t.\ node $t$ and $\mathcal{X}_{q,Q}$.
\end{problem}

If the minimum in Eq.~(\ref{eq:adv_goal}) is positive, it means that there exists \emph{no} adversarial example (within our defined admissible perturbations) that leads to the classifier changing its prediction to the other class $y$ -- i.e.\ the logits of class $y^*$ are always larger than the one of $y$. 

Setting reasonable constraints to adversarial attacks is important to obtain certificates that reflect realistic attacks. Works for \textit{classical} neural networks have constrained the adversarial examples to lie on a small $\epsilon$-ball around the original sample measured by, e.g., the infinity-norm or L2-norm \cite{kolterpolytope,semidefinite,croce2018provable}, often e.g. $\epsilon<0.1$ This is clearly not practical in our binary setting as an $\epsilon < 1$ would mean that \emph{no} attribute can be changed. To allow reasonable perturbations in a binary/discrete setting one has to allow much larger changes than the $\epsilon$-balls considered so far.

Therefore, motivated by the existing works on adversarial attacks to graphs \cite{zugner2018adversarial}, we consider a more realistic scenario: We define the set of admissible perturbations by limiting the \emph{number} of changes to the original attributes -- i.e.\  we assume a perturbation budget $Q\in \mathbb{N}$ and measure the $L_0$ norm in the change to $\Xslice$. It is important to note that in a graph setting an adversary can attack the target node by also changing the node attributes of its $\numlayers-1$ hop neighborhood. Thus, $Q$ acts as a \textit{global} perturbation budget.

However, since changing many attributes for a single node might not be desired, we also allow to limit the number of perturbations \textit{locally} -- i.e.\ \emph{for each node} in the $\numlayers - 1$ hop neighborhood we can consider a budget of $q\in \mathbb{N}$. Overall, in this work we consider admissible perturbations of the form:
\begin{align}
    \label{eq:admissible_perturbations}
    \mathcal{X}_{q,Q}(\Xslice) = &\left \{ \Xadv \; \middle | \; \Xadv_{nj} \in \{0,1\} \land \|\Xadv - \Xslice\|_{0} \leq Q \right. \\
    & \left. \land \; \|\Xadv_{n:} - \Xslice_{n:}\|_0 \leq q \; \forall  n \in \mathcal{N}_{\numlayers-1} \right  \}. \notag
\end{align}

\textbf{\textit{Challenges:}} There are two major obstacles preventing us from efficiently finding the minimum in Eq.~(\ref{eq:adv_goal}). First, our data domain is discrete, making optimization often intractable. Second, our function (i.e. the GNN) $f_\theta^t$ is nonconvex due to the nonlinear activation functions in the neural network. 
But there is hope: As we will show, we can efficiently find \textit{lower bounds} on the minimum of the original problem by performing specific relaxations of (i) the neural network, and (ii) the data domain. This means that if the lower bound is positive, we are certain that our classifier is robust w.r.t.\ the set of admissible perturbations. 
Remarkably, we will even see that our relaxation has an optimal solution which is integral. That is, we obtain an optimal solution (i.e.\ perturbation) which is binary -- thus, we can effectively handle the discrete data domain.

%% file: sections/model-relaxation.tex
\subsection{Convex Relaxations}
To make the objective function in Eq.~(\ref{eq:adv_goal}) convex, we have to find a convex relaxation of the $\relu$ activation function. While there are many ways to achieve this, we follow the approach of \cite{kolterpolytope} in this work. 
The core idea is (i) to treat the matrices $\hidden^{(\cdot)}$ and $\hiddenhat^{(\cdot)}$ in Eqs.~(\ref{eq:Hslice},\ref{eq:Hhatslice}) no longer as deterministic but as \textit{variables} one can optimize over (besides optimizing over $\Xadv$). In this view, Eqs.~(\ref{eq:Hslice},\ref{eq:Hhatslice}) become constraints the variables have to fulfill. Then, (ii) we relax the non-linear $\relu$ constraint of Eq. \eqref{eq:Hhatslice} by a set of convex ones.

In detail: Consider Eq. \eqref{eq:Hhatslice}.  Here,
 $\hiddenhat^{(\iterlayer)}_{nj} $ denotes the input to the $\relu$ activation function. Let us assume we have given some lower bounds $\lowerb^{(\iterlayer)}_{nj}$ and upper bounds $\upperb^{(\iterlayer)}_{nj}$ on this input based on the possible perturbations (in Section \ref{sec:bounds} we will discuss how to find these bounds).  We denote with $\mathcal{I}^{(\iterlayer)}$ the set of all tuples $(n,j)$ in layer $\iterlayer$ for which the lower and upper bounds differ in their sign, i.e. $\lowerb^{(\iterlayer)}_{nj}< 0 < \upperb_{nj}^{(\iterlayer)}$. We denote with $\mathcal{I}^{(\iterlayer)}_+$ and $\mathcal{I}^{(\iterlayer)}_-$ the tuples where both bounds are non-negative and non-positive, respectively.
 
Consider the case $\mathcal{I}^{(l)}$: We relax Eq. \eqref{eq:Hhatslice} using a convex envelope:
\begin{align}
    \hidden^{(\iterlayer)}_{nj} \geq \hiddenhat^{(\iterlayer)}_{nj}, & \qquad  \hidden^{(\iterlayer)}_{nj} \geq 0, \notag \\
    \hidden_{nj}^{(\iterlayer)} \left(\upperb_{nj}^{(\iterlayer)} - \lowerb^{(\iterlayer)}_{nj} \right) &\leq \upperb^{(\iterlayer)}_{nj} \left( \hiddenhat^{(\iterlayer)}_{nj} - \lowerb_{nj}^{(\iterlayer)} \right) && \text{if } (n,j) \in \mathcal{I}^{(\iterlayer)} \notag
\end{align}
 	\begin{wrapfigure}[7]{r}{0.36\columnwidth}
	\vspace*{-3.5mm}
	\hspace*{-4.5mm}
	\includegraphics[scale=0.7]{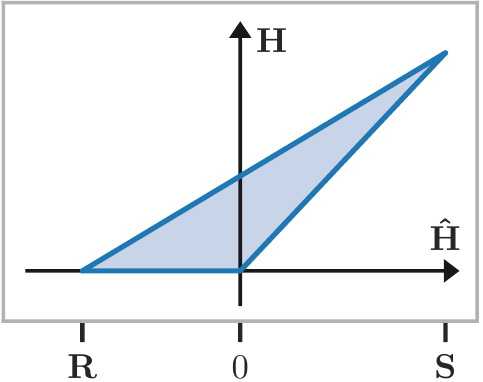}
	\vspace*{-7mm}
\end{wrapfigure}
The idea is illustrated in the figure on the right. Note that $\hidden^{(\iterlayer)}_{nj}$ is no longer the deterministic output of the $\relu$ given its input but it is a \textit{variable}. For a given input, the variable is constrained to lie on a vertical line above the input and below the upper line of the envelope.

Accordingly, but more simply, for the cases $\mathcal{I}^{(\iterlayer)}_+$ and $\mathcal{I}^{(\iterlayer)}_-$ we get:
\begin{align}
	\hidden^{(\iterlayer)}_{nj} &= \hiddenhat^{(\iterlayer)}_{nj} && \text{if } (n,j) \in \mathcal{I}^{(\iterlayer)}_+ \notag &&& 
	\hidden^{(\iterlayer)}_{nj} &= 0 && \text{if } (n,j) \in \mathcal{I}^{(\iterlayer)}_- \notag
\end{align}
which are actually not relaxations but exact conditions. Overall, Eq.~\eqref{eq:Hhatslice} has now been replaced by a set of linear (i.e.\ convex) constraints. Together with the linear constraints of Eq.~\eqref{eq:Hslice} they determine the set of admissible $\hidden^{(\cdot)}$ and $\hiddenhat^{(\cdot)}$ we can optimize over.  We denote the collection of these matrices that fulfill these constraints by ${\mathcal{Z}}_{q,Q}(\Xadv)$. 
Note that this set depends on $\Xadv$ since $\hidden^{(1)}=\Xadv$.

Overall, our problem becomes:
\begin{align}
	\label{eq:integer_linear_prog}
	\tilde{m}^\target(y^*,y):=~&\underset{\Xadv, \hidden^{(\cdot)},\hiddenhat^{(\cdot)}}{\mathrm{minimize}} \:\: \hiddenhat^{(\numlayers)}_{y^*} - \hiddenhat^{(\numlayers)}_{y} = \cvec^{\top} \hiddenhat^{(\numlayers)} \\
	&\textnormal{subject to }\Xadv \in \mathcal{X}_{q,Q}(\Xslice)~,~[\hidden^{(\cdot)},\hiddenhat^{(\cdot)}] \in {\mathcal{Z}}_{q,Q}(\Xadv)\notag 
\end{align}
Here we introduced the constant vector $\cvec=\bs{e}_{y^*}-\bs{e}_{y}$, which is $1$ at position $y^*$, $-1$ at $y$, and $0$ else. This notation clearly shows that the objective function is a simple linear function.

\begin{corollary}
	\label{coroll_lowerbound}
	The minimum in Eq.~(\ref{eq:integer_linear_prog}) is a lower bound on the minimum of the problem in Eq.~(\ref{eq:adv_goal}), i.e.\ $\tilde{m}^\target(y^*,y)\leq {m}^\target(y^*,y)$.
\end{corollary}
	\begin{proof}
		Let $\Xadv$ be the perturbation obtained by Problem~\ref{problem}, and $[\hidden^{(\cdot)},\hiddenhat^{(\cdot)}]$  the resulting \textit{exact} representations based on Eq.~\eqref{eq:Hslice}+\eqref{eq:Hhatslice}. By construction, $[\hidden^{(\cdot)},\hiddenhat^{(\cdot)}] \in \mathcal{Z}_{q,Q}(\Xslice) $. Since Eq.~\eqref{eq:integer_linear_prog} optimizes over the full set $\mathcal{Z}_{q,Q}(\Xslice) $ its minimum can not be larger.
	\end{proof}

From Corollary~\ref{coroll_lowerbound} it follows that if $\tilde{m}^\target(y^*,y) > 0$ for all $y\neq y^*$, the GNN is robust at node $t$. 
Directly solving Eq.~(\ref{eq:integer_linear_prog}), however, is still intractable due to the discrete data domain. 

As one core contribution, we will show that we can find the optimal solution in a tractable way. We proceed in two steps: (i) We first find a suitable continuous, convex relaxation of the discrete domain of possible adversarial examples. (ii) We show that the relaxed problem has an optimal solution which is integral; thus, by our specific construction the solution is binary.

More precisely, we relax the set ${\mathcal{X}}_{q,Q}(\Xslice)$ to:
\begin{align}
	\label{eq:relaxX}
    \hat{\mathcal{X}}_{q,Q}(\Xslice) = 
    &\left\{\Xadv \; \middle |  \; \Xadv_{nj} \in [0,1] \land \|\Xadv - \Xslice \|_1 \leq Q  \right. \\
    & \left .\land \; \|\Xadv_{n:} - \Xslice_{n:} \|_1 \leq q \; \forall n \in \mathcal{N}_{\numlayers-1} \right \} \notag 
\end{align}
Note that the entries of $\Xadv$ are now continuous between 0 and 1, and we have replaced the $L_0$ norm with the $L_1$ norm. This leads to:
\begin{align}
	\label{eq:linear_prog}
	\hat{m}^\target(y^*,y):=~&\underset{\Xadv, \hidden^{(\cdot)},\hiddenhat^{(\cdot)}}{\mathrm{minimize}} \:\: \hiddenhat^{(\numlayers)}_{y^*} - \hiddenhat^{(\numlayers)}_{y} = \cvec^{\top} \hiddenhat^{(\numlayers)} \\ 
	&\textnormal{subject to }\Xadv \in \hat{\mathcal{X}}_{q,Q}(\Xslice)~,~[\hidden^{(\cdot)},\hiddenhat^{(\cdot)}] \in {\mathcal{Z}}_{q,Q}(\Xadv) \notag 
\end{align}
It is worth mentioning that Eq. \eqref{eq:linear_prog} is a linear problem since besides the linear objective function also all constraints are linear. 
We provide the explicit form of this linear program in the appendix. Accordingly, Eq. \eqref{eq:linear_prog} can be solved optimally in a tractable way. 
Since $\hat{\mathcal{X}}_{q,Q}(\Xslice)\supset{\mathcal{X}}_{q,Q}(\Xslice)$ , we trivially have $\hat{m}^\target(y^*,y)\leq \tilde{m}^\target(y^*,y)$. But even more, we obtain:
\begin{theorem}
	\label{theorem:equality}
	The minimum in Eq.~(\ref{eq:integer_linear_prog}) is equal to the minimum in Eq.~(\ref{eq:linear_prog}), i.e.\ $\tilde{m}^\target(y^*,y)= \hat{m}^\target(y^*,y)$.
\end{theorem}

We will proof this theorem later (see Sec.~\ref{sec:certificate}) since it requires some further results.
In summary, using Theorem \ref{theorem:equality}, we can indeed handle the discrete data domain/discrete perturbations exactly and tractably by simply solving Eq. \eqref{eq:linear_prog} instead of Eq. \eqref{eq:integer_linear_prog}.

%% file: sections/model-dual-program.tex
\subsection{Efficient Lower Bounds via the Dual}
\label{sec:certificates}

In order to provide a robustness \emph{guarantee} w.r.t.\ the perturbations on $\Xslice$, we have to find the \textit{minimum} of the linear program in Eq. \eqref{eq:linear_prog} to ensure that we have covered the worst case. While it is possible to solve linear programs `efficiently' using highly optimized linear program solvers, the potentially large number of variables in a GNN makes this approach rather slow. As an alternative, we can consider the dual of the linear program \cite{kolterpolytope}. There, \emph{any} dual-feasible solution is a lower bound on the minimum of the primal problem. That is, if we find \textit{any} dual-feasible solution for which the objective function of the dual is positive, we know that the minimum of the primal problem has to be positive as well, guaranteeing robustness of the GNN w.r.t. any perturbation in the set. 

\begin{theorem}
    \label{thm:dual_problem}
    The dual of Eq.~(\ref{eq:linear_prog}) is equivalent to:
    \begin{align}
        \underset{\alphavec, \Mlocal, \Mglobal}{\mathrm{maximize}} \qquad  &g^\target_{q,Q}\left(\Xslice, \cvec, \alphavec, \Mlocal, \Mglobal \right) \\
        \textnormal{subject to} \notag\\
        & \alphavec^{(\iterlayer)} \in [0,1]^{|\mathcal{N}_{\numlayers-\iterlayer}| \times \latentsize} \text{ for } \iterlayer = \numlayers-1, ..., 2, \notag \\
        & \Mlocal \in \mathbb{R}_{\geq 0}^{|\mathcal{N}_{\numlayers-1}|}, ~~~ \Mglobal \in \mathbb{R}_{\geq 0} \notag
    \end{align}
    where
\begin{align}
    g^t_{q,Q}(...)=& \sum_{\iterlayer=2}^{\numlayers-1} \sum_{(n,j) \in \mathcal{I}^{(\iterlayer)}} \frac{\upperb_{nj}^{(\iterlayer)} \lowerb_{nj}^{(\iterlayer)}}{\upperb_{nj}^{(\iterlayer)} - \lowerb_{nj}^{(\iterlayer)}} \left[\Vhat_{nj}^{(\iterlayer)} \right]_+ 
     - \sum_{\iterlayer=1}^{\numlayers-1} \bs{1}^\top \V^{{(\iterlayer+1)}} \bvec^{(\iterlayer)}  \notag \\
     &  - \Tr \left[ \Xslice^\top \Vhat^{(1)} \right ]  \notag  
     - \|  \betavec\|_1 
     - q\cdot \sum_{n}\Mlocal_n - Q\cdot \Mglobal  \notag 
\end{align}
and 
\begin{align}
    & \V^{(\numlayers)} = -\cvec \in \mathbb{R}^{k} \notag\\
    & \Vhat^{(\iterlayer)} = \Aslice^{(\iterlayer)\top} \V^{(\iterlayer+1)} \W^{(\iterlayer)\top}  \in \mathbb{R}^{|\mathcal{N}_{\numlayers-\iterlayer}| \times \latentsize } \text{ for } \iterlayer = \numlayers-1, ..., 1 \notag\\
    & \V_{nj}^{(\iterlayer)} =  \begin{cases}
        0  & \textnormal{ if } (n,j) \in \mathcal{I}^{(\iterlayer)}_{-}\\
        \Vhat_{nj}  & \textnormal{ if } (n,j) \in \mathcal{I}^{(\iterlayer)}_{+}\\
        \frac{\upperb_{nj}^{(\iterlayer)}}{\upperb_{nj}^{(\iterlayer)} - \lowerb_{nj}^{(\iterlayer)}}\left[\Vhat_{nj}^{(\iterlayer)} \right]_+ - \alphavec^{(\iterlayer)}_{nj} \left[\Vhat_{nj}^{(\iterlayer)} \right]_- & \textnormal{ if } (n,j) \in \mathcal{I}^{(\iterlayer)}   
    \end{cases} \notag \\
    & \qquad \qquad \qquad \qquad \qquad \qquad \quad \: \:  \text{for } \iterlayer = \numlayers-1, ..., 2 \notag \\
    & \betavec_{nd} =  \max\left\{\bs \Delta_{nd}  - (\Mlocal_n + \Mglobal), 0 \right\} \notag \\
     & \bs \Delta_{nd} = \left[ \Vhat_{nd}^{(1)} \right]_+ \cdot (1-\Xslice_{nd})  + \left[ \Vhat_{nd}^{(1)} \right]_{-} \cdot \Xslice_{nd} \notag
\end{align}
\end{theorem}
The proof is given in the appendix. Note that parts of the dual problem in Theorem~\ref{thm:dual_problem} have a similar form to the problem in \cite{kolterpolytope}. For instance, we can interpret this dual problem as a \emph{backward} pass on a GNN, where the $\Vhat^{(\iterlayer)}$ and $\V^{(\iterlayer)}$ are the hidden representations of the respective nodes in the graph. Crucially different, however, is the propagation in the dual problem with the message passing matrices $\Aslice$ coming from the GNN formulation where neighboring nodes influence each other. Furthermore, our novel perturbation constraints from Eq. \eqref{eq:relaxX} lead to the dual variables $\Mlocal$ and $\Mglobal$, which have their origin in the local ($q$) and global ($Q$) constraints, respectively. Note that, in principle, our framework allows for different budgets $q$ per node. The term $\betavec$ has its origin in the constraint $\Xadv_{nj} \in [0,1]$.
While on the first look, the above dual problem seems rather complicated, its specific form makes it amenable for \textit{easy optimization}. The variables $\alphavec, \Mlocal, \Mglobal$ have only simple, element-wise constraints (e.g. clipping between $[0,1]$). All other terms are just deterministic assignments. Thus, straightforward optimization using (projected) gradient ascent in combination with any modern automatic differentiation framework (e.g.\ TensorFlow, PyTorch) is possible.

Furthermore, while in the above dual we need to optimize over $\Mlocal$ and $\Mglobal$, it turns out that we can simplify it even further: for any feasible $\alphavec$, we get an optimal closed-form solution for $\Mlocal, \Mglobal$. 
\begin{theorem}
	\label{thm:M_opt}
	Given the dual problem from Theorem~\ref{thm:dual_problem} and any dual-feasible value for $\alphavec$. For each node $n \in \mathcal{N}_{\numlayers - 1}$, let $S_n$ be the set of dimensions $d$ corresponding to the $q$ largest values from the vector $\bs \Delta_{n:}$ (ties broken arbitrarily). Further, denote with $o_n=\min_{d\in S_n} \Delta_{nd}$ the smallest of these values. The optimal $\Mglobal$ that maximizes the dual is the $Q$-th largest value from $[\Delta_{nd}]_{n\in \mathcal{N}_{\numlayers - 1},d\in S_n}$. For later use we denote with $S_Q$ the  set of tuples $(n,d)$ corresponding to these $Q$-largest values.
	Moreover, the optimal $\Mlocal_n$ is $\Mlocal_n=\max\left \{0, o_n - \Mglobal\right \}$. 
\end{theorem}
The proof is given in the appendix. Using Theo. \ref{thm:M_opt}, we obtain an even more compact dual where we only have to optimize over $\alphavec$. Importantly, the calculations done in Theo.~\ref{thm:M_opt} are also available in many modern automatic differentiation frameworks (i.e.\ we can back-propagate through them). Thus, we still get very efficient (and easy to implement) optimization.

\textbf{\textit{Default value:}} As mentioned before, it is not required to solve the dual problem optimally. Any dual-feasible solution leads to a lower bound on the original problem. Specifically, we can also just evaluate the function $g^t_{q,Q}$ \textit{once} given a single instantiation for $\alphavec$. This makes the computation of robustness certificates extremely fast.
For example, adopting the result of \cite{kolterpolytope}, instead of optimizing over $\alphavec$ we can set it to
\begin{equation}\label{eq:defaultvalue}
\alphavec^{(l)}_{nj} = {\upperb_{nj}^{(l)}}\cdot {(\upperb_{nj}^{(l)} - \lowerb_{nj}^{(l)})^{-1}},
\end{equation}
which is dual-feasible, and still obtain strong robustness certificates. In our experimental section, we compare the results obtained using this default value to results for optimizing over $\alphavec$. Note that using Theo.~\ref{thm:M_opt} we always ensure to use the optimal $\Mlocal, \Mglobal$ w.r.t.\ $\alphavec$.

\subsection{Primal Solutions and Certificates}\label{sec:certificate}

Based on the above results, we can now prove the following:
\begin{corollary}\label{corr:integral}
Eq.~\eqref{eq:linear_prog} is an integral linear program with respect to the variables $\Xadv$.
\end{corollary}

The proof is given in the appendix. Using this result, it is now straightforward to prove Theo. \ref{theorem:equality} from the beginning.
\begin{proof}
Since Eq.~\eqref{eq:linear_prog} has an optimal (thus, feasible) solution where $\Xadv$ is integral, we have $\Xadv\in \hat{\mathcal{X}}_{q,Q}(\Xslice)$ and, thus, $\Xadv$ has to be binary to be integral. Since in this case the $L_1$ constraints are equivalent to the $L_0$ constraints, it follows that $\Xadv\in \mathcal{X}_{q,Q}(\Xslice)$. Thus, this optimal solution of Eq.~\ref{eq:linear_prog} is feasible for Eq.~\ref{eq:integer_linear_prog}  as well. Together with $\hat{m}^\target(y^*,y)\leq \tilde{m}^\target(y^*,y)$ it follows that $\hat{m}^\target(y^*,y)= \tilde{m}^\target(y^*,y)$.
\end{proof}

In the proof of Corollary \ref{corr:integral}, we have seen that in the optimal solution, the set $\{(n,d)\in S_Q\mid \bs \Delta_{nd} > 0\}=:P$ indicates those elements which are perturbed.
That is, we constructed the worst-case perturbation. Clearly, this mechanism can also be used even if $\alphavec$ (and, thus, $\bs \Delta$) is not optimal: simply perturbing the elements in $P$. In this case, of course, the primal solution might not be optimal and we cannot use it for a robustness certificate. However, since the resulting perturbation is primal feasible (regarding the set $\mathcal{X}_{q,Q}(\Xslice)$), we can use it for our non-robustness certificate: After constructing the perturbation $\Xadv$ based on $P$, we pass it through the \textit{exact} GNN, i.e.\ we evaluate Eq.~\eqref{eq:adv_goal}. If the value is negative, we found a harmful perturbation, certifying non-robustness.

\textbf{\textit{In summary:}} By considering the dual program, we obtain robustness certificates if the obtained (dual) values are positive for every $y\neq y^*$. In contrast, by constructing the primal feasible perturbation using $P$, we obtain non-robustness certificates if the obtained (exact, primal) values are negative for one $y\neq y^*$. For some nodes, neither of these certificates can be given. We analyze this aspect in more detail in our experiments.

%% file: sections/model-bounds.tex
\subsection{Activation Bounds}\label{sec:bounds}
One crucial component of our method, the computation of the bounds $\lowerb^{(\iterlayer)}$ and $\upperb^{(\iterlayer)}$ on the activations in the relaxed GNN, remains to be defined.
Again, existing bounds for classical neural networks are not applicable since they neither consider $L_0$ constraints nor do they take neighboring instances into account. Obtaining good upper and lower bounds is crucial to obtain robustness certificates, as tighter bounds lead to lower relaxation error of the GNN activations. 

While in Sec.~\ref{sec:certificates}, we relax the discreteness condition of the node attributes $\Xslice$ in the linear program, it turns out that for the bounds the binary nature of the data can be exploited. 
More precisely, for every node $m \in \mathcal{N}_{\numlayers-2}(t)$, we compute the upper bound $\upperb^{(2)}_{mj}$ in the second layer for latent dimension $j$ as
\begin{align}
	\label{eq:upperbound}
	\upperb^{(2)}_{mj} &= \sumtopQ \left([  \Aslice^{(1)}_{mn} \hat{\bs \upperb}^{(2)}_{n j i}
	]_{n \in \mathcal{N}_{1}(m), i \in \{1,\ldots,q\}}  \right) + \unperturbed_{mj} \\
	\hat{\bs \upperb}_{nji}^{(2)} &= \ilargest\left((1-\Xslice_{n:}) \odot \left[\W^{(1)}_{:j}\right]_+ + \Xslice_{n:} \odot \left[\W^{(1)}_{:j}\right]_- \right) \notag 
\end{align}
Here, $\ilargest(\cdot)$ denotes the selection of the $i$-th largest element from the corresponding vector, and $\sumtopQ(\cdot)$ the sum of the $Q$ largest elements from the corresponding list. The first term of the sum in Eq.~(\ref{eq:upperbound}) is an upper bound on the \emph{change/increase} in the first hidden layer's activations of node $m$ and hidden dimension $j$ for any admissible perturbation on the attributes $\Xslice$. The second term are the hidden activations obtained for the (un-perturbed) input $\Xslice$, i.e.\ $\unperturbed_{mj}= \Aslice^{(1)} \Xslice\W^{(1)}  + \bvec^{(1)}$. In sum we have an upper bound on the hidden activations in the first hidden layer for the perturbed input $\Xadv$. Note that, reflecting the interdependence of nodes in the graph, the bounds of a node $m$ depend on the attributes of its neighbors $n$.

Likewise for the lower bound we use:
\begin{align}
	\label{eq:lowerbound}
	\lowerb^{(2)}_{mj} &= \textnormal{-}\sumtopQ \left([  \Aslice^{(1)}_{mn} \hat{\bs {\lowerb}}^{(2)}_{n j i}
	]_{n \in \mathcal{N}_{1}(m), i \in \{1,\ldots,q\}}  \right)\!+\!\unperturbed_{mj} \\
	\hat{\bs \lowerb}_{nji}^{(2)} &= \ilargest\left(\Xslice_{n:} \odot \left[\W^{(1)}_{:j}\right]_+ + (1-\Xslice_{n:}) \odot \left[\W^{(1)}_{:j}\right]_- \right)  \notag 
\end{align}

We need to compute the bounds for each node in the $\numlayers-2$ hop neighborhood of the target, i.e. for a GNN with a single hidden layer ($\numlayers=3$) we have $\lowerb^{(2)}, \upperb^{(2)} \in \mathbb{R}^{\mathcal{N}_{1}(t) \times h^{(2)}}$.

\begin{corollary}\label{cor:bounds}
Eqs.~(\ref{eq:upperbound}) and (\ref{eq:lowerbound}) are valid, and the tightest possible, lower/upper bounds w.r.t.\ the set of admissible perturbations.
\end{corollary}

The proof is in the appendix. 
For the remaining layers, since the input to them is no longer binary, we adapt the bounds proposed in \cite{semidefinite}. Generalized to the GNN we therefore obtain:
\begin{align}
    \lowerb^{(\iterlayer)} &= \Aslice^{(\iterlayer-1)} \left(\lowerb^{(\iterlayer-1)} \left[\W^{(\iterlayer-1)}\right]_+ - \upperb^{(\iterlayer-1)} \left[\W^{(\iterlayer-1)}\right]_- \right) \notag \\
    \upperb^{(\iterlayer)} &= \Aslice^{(\iterlayer-1)} \left(\upperb^{(\iterlayer-1)} \left[\W^{(\iterlayer-1)}\right]_+ - \lowerb^{(\iterlayer-1)} \left[\W^{(\iterlayer-1)}\right]_- \right) \notag \\
    \text{for }\iterlayer&=3,\ldots,\numlayers-1. \notag
\end{align}
Intuitively, for the upper bounds we assume that the activations in the previous layer take their respective upper bound wherever we have positive weights, and their lower bounds whenever we have negative weights (and the lower bounds are analogous to this). While there exist more computationally involved algorithms to compute more accurate bounds \cite{kolterpolytope}, we leave adaptation of such bounds to the graph domain for future work.

It is important to note that all bounds can be computed highly efficiently and one can even back-propagate through them -- important aspects for the robust training (Sec.~\ref{sec:training}). 
Specifically, one can compute Eqs.~(\ref{eq:upperbound}) and (\ref{eq:lowerbound}) for all $m\in \mathcal{V}$ (!) and all $j$ \textit{together} in time  $\mathcal{O}(h^{(2)}\cdot (N\cdot  \numdim +  E \cdot q ))$ where $E$ is the number of edges in the graph.
Note that $\hat{\bs \lowerb}_{nj:}^{(2)}$ can be computed in time $\mathcal{O}(\numdim)$ by unordered partial sorting; overall leading to the complexity $\mathcal{O}(N \cdot h^{(2)}\cdot \numdim)$. Likewise the sum of top Q elements can be computed in time $\mathcal{O}(\mathcal{N}_{1}(m) \cdot q)$ for every $1 \leq j \leq h^{(2)}$ and $m \in \mathcal{V}$, together leading to $\mathcal{O}(E \cdot q \cdot h^{(2)})$.

%% file: sections/model-robust-training.tex
\section{Robust Training of GNNs}\label{sec:training}

While being able to certify robustness of a given GNN by itself is extremely valuable for being able to trust the model's output in real-world applications, it is also highly desirable to train classifiers that are (certifiably) robust to adversarial attacks. In this section we show how to use our findings from before to train robust GNNs. 

Recall that the value of the dual $g$ can be interpreted as a lower bound on the margin between the two considered classes. As a shortcut, we denote with $\pvec^t_\theta(y, \alphavec^{(\cdot)})=\left[-g_{q,Q}^{\target}\left(\Xslice,\cvec^k, \alphavec^{k}\right)\right]_{1 \leq k \leq \numclasses}$ \ the $\numclasses$-dimensional vector containing the (negative) dual objective function values for any class $k$ compared to the given class $y$, i.e.\ $\cvec^{k}=\bs{e}_{y}-\bs{e}_{k}$. That is, node $t$ with class $\yt$ is certifiably robust if $\pvec^t_\theta< 0$ for all entries (except the entry at $\yt$ which is always 0). Here, $\theta$ denotes the parameters of the GNN.

First consider the training objective typically used to train GNNs for node classification:
\begin{equation}
    \underset{\theta}{\mathrm{minimize}} \quad \sum_{\target \in \mathcal{V}_{L}} \mathcal{L}\left(f^\target_{\theta}(\Xslice, \Aslice),\yt\right),
\end{equation}
where $\mathcal{L}$ is the cross entropy function (operating on the logits) and $\mathcal{V}_{L}$ the set of labeled nodes in the graph. $\yt$ denotes the (known) class label of node $\target$. 
To improve robustness, in \cite{kolterpolytope} (for classical neural networks) it has been proposed to instead optimize
\begin{equation}
    \label{eq:robust_ce}
    \underset{\theta, \left \{ \alphavec^{\target,k} \right \}_{\target \in \mathcal{V}_L, 1 \leq k \leq K}}{\mathrm{minimize}} \quad\sum_{\target \in \mathcal{V}_{L}} \mathcal{L}\left(\pvec^t_\theta(\yt, \alphavec^{\target,\cdot}), \yt \right)
\end{equation}
which is an {upper bound} on the worst-case loss achievable. 
Note that we can omit optimizing over $\alphavec$ by setting it to Eq.~\eqref{eq:defaultvalue}. We refer to the loss function in Eq.~(\ref{eq:robust_ce}) as \emph{robust cross entropy} loss.

One common issue with deep learning models is overconfidence \cite{overconfidence}, i.e. the models predicting effectively a probability of 1 for one and 0 for the other classes. Applied to Eq. \eqref{eq:robust_ce}, this means that the vector $\pvec^t_\theta$ is pushed to contain very large negative numbers: the predictions will not only be robust but  also very certain even  under the worst perturbation. To facilitate \emph{true} robustness and not false certainty in our model's predictions, we therefore propose an alternative robust loss that we refer to as \emph{robust hinge loss}:
\begin{align}
    \label{eq:robust_hinge}
    \hat{\mathcal{L}}_{\margin}\left(\bs p, y^*  \right) &= \sum_{k\neq y*} \max \left\{0, \bs{p}_k + \margin \right\}.
\end{align}
This loss is positive if $-\bs{p}^\target_{\theta k} = g_{q,Q}^{\target}\left(\Xslice,\cvec^k, \alphavec^{k}\right) < \margin$; and zero otherwise. Put simply: If the loss is zero, the node $t$ is certifiably robust -- in this case even guaranteeing a margin of at least $\margin$ to the decision boundary. Importantly, realizing even larger margins (for the worst-case) is not `rewarded'.
 
 We combine the \emph{robust hinge loss} with standard cross entropy to obtain the following robust optimization problem 
\begin{align}
\label{eq:robust_hinge_combi}
\underset{\theta, \alphavec}{\min}  &\sum_{\target \in \mathcal{V}_{L}} \hat{\mathcal{L}}_{\margin}\left( \pvec^t_\theta(\yt, \alphavec^{\target,\cdot}),\yt \right) + \mathcal{L}\left(f^t_\theta(\Xslice, \Aslice), \yt \right ).
\end{align}

Note that the cross entropy term is operating on the \emph{exact}, non-relaxed GNN, which is a strong advantage over the robust cross entropy loss that only uses the \emph{relaxed} GNN. Thus, we are using the exact GNN model for the node predictions, while the relaxed GNN is only used to ensure robustness. Effectively, if all nodes are robust, the term $\hat{\mathcal{L}}_{\margin}$ becomes zero, thus, reducing to the standard cross-entropy loss on the exact GNN (with robustness guarantee). 

	\begin{figure*}[!t!]
	\begin{minipage}[t]{0.35\textwidth}
		\centering
		\captionsetup{width=.95\linewidth}
		\input{figures/cora_ml/q_Q_plot.pgf}
		\vspace*{-9mm}
		\caption{Certificates for a GNN trained with standard training on \textsc{Cora-ML}.}\label{fig:Cora-q-Q-plot}		
	\end{minipage}%
	\begin{minipage}[t]{0.31\textwidth}
		\centering
		\captionsetup{width=.92\linewidth}	
		\resizebox{0.94\textwidth}{!}{\input{figures/cora_ml/purity_robustness_new.pgf}	}
		\vspace*{-5mm}
		\caption{Neighborhood purity correlates with robustness.}\label{fig:corapurity}
	\end{minipage}%
	\begin{minipage}[t]{0.31\textwidth}
	\centering
	\captionsetup{width=.9\linewidth}
	\resizebox{0.94\textwidth}{!}{	
	\input{figures/cora_ml/degree_robustness_new.pgf}}
	\vspace*{-5mm}
	\caption{Robustness of nodes vs.\ their degree.}\label{fig:coradegree}
\end{minipage}%
\end{figure*}

\textit{\textbf{Robustness in the semi-supervised setting:}} 
While Eq. \eqref{eq:robust_hinge_combi} improves the robustness regarding the labeled nodes, we do not consider the given unlabeled nodes. How to handle the semi-supervised setting which is prevalent in the graph domain, ensuring also robustness for the unlabeled nodes? Note that for the unlabeled nodes, we do not necessarily want robustness certificates with a very large margin (i.e. strongly negative $\pvec^t_\theta$) since the classifier's prediction may be wrong in the first place; this would mean that we encourage the classifier to make very certain predictions even when the predictions are wrong.
 Instead, we want to reflect in our model that some unlabeled nodes might be close to the decision boundary and not make overconfident predictions in these cases.

Our robust hinge loss provides a natural way to incorporate these goals. By setting a smaller margin $\margin_2$ for the unlabeled nodes, we can train our classifier to be robust, but does not encourage worst-case logit differences larger than the specified $\margin_2$. Importantly, this does \emph{not} mean that the classifier will be less certain in general, since the cross entropy term is unchanged and if the classifier is already robust, the robust hinge loss is 0. Overall:
\begin{align}
	\label{eq:robust_hinge_combi_semi}
	\underset{\theta, \alphavec}{\min}  &\sum_{\target \in \mathcal{V}_{L}} \hat{\mathcal{L}}_{\margin_1}\left( \pvec^t_\theta(\yt, \alphavec^{\target,\cdot}),\yt \right) + \mathcal{L}\left(f^t_\theta(\Xslice, \Aslice), \yt \right ) \\
	&+\sum_{\target \in \mathcal{V}\backslash\mathcal{V}_{L}} \hat{\mathcal{L}}_{\margin_2}\left(\pvec^t_\theta(\tilde{y}_{\target}, \alphavec^{\target,\cdot}), \tilde{y}_{\target} \right) \notag
\end{align}
where $\tilde{y}_t=\arg\max_k f^t_\theta(\Xslice, \Aslice)_k$ is the predicted label for node~$t$.
Note again that the unlabeled nodes are used for robustness purposes only -- making it very different to the principle of self-training (see below). Overall, Eq. \eqref{eq:robust_hinge_combi_semi} aims to correctly classify all labeled nodes using the exact GNN, while making sure that \textit{every} node has at least a margin of $M_*$ from the decision boundary even under \textit{worst-case perturbations}.

Eq. \eqref{eq:robust_hinge_combi_semi} can be optimized as is. In practice, however, we proceed as follows: We first train the GNN on the \emph{labeled} nodes using Eq.~\eqref{eq:robust_hinge_combi} until convergence. Then we train on \emph{all} nodes using Eq. \eqref{eq:robust_hinge_combi_semi} until convergence.

\textbf{\textit{Discussion:}} Note that the above idea is not applicable to the robust cross entropy loss from Eq. \eqref{eq:robust_ce}. One might argue that one could use a GNN trained using Eq. \eqref{eq:robust_ce} to compute predictions for all (or some of the) unlabeled nodes. Then, treating these predictions as the correct (soft-)labels for the nodes and recursively apply the training. This has two undesired effects: If the prediction is very uncertain (i.e.\ the soft-labels are flat), Eq.~\eqref{eq:robust_ce} tries to find a GNN where the worst-case margin exactly matches these uncertain labels (since this minimizes the cross-entropy). The GNN will be forced to keep the prediction uncertain for such instances even if it could do better. On the other hand, if the prediction is very certain (i.e.\ very peaky), Eq.~\eqref{eq:robust_ce} tries to make sure that even in the worst-case the prediction has such high certainty -- thus being overconfident in the prediction (which might even be wrong in the first place). Indeed, this case mimics the idea of self-training: In self-training, we first train our model on the labeled nodes. Subsequently, we use the predicted classes of (some of) the unlabeled nodes, pretending these are their true labels; and continue training with them as well. Self-training, however, serves an orthogonal purpose and, in principle, can be used with any of the above models.

\textbf{\textit{Summary:}} When training the GNN, the lower and upper activation bounds are treated as a function of $\theta$, i.e.\ they are updated accordingly. While this can be done efficiently as discussed in Sec.~\ref{sec:bounds}, it is still the least efficient part of our model and future work might consider incremental computations. Overall, since the dual program in Theorem~\ref{thm:dual_problem} and the upper/lower activations bounds are differentiable, we can train a robust GNN with gradient descent and standard deep learning libraries. Note again that by setting $\alphavec$ to its default value, we actually only have to optimize over $\theta$ -- like in standard training. Furthermore, computing $\pvec^t_\theta$ for the default parameters has roughly the same cost as evaluating a usual (sliced) GNN $K$ many times, i.e. it is very efficient. 

%% file: figures/cora_ml/q_Q_plot.pgf
\begingroup%
\makeatletter%
\begin{pgfpicture}%
\pgfpathrectangle{\pgfpointorigin}{\pgfqpoint{2.475000in}{1.529634in}}%
\pgfusepath{use as bounding box, clip}%
\begin{pgfscope}%
\pgfsetbuttcap%
\pgfsetmiterjoin%
\definecolor{currentfill}{rgb}{1.000000,1.000000,1.000000}%
\pgfsetfillcolor{currentfill}%
\pgfsetlinewidth{0.000000pt}%
\definecolor{currentstroke}{rgb}{1.000000,1.000000,1.000000}%
\pgfsetstrokecolor{currentstroke}%
\pgfsetdash{}{0pt}%
\pgfpathmoveto{\pgfqpoint{0.000000in}{0.000000in}}%
\pgfpathlineto{\pgfqpoint{2.475000in}{0.000000in}}%
\pgfpathlineto{\pgfqpoint{2.475000in}{1.529634in}}%
\pgfpathlineto{\pgfqpoint{0.000000in}{1.529634in}}%
\pgfpathclose%
\pgfusepath{fill}%
\end{pgfscope}%
\begin{pgfscope}%
\pgfsetbuttcap%
\pgfsetmiterjoin%
\definecolor{currentfill}{rgb}{1.000000,1.000000,1.000000}%
\pgfsetfillcolor{currentfill}%
\pgfsetlinewidth{0.000000pt}%
\definecolor{currentstroke}{rgb}{0.000000,0.000000,0.000000}%
\pgfsetstrokecolor{currentstroke}%
\pgfsetstrokeopacity{0.000000}%
\pgfsetdash{}{0pt}%
\pgfpathmoveto{\pgfqpoint{0.461147in}{0.451389in}}%
\pgfpathlineto{\pgfqpoint{2.322222in}{0.451389in}}%
\pgfpathlineto{\pgfqpoint{2.322222in}{1.376856in}}%
\pgfpathlineto{\pgfqpoint{0.461147in}{1.376856in}}%
\pgfpathclose%
\pgfusepath{fill}%
\end{pgfscope}%
\begin{pgfscope}%
\pgfsetbuttcap%
\pgfsetroundjoin%
\definecolor{currentfill}{rgb}{0.150000,0.150000,0.150000}%
\pgfsetfillcolor{currentfill}%
\pgfsetlinewidth{1.254687pt}%
\definecolor{currentstroke}{rgb}{0.150000,0.150000,0.150000}%
\pgfsetstrokecolor{currentstroke}%
\pgfsetdash{}{0pt}%
\pgfsys@defobject{currentmarker}{\pgfqpoint{0.000000in}{-0.083333in}}{\pgfqpoint{0.000000in}{0.000000in}}{%
\pgfpathmoveto{\pgfqpoint{0.000000in}{0.000000in}}%
\pgfpathlineto{\pgfqpoint{0.000000in}{-0.083333in}}%
\pgfusepath{stroke,fill}%
}%
\begin{pgfscope}%
\pgfsys@transformshift{0.545741in}{0.451389in}%
\pgfsys@useobject{currentmarker}{}%
\end{pgfscope}%
\end{pgfscope}%
\begin{pgfscope}%
\definecolor{textcolor}{rgb}{0.150000,0.150000,0.150000}%
\pgfsetstrokecolor{textcolor}%
\pgfsetfillcolor{textcolor}%
\pgftext[x=0.545741in,y=0.319444in,,top]{\color{textcolor}\rmfamily\fontsize{8.000000}{9.600000}\selectfont \(\displaystyle 0\)}%
\end{pgfscope}%
\begin{pgfscope}%
\pgfsetbuttcap%
\pgfsetroundjoin%
\definecolor{currentfill}{rgb}{0.150000,0.150000,0.150000}%
\pgfsetfillcolor{currentfill}%
\pgfsetlinewidth{1.254687pt}%
\definecolor{currentstroke}{rgb}{0.150000,0.150000,0.150000}%
\pgfsetstrokecolor{currentstroke}%
\pgfsetdash{}{0pt}%
\pgfsys@defobject{currentmarker}{\pgfqpoint{0.000000in}{-0.083333in}}{\pgfqpoint{0.000000in}{0.000000in}}{%
\pgfpathmoveto{\pgfqpoint{0.000000in}{0.000000in}}%
\pgfpathlineto{\pgfqpoint{0.000000in}{-0.083333in}}%
\pgfusepath{stroke,fill}%
}%
\begin{pgfscope}%
\pgfsys@transformshift{0.977345in}{0.451389in}%
\pgfsys@useobject{currentmarker}{}%
\end{pgfscope}%
\end{pgfscope}%
\begin{pgfscope}%
\definecolor{textcolor}{rgb}{0.150000,0.150000,0.150000}%
\pgfsetstrokecolor{textcolor}%
\pgfsetfillcolor{textcolor}%
\pgftext[x=0.977345in,y=0.319444in,,top]{\color{textcolor}\rmfamily\fontsize{8.000000}{9.600000}\selectfont \(\displaystyle 25\)}%
\end{pgfscope}%
\begin{pgfscope}%
\pgfsetbuttcap%
\pgfsetroundjoin%
\definecolor{currentfill}{rgb}{0.150000,0.150000,0.150000}%
\pgfsetfillcolor{currentfill}%
\pgfsetlinewidth{1.254687pt}%
\definecolor{currentstroke}{rgb}{0.150000,0.150000,0.150000}%
\pgfsetstrokecolor{currentstroke}%
\pgfsetdash{}{0pt}%
\pgfsys@defobject{currentmarker}{\pgfqpoint{0.000000in}{-0.083333in}}{\pgfqpoint{0.000000in}{0.000000in}}{%
\pgfpathmoveto{\pgfqpoint{0.000000in}{0.000000in}}%
\pgfpathlineto{\pgfqpoint{0.000000in}{-0.083333in}}%
\pgfusepath{stroke,fill}%
}%
\begin{pgfscope}%
\pgfsys@transformshift{1.408948in}{0.451389in}%
\pgfsys@useobject{currentmarker}{}%
\end{pgfscope}%
\end{pgfscope}%
\begin{pgfscope}%
\definecolor{textcolor}{rgb}{0.150000,0.150000,0.150000}%
\pgfsetstrokecolor{textcolor}%
\pgfsetfillcolor{textcolor}%
\pgftext[x=1.408948in,y=0.319444in,,top]{\color{textcolor}\rmfamily\fontsize{8.000000}{9.600000}\selectfont \(\displaystyle 50\)}%
\end{pgfscope}%
\begin{pgfscope}%
\pgfsetbuttcap%
\pgfsetroundjoin%
\definecolor{currentfill}{rgb}{0.150000,0.150000,0.150000}%
\pgfsetfillcolor{currentfill}%
\pgfsetlinewidth{1.254687pt}%
\definecolor{currentstroke}{rgb}{0.150000,0.150000,0.150000}%
\pgfsetstrokecolor{currentstroke}%
\pgfsetdash{}{0pt}%
\pgfsys@defobject{currentmarker}{\pgfqpoint{0.000000in}{-0.083333in}}{\pgfqpoint{0.000000in}{0.000000in}}{%
\pgfpathmoveto{\pgfqpoint{0.000000in}{0.000000in}}%
\pgfpathlineto{\pgfqpoint{0.000000in}{-0.083333in}}%
\pgfusepath{stroke,fill}%
}%
\begin{pgfscope}%
\pgfsys@transformshift{1.840552in}{0.451389in}%
\pgfsys@useobject{currentmarker}{}%
\end{pgfscope}%
\end{pgfscope}%
\begin{pgfscope}%
\definecolor{textcolor}{rgb}{0.150000,0.150000,0.150000}%
\pgfsetstrokecolor{textcolor}%
\pgfsetfillcolor{textcolor}%
\pgftext[x=1.840552in,y=0.319444in,,top]{\color{textcolor}\rmfamily\fontsize{8.000000}{9.600000}\selectfont \(\displaystyle 75\)}%
\end{pgfscope}%
\begin{pgfscope}%
\pgfsetbuttcap%
\pgfsetroundjoin%
\definecolor{currentfill}{rgb}{0.150000,0.150000,0.150000}%
\pgfsetfillcolor{currentfill}%
\pgfsetlinewidth{1.254687pt}%
\definecolor{currentstroke}{rgb}{0.150000,0.150000,0.150000}%
\pgfsetstrokecolor{currentstroke}%
\pgfsetdash{}{0pt}%
\pgfsys@defobject{currentmarker}{\pgfqpoint{0.000000in}{-0.083333in}}{\pgfqpoint{0.000000in}{0.000000in}}{%
\pgfpathmoveto{\pgfqpoint{0.000000in}{0.000000in}}%
\pgfpathlineto{\pgfqpoint{0.000000in}{-0.083333in}}%
\pgfusepath{stroke,fill}%
}%
\begin{pgfscope}%
\pgfsys@transformshift{2.272156in}{0.451389in}%
\pgfsys@useobject{currentmarker}{}%
\end{pgfscope}%
\end{pgfscope}%
\begin{pgfscope}%
\definecolor{textcolor}{rgb}{0.150000,0.150000,0.150000}%
\pgfsetstrokecolor{textcolor}%
\pgfsetfillcolor{textcolor}%
\pgftext[x=2.272156in,y=0.319444in,,top]{\color{textcolor}\rmfamily\fontsize{8.000000}{9.600000}\selectfont \(\displaystyle 100\)}%
\end{pgfscope}%
\begin{pgfscope}%
\definecolor{textcolor}{rgb}{0.150000,0.150000,0.150000}%
\pgfsetstrokecolor{textcolor}%
\pgfsetfillcolor{textcolor}%
\pgftext[x=1.391684in,y=0.221320in,,top]{\color{textcolor}\rmfamily\fontsize{8.000000}{9.600000}\selectfont Certificate w.r.t \(\displaystyle Q\)}%
\end{pgfscope}%
\begin{pgfscope}%
\pgfsetbuttcap%
\pgfsetroundjoin%
\definecolor{currentfill}{rgb}{0.150000,0.150000,0.150000}%
\pgfsetfillcolor{currentfill}%
\pgfsetlinewidth{1.254687pt}%
\definecolor{currentstroke}{rgb}{0.150000,0.150000,0.150000}%
\pgfsetstrokecolor{currentstroke}%
\pgfsetdash{}{0pt}%
\pgfsys@defobject{currentmarker}{\pgfqpoint{-0.083333in}{0.000000in}}{\pgfqpoint{0.000000in}{0.000000in}}{%
\pgfpathmoveto{\pgfqpoint{0.000000in}{0.000000in}}%
\pgfpathlineto{\pgfqpoint{-0.083333in}{0.000000in}}%
\pgfusepath{stroke,fill}%
}%
\begin{pgfscope}%
\pgfsys@transformshift{0.461147in}{0.493456in}%
\pgfsys@useobject{currentmarker}{}%
\end{pgfscope}%
\end{pgfscope}%
\begin{pgfscope}%
\definecolor{textcolor}{rgb}{0.150000,0.150000,0.150000}%
\pgfsetstrokecolor{textcolor}%
\pgfsetfillcolor{textcolor}%
\pgftext[x=0.270173in,y=0.455193in,left,base]{\color{textcolor}\rmfamily\fontsize{8.000000}{9.600000}\selectfont \(\displaystyle 0\)}%
\end{pgfscope}%
\begin{pgfscope}%
\pgfsetbuttcap%
\pgfsetroundjoin%
\definecolor{currentfill}{rgb}{0.150000,0.150000,0.150000}%
\pgfsetfillcolor{currentfill}%
\pgfsetlinewidth{1.254687pt}%
\definecolor{currentstroke}{rgb}{0.150000,0.150000,0.150000}%
\pgfsetstrokecolor{currentstroke}%
\pgfsetdash{}{0pt}%
\pgfsys@defobject{currentmarker}{\pgfqpoint{-0.083333in}{0.000000in}}{\pgfqpoint{0.000000in}{0.000000in}}{%
\pgfpathmoveto{\pgfqpoint{0.000000in}{0.000000in}}%
\pgfpathlineto{\pgfqpoint{-0.083333in}{0.000000in}}%
\pgfusepath{stroke,fill}%
}%
\begin{pgfscope}%
\pgfsys@transformshift{0.461147in}{0.901870in}%
\pgfsys@useobject{currentmarker}{}%
\end{pgfscope}%
\end{pgfscope}%
\begin{pgfscope}%
\definecolor{textcolor}{rgb}{0.150000,0.150000,0.150000}%
\pgfsetstrokecolor{textcolor}%
\pgfsetfillcolor{textcolor}%
\pgftext[x=0.211145in,y=0.863608in,left,base]{\color{textcolor}\rmfamily\fontsize{8.000000}{9.600000}\selectfont \(\displaystyle 50\)}%
\end{pgfscope}%
\begin{pgfscope}%
\pgfsetbuttcap%
\pgfsetroundjoin%
\definecolor{currentfill}{rgb}{0.150000,0.150000,0.150000}%
\pgfsetfillcolor{currentfill}%
\pgfsetlinewidth{1.254687pt}%
\definecolor{currentstroke}{rgb}{0.150000,0.150000,0.150000}%
\pgfsetstrokecolor{currentstroke}%
\pgfsetdash{}{0pt}%
\pgfsys@defobject{currentmarker}{\pgfqpoint{-0.083333in}{0.000000in}}{\pgfqpoint{0.000000in}{0.000000in}}{%
\pgfpathmoveto{\pgfqpoint{0.000000in}{0.000000in}}%
\pgfpathlineto{\pgfqpoint{-0.083333in}{0.000000in}}%
\pgfusepath{stroke,fill}%
}%
\begin{pgfscope}%
\pgfsys@transformshift{0.461147in}{1.310285in}%
\pgfsys@useobject{currentmarker}{}%
\end{pgfscope}%
\end{pgfscope}%
\begin{pgfscope}%
\definecolor{textcolor}{rgb}{0.150000,0.150000,0.150000}%
\pgfsetstrokecolor{textcolor}%
\pgfsetfillcolor{textcolor}%
\pgftext[x=0.152116in,y=1.272022in,left,base]{\color{textcolor}\rmfamily\fontsize{8.000000}{9.600000}\selectfont \(\displaystyle 100\)}%
\end{pgfscope}%
\begin{pgfscope}%
\definecolor{textcolor}{rgb}{0.150000,0.150000,0.150000}%
\pgfsetstrokecolor{textcolor}%
\pgfsetfillcolor{textcolor}%
\pgftext[x=0.193783in,y=0.914123in,,bottom,rotate=90.000000]{\color{textcolor}\rmfamily\fontsize{8.000000}{9.600000}\selectfont \% Nodes}%
\end{pgfscope}%
\begin{pgfscope}%
\pgfpathrectangle{\pgfqpoint{0.461147in}{0.451389in}}{\pgfqpoint{1.861076in}{0.925467in}}%
\pgfusepath{clip}%
\pgfsetbuttcap%
\pgfsetroundjoin%
\definecolor{currentfill}{rgb}{0.121569,0.466667,0.705882}%
\pgfsetfillcolor{currentfill}%
\pgfsetfillopacity{0.200000}%
\pgfsetlinewidth{1.003750pt}%
\definecolor{currentstroke}{rgb}{1.000000,1.000000,1.000000}%
\pgfsetstrokecolor{currentstroke}%
\pgfsetstrokeopacity{0.200000}%
\pgfsetdash{}{0pt}%
\pgfpathmoveto{\pgfqpoint{0.545741in}{1.310285in}}%
\pgfpathlineto{\pgfqpoint{0.545741in}{0.493456in}}%
\pgfpathlineto{\pgfqpoint{0.563005in}{0.493456in}}%
\pgfpathlineto{\pgfqpoint{0.580269in}{0.493456in}}%
\pgfpathlineto{\pgfqpoint{0.597533in}{0.493456in}}%
\pgfpathlineto{\pgfqpoint{0.614797in}{0.493456in}}%
\pgfpathlineto{\pgfqpoint{0.632062in}{0.493456in}}%
\pgfpathlineto{\pgfqpoint{0.649326in}{0.493456in}}%
\pgfpathlineto{\pgfqpoint{0.666590in}{0.493456in}}%
\pgfpathlineto{\pgfqpoint{0.683854in}{0.493456in}}%
\pgfpathlineto{\pgfqpoint{0.701118in}{0.493456in}}%
\pgfpathlineto{\pgfqpoint{0.718382in}{0.493456in}}%
\pgfpathlineto{\pgfqpoint{0.735647in}{0.493456in}}%
\pgfpathlineto{\pgfqpoint{0.752911in}{0.493456in}}%
\pgfpathlineto{\pgfqpoint{0.770175in}{0.493456in}}%
\pgfpathlineto{\pgfqpoint{0.787439in}{0.493456in}}%
\pgfpathlineto{\pgfqpoint{0.804703in}{0.493456in}}%
\pgfpathlineto{\pgfqpoint{0.821967in}{0.493456in}}%
\pgfpathlineto{\pgfqpoint{0.839231in}{0.493456in}}%
\pgfpathlineto{\pgfqpoint{0.856496in}{0.493456in}}%
\pgfpathlineto{\pgfqpoint{0.873760in}{0.493456in}}%
\pgfpathlineto{\pgfqpoint{0.891024in}{0.493456in}}%
\pgfpathlineto{\pgfqpoint{0.908288in}{0.493456in}}%
\pgfpathlineto{\pgfqpoint{0.925552in}{0.493456in}}%
\pgfpathlineto{\pgfqpoint{0.942816in}{0.493456in}}%
\pgfpathlineto{\pgfqpoint{0.960081in}{0.493456in}}%
\pgfpathlineto{\pgfqpoint{0.977345in}{0.493456in}}%
\pgfpathlineto{\pgfqpoint{0.994609in}{0.493456in}}%
\pgfpathlineto{\pgfqpoint{1.011873in}{0.493456in}}%
\pgfpathlineto{\pgfqpoint{1.029137in}{0.493456in}}%
\pgfpathlineto{\pgfqpoint{1.046401in}{0.493456in}}%
\pgfpathlineto{\pgfqpoint{1.063665in}{0.493456in}}%
\pgfpathlineto{\pgfqpoint{1.080930in}{0.493456in}}%
\pgfpathlineto{\pgfqpoint{1.098194in}{0.493456in}}%
\pgfpathlineto{\pgfqpoint{1.115458in}{0.493456in}}%
\pgfpathlineto{\pgfqpoint{1.132722in}{0.493456in}}%
\pgfpathlineto{\pgfqpoint{1.149986in}{0.493456in}}%
\pgfpathlineto{\pgfqpoint{1.167250in}{0.493456in}}%
\pgfpathlineto{\pgfqpoint{1.184515in}{0.493456in}}%
\pgfpathlineto{\pgfqpoint{1.201779in}{0.493456in}}%
\pgfpathlineto{\pgfqpoint{1.219043in}{0.493456in}}%
\pgfpathlineto{\pgfqpoint{1.236307in}{0.493456in}}%
\pgfpathlineto{\pgfqpoint{1.253571in}{0.493456in}}%
\pgfpathlineto{\pgfqpoint{1.270835in}{0.493456in}}%
\pgfpathlineto{\pgfqpoint{1.288099in}{0.493456in}}%
\pgfpathlineto{\pgfqpoint{1.305364in}{0.493456in}}%
\pgfpathlineto{\pgfqpoint{1.322628in}{0.493456in}}%
\pgfpathlineto{\pgfqpoint{1.339892in}{0.493456in}}%
\pgfpathlineto{\pgfqpoint{1.357156in}{0.493456in}}%
\pgfpathlineto{\pgfqpoint{1.374420in}{0.493456in}}%
\pgfpathlineto{\pgfqpoint{1.391684in}{0.493456in}}%
\pgfpathlineto{\pgfqpoint{1.408948in}{0.493456in}}%
\pgfpathlineto{\pgfqpoint{1.426213in}{0.493456in}}%
\pgfpathlineto{\pgfqpoint{1.443477in}{0.493456in}}%
\pgfpathlineto{\pgfqpoint{1.460741in}{0.493456in}}%
\pgfpathlineto{\pgfqpoint{1.478005in}{0.493456in}}%
\pgfpathlineto{\pgfqpoint{1.495269in}{0.493456in}}%
\pgfpathlineto{\pgfqpoint{1.512533in}{0.493456in}}%
\pgfpathlineto{\pgfqpoint{1.529798in}{0.493456in}}%
\pgfpathlineto{\pgfqpoint{1.547062in}{0.493456in}}%
\pgfpathlineto{\pgfqpoint{1.564326in}{0.493456in}}%
\pgfpathlineto{\pgfqpoint{1.581590in}{0.493456in}}%
\pgfpathlineto{\pgfqpoint{1.598854in}{0.493456in}}%
\pgfpathlineto{\pgfqpoint{1.616118in}{0.493456in}}%
\pgfpathlineto{\pgfqpoint{1.633382in}{0.493456in}}%
\pgfpathlineto{\pgfqpoint{1.650647in}{0.493456in}}%
\pgfpathlineto{\pgfqpoint{1.667911in}{0.493456in}}%
\pgfpathlineto{\pgfqpoint{1.685175in}{0.493456in}}%
\pgfpathlineto{\pgfqpoint{1.702439in}{0.493456in}}%
\pgfpathlineto{\pgfqpoint{1.719703in}{0.493456in}}%
\pgfpathlineto{\pgfqpoint{1.736967in}{0.493456in}}%
\pgfpathlineto{\pgfqpoint{1.754232in}{0.493456in}}%
\pgfpathlineto{\pgfqpoint{1.771496in}{0.493456in}}%
\pgfpathlineto{\pgfqpoint{1.788760in}{0.493456in}}%
\pgfpathlineto{\pgfqpoint{1.806024in}{0.493456in}}%
\pgfpathlineto{\pgfqpoint{1.823288in}{0.493456in}}%
\pgfpathlineto{\pgfqpoint{1.840552in}{0.493456in}}%
\pgfpathlineto{\pgfqpoint{1.857816in}{0.493456in}}%
\pgfpathlineto{\pgfqpoint{1.875081in}{0.493456in}}%
\pgfpathlineto{\pgfqpoint{1.892345in}{0.493456in}}%
\pgfpathlineto{\pgfqpoint{1.909609in}{0.493456in}}%
\pgfpathlineto{\pgfqpoint{1.926873in}{0.493456in}}%
\pgfpathlineto{\pgfqpoint{1.944137in}{0.493456in}}%
\pgfpathlineto{\pgfqpoint{1.961401in}{0.493456in}}%
\pgfpathlineto{\pgfqpoint{1.978666in}{0.493456in}}%
\pgfpathlineto{\pgfqpoint{1.995930in}{0.493456in}}%
\pgfpathlineto{\pgfqpoint{2.013194in}{0.493456in}}%
\pgfpathlineto{\pgfqpoint{2.030458in}{0.493456in}}%
\pgfpathlineto{\pgfqpoint{2.047722in}{0.493456in}}%
\pgfpathlineto{\pgfqpoint{2.064986in}{0.493456in}}%
\pgfpathlineto{\pgfqpoint{2.082250in}{0.493456in}}%
\pgfpathlineto{\pgfqpoint{2.099515in}{0.493456in}}%
\pgfpathlineto{\pgfqpoint{2.116779in}{0.493456in}}%
\pgfpathlineto{\pgfqpoint{2.134043in}{0.493456in}}%
\pgfpathlineto{\pgfqpoint{2.151307in}{0.493456in}}%
\pgfpathlineto{\pgfqpoint{2.168571in}{0.493456in}}%
\pgfpathlineto{\pgfqpoint{2.185835in}{0.493456in}}%
\pgfpathlineto{\pgfqpoint{2.203100in}{0.493456in}}%
\pgfpathlineto{\pgfqpoint{2.220364in}{0.493456in}}%
\pgfpathlineto{\pgfqpoint{2.237628in}{0.493456in}}%
\pgfpathlineto{\pgfqpoint{2.237628in}{0.493728in}}%
\pgfpathlineto{\pgfqpoint{2.237628in}{0.493728in}}%
\pgfpathlineto{\pgfqpoint{2.220364in}{0.493728in}}%
\pgfpathlineto{\pgfqpoint{2.203100in}{0.493728in}}%
\pgfpathlineto{\pgfqpoint{2.185835in}{0.494001in}}%
\pgfpathlineto{\pgfqpoint{2.168571in}{0.494274in}}%
\pgfpathlineto{\pgfqpoint{2.151307in}{0.494819in}}%
\pgfpathlineto{\pgfqpoint{2.134043in}{0.494819in}}%
\pgfpathlineto{\pgfqpoint{2.116779in}{0.495092in}}%
\pgfpathlineto{\pgfqpoint{2.099515in}{0.495637in}}%
\pgfpathlineto{\pgfqpoint{2.082250in}{0.496183in}}%
\pgfpathlineto{\pgfqpoint{2.064986in}{0.496183in}}%
\pgfpathlineto{\pgfqpoint{2.047722in}{0.497001in}}%
\pgfpathlineto{\pgfqpoint{2.030458in}{0.497001in}}%
\pgfpathlineto{\pgfqpoint{2.013194in}{0.497001in}}%
\pgfpathlineto{\pgfqpoint{1.995930in}{0.497547in}}%
\pgfpathlineto{\pgfqpoint{1.978666in}{0.498365in}}%
\pgfpathlineto{\pgfqpoint{1.961401in}{0.498910in}}%
\pgfpathlineto{\pgfqpoint{1.944137in}{0.498910in}}%
\pgfpathlineto{\pgfqpoint{1.926873in}{0.498910in}}%
\pgfpathlineto{\pgfqpoint{1.909609in}{0.499456in}}%
\pgfpathlineto{\pgfqpoint{1.892345in}{0.499728in}}%
\pgfpathlineto{\pgfqpoint{1.875081in}{0.500274in}}%
\pgfpathlineto{\pgfqpoint{1.857816in}{0.500547in}}%
\pgfpathlineto{\pgfqpoint{1.840552in}{0.500547in}}%
\pgfpathlineto{\pgfqpoint{1.823288in}{0.501365in}}%
\pgfpathlineto{\pgfqpoint{1.806024in}{0.501638in}}%
\pgfpathlineto{\pgfqpoint{1.788760in}{0.502183in}}%
\pgfpathlineto{\pgfqpoint{1.771496in}{0.503001in}}%
\pgfpathlineto{\pgfqpoint{1.754232in}{0.503819in}}%
\pgfpathlineto{\pgfqpoint{1.736967in}{0.503819in}}%
\pgfpathlineto{\pgfqpoint{1.719703in}{0.504092in}}%
\pgfpathlineto{\pgfqpoint{1.702439in}{0.504092in}}%
\pgfpathlineto{\pgfqpoint{1.685175in}{0.504638in}}%
\pgfpathlineto{\pgfqpoint{1.667911in}{0.505456in}}%
\pgfpathlineto{\pgfqpoint{1.650647in}{0.506001in}}%
\pgfpathlineto{\pgfqpoint{1.633382in}{0.506547in}}%
\pgfpathlineto{\pgfqpoint{1.616118in}{0.507092in}}%
\pgfpathlineto{\pgfqpoint{1.598854in}{0.507910in}}%
\pgfpathlineto{\pgfqpoint{1.581590in}{0.508456in}}%
\pgfpathlineto{\pgfqpoint{1.564326in}{0.509274in}}%
\pgfpathlineto{\pgfqpoint{1.547062in}{0.509819in}}%
\pgfpathlineto{\pgfqpoint{1.529798in}{0.511183in}}%
\pgfpathlineto{\pgfqpoint{1.512533in}{0.512547in}}%
\pgfpathlineto{\pgfqpoint{1.495269in}{0.514456in}}%
\pgfpathlineto{\pgfqpoint{1.478005in}{0.515001in}}%
\pgfpathlineto{\pgfqpoint{1.460741in}{0.515820in}}%
\pgfpathlineto{\pgfqpoint{1.443477in}{0.516910in}}%
\pgfpathlineto{\pgfqpoint{1.426213in}{0.518001in}}%
\pgfpathlineto{\pgfqpoint{1.408948in}{0.519910in}}%
\pgfpathlineto{\pgfqpoint{1.391684in}{0.521547in}}%
\pgfpathlineto{\pgfqpoint{1.374420in}{0.522365in}}%
\pgfpathlineto{\pgfqpoint{1.357156in}{0.524820in}}%
\pgfpathlineto{\pgfqpoint{1.339892in}{0.525638in}}%
\pgfpathlineto{\pgfqpoint{1.322628in}{0.528365in}}%
\pgfpathlineto{\pgfqpoint{1.305364in}{0.531092in}}%
\pgfpathlineto{\pgfqpoint{1.288099in}{0.533547in}}%
\pgfpathlineto{\pgfqpoint{1.270835in}{0.534911in}}%
\pgfpathlineto{\pgfqpoint{1.253571in}{0.538456in}}%
\pgfpathlineto{\pgfqpoint{1.236307in}{0.541729in}}%
\pgfpathlineto{\pgfqpoint{1.219043in}{0.544729in}}%
\pgfpathlineto{\pgfqpoint{1.201779in}{0.549638in}}%
\pgfpathlineto{\pgfqpoint{1.184515in}{0.554547in}}%
\pgfpathlineto{\pgfqpoint{1.167250in}{0.558911in}}%
\pgfpathlineto{\pgfqpoint{1.149986in}{0.566547in}}%
\pgfpathlineto{\pgfqpoint{1.132722in}{0.574457in}}%
\pgfpathlineto{\pgfqpoint{1.115458in}{0.585911in}}%
\pgfpathlineto{\pgfqpoint{1.098194in}{0.593821in}}%
\pgfpathlineto{\pgfqpoint{1.080930in}{0.602275in}}%
\pgfpathlineto{\pgfqpoint{1.063665in}{0.608003in}}%
\pgfpathlineto{\pgfqpoint{1.046401in}{0.618912in}}%
\pgfpathlineto{\pgfqpoint{1.029137in}{0.630094in}}%
\pgfpathlineto{\pgfqpoint{1.011873in}{0.644003in}}%
\pgfpathlineto{\pgfqpoint{0.994609in}{0.654912in}}%
\pgfpathlineto{\pgfqpoint{0.977345in}{0.669367in}}%
\pgfpathlineto{\pgfqpoint{0.960081in}{0.686004in}}%
\pgfpathlineto{\pgfqpoint{0.942816in}{0.702640in}}%
\pgfpathlineto{\pgfqpoint{0.925552in}{0.720368in}}%
\pgfpathlineto{\pgfqpoint{0.908288in}{0.739186in}}%
\pgfpathlineto{\pgfqpoint{0.891024in}{0.753914in}}%
\pgfpathlineto{\pgfqpoint{0.873760in}{0.776278in}}%
\pgfpathlineto{\pgfqpoint{0.856496in}{0.801914in}}%
\pgfpathlineto{\pgfqpoint{0.839231in}{0.823460in}}%
\pgfpathlineto{\pgfqpoint{0.821967in}{0.846097in}}%
\pgfpathlineto{\pgfqpoint{0.804703in}{0.867915in}}%
\pgfpathlineto{\pgfqpoint{0.787439in}{0.892188in}}%
\pgfpathlineto{\pgfqpoint{0.770175in}{0.911552in}}%
\pgfpathlineto{\pgfqpoint{0.752911in}{0.938007in}}%
\pgfpathlineto{\pgfqpoint{0.735647in}{0.965553in}}%
\pgfpathlineto{\pgfqpoint{0.718382in}{0.993644in}}%
\pgfpathlineto{\pgfqpoint{0.701118in}{1.024735in}}%
\pgfpathlineto{\pgfqpoint{0.683854in}{1.057736in}}%
\pgfpathlineto{\pgfqpoint{0.666590in}{1.084464in}}%
\pgfpathlineto{\pgfqpoint{0.649326in}{1.117191in}}%
\pgfpathlineto{\pgfqpoint{0.632062in}{1.153192in}}%
\pgfpathlineto{\pgfqpoint{0.614797in}{1.185919in}}%
\pgfpathlineto{\pgfqpoint{0.597533in}{1.218920in}}%
\pgfpathlineto{\pgfqpoint{0.580269in}{1.255466in}}%
\pgfpathlineto{\pgfqpoint{0.563005in}{1.310285in}}%
\pgfpathlineto{\pgfqpoint{0.545741in}{1.310285in}}%
\pgfpathclose%
\pgfusepath{stroke,fill}%
\end{pgfscope}%
\begin{pgfscope}%
\pgfpathrectangle{\pgfqpoint{0.461147in}{0.451389in}}{\pgfqpoint{1.861076in}{0.925467in}}%
\pgfusepath{clip}%
\pgfsetbuttcap%
\pgfsetroundjoin%
\definecolor{currentfill}{rgb}{1.000000,0.498039,0.054902}%
\pgfsetfillcolor{currentfill}%
\pgfsetfillopacity{0.200000}%
\pgfsetlinewidth{1.003750pt}%
\definecolor{currentstroke}{rgb}{1.000000,1.000000,1.000000}%
\pgfsetstrokecolor{currentstroke}%
\pgfsetstrokeopacity{0.200000}%
\pgfsetdash{}{0pt}%
\pgfpathmoveto{\pgfqpoint{0.545741in}{1.310285in}}%
\pgfpathlineto{\pgfqpoint{0.545741in}{1.334790in}}%
\pgfpathlineto{\pgfqpoint{0.563005in}{1.334790in}}%
\pgfpathlineto{\pgfqpoint{0.580269in}{1.334790in}}%
\pgfpathlineto{\pgfqpoint{0.597533in}{1.334790in}}%
\pgfpathlineto{\pgfqpoint{0.614797in}{1.334790in}}%
\pgfpathlineto{\pgfqpoint{0.632062in}{1.334790in}}%
\pgfpathlineto{\pgfqpoint{0.649326in}{1.334790in}}%
\pgfpathlineto{\pgfqpoint{0.666590in}{1.334790in}}%
\pgfpathlineto{\pgfqpoint{0.683854in}{1.334790in}}%
\pgfpathlineto{\pgfqpoint{0.701118in}{1.334790in}}%
\pgfpathlineto{\pgfqpoint{0.718382in}{1.334790in}}%
\pgfpathlineto{\pgfqpoint{0.735647in}{1.334790in}}%
\pgfpathlineto{\pgfqpoint{0.752911in}{1.334790in}}%
\pgfpathlineto{\pgfqpoint{0.770175in}{1.334790in}}%
\pgfpathlineto{\pgfqpoint{0.787439in}{1.334790in}}%
\pgfpathlineto{\pgfqpoint{0.804703in}{1.334790in}}%
\pgfpathlineto{\pgfqpoint{0.821967in}{1.334790in}}%
\pgfpathlineto{\pgfqpoint{0.839231in}{1.334790in}}%
\pgfpathlineto{\pgfqpoint{0.856496in}{1.334790in}}%
\pgfpathlineto{\pgfqpoint{0.873760in}{1.334790in}}%
\pgfpathlineto{\pgfqpoint{0.891024in}{1.334790in}}%
\pgfpathlineto{\pgfqpoint{0.908288in}{1.334790in}}%
\pgfpathlineto{\pgfqpoint{0.925552in}{1.334790in}}%
\pgfpathlineto{\pgfqpoint{0.942816in}{1.334790in}}%
\pgfpathlineto{\pgfqpoint{0.960081in}{1.334790in}}%
\pgfpathlineto{\pgfqpoint{0.977345in}{1.334790in}}%
\pgfpathlineto{\pgfqpoint{0.994609in}{1.334790in}}%
\pgfpathlineto{\pgfqpoint{1.011873in}{1.334790in}}%
\pgfpathlineto{\pgfqpoint{1.029137in}{1.334790in}}%
\pgfpathlineto{\pgfqpoint{1.046401in}{1.334790in}}%
\pgfpathlineto{\pgfqpoint{1.063665in}{1.334790in}}%
\pgfpathlineto{\pgfqpoint{1.080930in}{1.334790in}}%
\pgfpathlineto{\pgfqpoint{1.098194in}{1.334790in}}%
\pgfpathlineto{\pgfqpoint{1.115458in}{1.334790in}}%
\pgfpathlineto{\pgfqpoint{1.132722in}{1.334790in}}%
\pgfpathlineto{\pgfqpoint{1.149986in}{1.334790in}}%
\pgfpathlineto{\pgfqpoint{1.167250in}{1.334790in}}%
\pgfpathlineto{\pgfqpoint{1.184515in}{1.334790in}}%
\pgfpathlineto{\pgfqpoint{1.201779in}{1.334790in}}%
\pgfpathlineto{\pgfqpoint{1.219043in}{1.334790in}}%
\pgfpathlineto{\pgfqpoint{1.236307in}{1.334790in}}%
\pgfpathlineto{\pgfqpoint{1.253571in}{1.334790in}}%
\pgfpathlineto{\pgfqpoint{1.270835in}{1.334790in}}%
\pgfpathlineto{\pgfqpoint{1.288099in}{1.334790in}}%
\pgfpathlineto{\pgfqpoint{1.305364in}{1.334790in}}%
\pgfpathlineto{\pgfqpoint{1.322628in}{1.334790in}}%
\pgfpathlineto{\pgfqpoint{1.339892in}{1.334790in}}%
\pgfpathlineto{\pgfqpoint{1.357156in}{1.334790in}}%
\pgfpathlineto{\pgfqpoint{1.374420in}{1.334790in}}%
\pgfpathlineto{\pgfqpoint{1.391684in}{1.334790in}}%
\pgfpathlineto{\pgfqpoint{1.408948in}{1.334790in}}%
\pgfpathlineto{\pgfqpoint{1.426213in}{1.334790in}}%
\pgfpathlineto{\pgfqpoint{1.443477in}{1.334790in}}%
\pgfpathlineto{\pgfqpoint{1.460741in}{1.334790in}}%
\pgfpathlineto{\pgfqpoint{1.478005in}{1.334790in}}%
\pgfpathlineto{\pgfqpoint{1.495269in}{1.334790in}}%
\pgfpathlineto{\pgfqpoint{1.512533in}{1.334790in}}%
\pgfpathlineto{\pgfqpoint{1.529798in}{1.334790in}}%
\pgfpathlineto{\pgfqpoint{1.547062in}{1.334790in}}%
\pgfpathlineto{\pgfqpoint{1.564326in}{1.334790in}}%
\pgfpathlineto{\pgfqpoint{1.581590in}{1.334790in}}%
\pgfpathlineto{\pgfqpoint{1.598854in}{1.334790in}}%
\pgfpathlineto{\pgfqpoint{1.616118in}{1.334790in}}%
\pgfpathlineto{\pgfqpoint{1.633382in}{1.334790in}}%
\pgfpathlineto{\pgfqpoint{1.650647in}{1.334790in}}%
\pgfpathlineto{\pgfqpoint{1.667911in}{1.334790in}}%
\pgfpathlineto{\pgfqpoint{1.685175in}{1.334790in}}%
\pgfpathlineto{\pgfqpoint{1.702439in}{1.334790in}}%
\pgfpathlineto{\pgfqpoint{1.719703in}{1.334790in}}%
\pgfpathlineto{\pgfqpoint{1.736967in}{1.334790in}}%
\pgfpathlineto{\pgfqpoint{1.754232in}{1.334790in}}%
\pgfpathlineto{\pgfqpoint{1.771496in}{1.334790in}}%
\pgfpathlineto{\pgfqpoint{1.788760in}{1.334790in}}%
\pgfpathlineto{\pgfqpoint{1.806024in}{1.334790in}}%
\pgfpathlineto{\pgfqpoint{1.823288in}{1.334790in}}%
\pgfpathlineto{\pgfqpoint{1.840552in}{1.334790in}}%
\pgfpathlineto{\pgfqpoint{1.857816in}{1.334790in}}%
\pgfpathlineto{\pgfqpoint{1.875081in}{1.334790in}}%
\pgfpathlineto{\pgfqpoint{1.892345in}{1.334790in}}%
\pgfpathlineto{\pgfqpoint{1.909609in}{1.334790in}}%
\pgfpathlineto{\pgfqpoint{1.926873in}{1.334790in}}%
\pgfpathlineto{\pgfqpoint{1.944137in}{1.334790in}}%
\pgfpathlineto{\pgfqpoint{1.961401in}{1.334790in}}%
\pgfpathlineto{\pgfqpoint{1.978666in}{1.334790in}}%
\pgfpathlineto{\pgfqpoint{1.995930in}{1.334790in}}%
\pgfpathlineto{\pgfqpoint{2.013194in}{1.334790in}}%
\pgfpathlineto{\pgfqpoint{2.030458in}{1.334790in}}%
\pgfpathlineto{\pgfqpoint{2.047722in}{1.334790in}}%
\pgfpathlineto{\pgfqpoint{2.064986in}{1.334790in}}%
\pgfpathlineto{\pgfqpoint{2.082250in}{1.334790in}}%
\pgfpathlineto{\pgfqpoint{2.099515in}{1.334790in}}%
\pgfpathlineto{\pgfqpoint{2.116779in}{1.334790in}}%
\pgfpathlineto{\pgfqpoint{2.134043in}{1.334790in}}%
\pgfpathlineto{\pgfqpoint{2.151307in}{1.334790in}}%
\pgfpathlineto{\pgfqpoint{2.168571in}{1.334790in}}%
\pgfpathlineto{\pgfqpoint{2.185835in}{1.334790in}}%
\pgfpathlineto{\pgfqpoint{2.203100in}{1.334790in}}%
\pgfpathlineto{\pgfqpoint{2.220364in}{1.334790in}}%
\pgfpathlineto{\pgfqpoint{2.237628in}{1.334790in}}%
\pgfpathlineto{\pgfqpoint{2.237628in}{0.533002in}}%
\pgfpathlineto{\pgfqpoint{2.237628in}{0.533002in}}%
\pgfpathlineto{\pgfqpoint{2.220364in}{0.533274in}}%
\pgfpathlineto{\pgfqpoint{2.203100in}{0.534638in}}%
\pgfpathlineto{\pgfqpoint{2.185835in}{0.535183in}}%
\pgfpathlineto{\pgfqpoint{2.168571in}{0.535729in}}%
\pgfpathlineto{\pgfqpoint{2.151307in}{0.536547in}}%
\pgfpathlineto{\pgfqpoint{2.134043in}{0.537093in}}%
\pgfpathlineto{\pgfqpoint{2.116779in}{0.538183in}}%
\pgfpathlineto{\pgfqpoint{2.099515in}{0.538729in}}%
\pgfpathlineto{\pgfqpoint{2.082250in}{0.539274in}}%
\pgfpathlineto{\pgfqpoint{2.064986in}{0.539547in}}%
\pgfpathlineto{\pgfqpoint{2.047722in}{0.540365in}}%
\pgfpathlineto{\pgfqpoint{2.030458in}{0.541456in}}%
\pgfpathlineto{\pgfqpoint{2.013194in}{0.542820in}}%
\pgfpathlineto{\pgfqpoint{1.995930in}{0.544184in}}%
\pgfpathlineto{\pgfqpoint{1.978666in}{0.544456in}}%
\pgfpathlineto{\pgfqpoint{1.961401in}{0.545002in}}%
\pgfpathlineto{\pgfqpoint{1.944137in}{0.545820in}}%
\pgfpathlineto{\pgfqpoint{1.926873in}{0.546638in}}%
\pgfpathlineto{\pgfqpoint{1.909609in}{0.548002in}}%
\pgfpathlineto{\pgfqpoint{1.892345in}{0.548547in}}%
\pgfpathlineto{\pgfqpoint{1.875081in}{0.549911in}}%
\pgfpathlineto{\pgfqpoint{1.857816in}{0.551275in}}%
\pgfpathlineto{\pgfqpoint{1.840552in}{0.552093in}}%
\pgfpathlineto{\pgfqpoint{1.823288in}{0.555911in}}%
\pgfpathlineto{\pgfqpoint{1.806024in}{0.557275in}}%
\pgfpathlineto{\pgfqpoint{1.788760in}{0.560547in}}%
\pgfpathlineto{\pgfqpoint{1.771496in}{0.561911in}}%
\pgfpathlineto{\pgfqpoint{1.754232in}{0.564366in}}%
\pgfpathlineto{\pgfqpoint{1.736967in}{0.566820in}}%
\pgfpathlineto{\pgfqpoint{1.719703in}{0.567366in}}%
\pgfpathlineto{\pgfqpoint{1.702439in}{0.570911in}}%
\pgfpathlineto{\pgfqpoint{1.685175in}{0.572548in}}%
\pgfpathlineto{\pgfqpoint{1.667911in}{0.575820in}}%
\pgfpathlineto{\pgfqpoint{1.650647in}{0.577729in}}%
\pgfpathlineto{\pgfqpoint{1.633382in}{0.579911in}}%
\pgfpathlineto{\pgfqpoint{1.616118in}{0.582366in}}%
\pgfpathlineto{\pgfqpoint{1.598854in}{0.584820in}}%
\pgfpathlineto{\pgfqpoint{1.581590in}{0.589730in}}%
\pgfpathlineto{\pgfqpoint{1.564326in}{0.592730in}}%
\pgfpathlineto{\pgfqpoint{1.547062in}{0.596275in}}%
\pgfpathlineto{\pgfqpoint{1.529798in}{0.599821in}}%
\pgfpathlineto{\pgfqpoint{1.512533in}{0.602003in}}%
\pgfpathlineto{\pgfqpoint{1.495269in}{0.605275in}}%
\pgfpathlineto{\pgfqpoint{1.478005in}{0.606639in}}%
\pgfpathlineto{\pgfqpoint{1.460741in}{0.611003in}}%
\pgfpathlineto{\pgfqpoint{1.443477in}{0.614548in}}%
\pgfpathlineto{\pgfqpoint{1.426213in}{0.619730in}}%
\pgfpathlineto{\pgfqpoint{1.408948in}{0.623276in}}%
\pgfpathlineto{\pgfqpoint{1.391684in}{0.627366in}}%
\pgfpathlineto{\pgfqpoint{1.374420in}{0.632276in}}%
\pgfpathlineto{\pgfqpoint{1.357156in}{0.636912in}}%
\pgfpathlineto{\pgfqpoint{1.339892in}{0.640730in}}%
\pgfpathlineto{\pgfqpoint{1.322628in}{0.645367in}}%
\pgfpathlineto{\pgfqpoint{1.305364in}{0.649185in}}%
\pgfpathlineto{\pgfqpoint{1.288099in}{0.655458in}}%
\pgfpathlineto{\pgfqpoint{1.270835in}{0.661731in}}%
\pgfpathlineto{\pgfqpoint{1.253571in}{0.666367in}}%
\pgfpathlineto{\pgfqpoint{1.236307in}{0.670731in}}%
\pgfpathlineto{\pgfqpoint{1.219043in}{0.677276in}}%
\pgfpathlineto{\pgfqpoint{1.201779in}{0.686549in}}%
\pgfpathlineto{\pgfqpoint{1.184515in}{0.693913in}}%
\pgfpathlineto{\pgfqpoint{1.167250in}{0.699367in}}%
\pgfpathlineto{\pgfqpoint{1.149986in}{0.709458in}}%
\pgfpathlineto{\pgfqpoint{1.132722in}{0.715186in}}%
\pgfpathlineto{\pgfqpoint{1.115458in}{0.722550in}}%
\pgfpathlineto{\pgfqpoint{1.098194in}{0.731550in}}%
\pgfpathlineto{\pgfqpoint{1.080930in}{0.743004in}}%
\pgfpathlineto{\pgfqpoint{1.063665in}{0.752550in}}%
\pgfpathlineto{\pgfqpoint{1.046401in}{0.765914in}}%
\pgfpathlineto{\pgfqpoint{1.029137in}{0.780096in}}%
\pgfpathlineto{\pgfqpoint{1.011873in}{0.790732in}}%
\pgfpathlineto{\pgfqpoint{0.994609in}{0.804369in}}%
\pgfpathlineto{\pgfqpoint{0.977345in}{0.814460in}}%
\pgfpathlineto{\pgfqpoint{0.960081in}{0.830006in}}%
\pgfpathlineto{\pgfqpoint{0.942816in}{0.846097in}}%
\pgfpathlineto{\pgfqpoint{0.925552in}{0.859188in}}%
\pgfpathlineto{\pgfqpoint{0.908288in}{0.871733in}}%
\pgfpathlineto{\pgfqpoint{0.891024in}{0.882915in}}%
\pgfpathlineto{\pgfqpoint{0.873760in}{0.898734in}}%
\pgfpathlineto{\pgfqpoint{0.856496in}{0.915916in}}%
\pgfpathlineto{\pgfqpoint{0.839231in}{0.934734in}}%
\pgfpathlineto{\pgfqpoint{0.821967in}{0.950007in}}%
\pgfpathlineto{\pgfqpoint{0.804703in}{0.967735in}}%
\pgfpathlineto{\pgfqpoint{0.787439in}{0.987644in}}%
\pgfpathlineto{\pgfqpoint{0.770175in}{1.006735in}}%
\pgfpathlineto{\pgfqpoint{0.752911in}{1.029099in}}%
\pgfpathlineto{\pgfqpoint{0.735647in}{1.050918in}}%
\pgfpathlineto{\pgfqpoint{0.718382in}{1.074645in}}%
\pgfpathlineto{\pgfqpoint{0.701118in}{1.097009in}}%
\pgfpathlineto{\pgfqpoint{0.683854in}{1.121555in}}%
\pgfpathlineto{\pgfqpoint{0.666590in}{1.149919in}}%
\pgfpathlineto{\pgfqpoint{0.649326in}{1.174737in}}%
\pgfpathlineto{\pgfqpoint{0.632062in}{1.197647in}}%
\pgfpathlineto{\pgfqpoint{0.614797in}{1.225465in}}%
\pgfpathlineto{\pgfqpoint{0.597533in}{1.256829in}}%
\pgfpathlineto{\pgfqpoint{0.580269in}{1.310285in}}%
\pgfpathlineto{\pgfqpoint{0.563005in}{1.310285in}}%
\pgfpathlineto{\pgfqpoint{0.545741in}{1.310285in}}%
\pgfpathclose%
\pgfusepath{stroke,fill}%
\end{pgfscope}%
\begin{pgfscope}%
\pgfpathrectangle{\pgfqpoint{0.461147in}{0.451389in}}{\pgfqpoint{1.861076in}{0.925467in}}%
\pgfusepath{clip}%
\pgfsetroundcap%
\pgfsetroundjoin%
\pgfsetlinewidth{1.505625pt}%
\definecolor{currentstroke}{rgb}{0.298039,0.447059,0.690196}%
\pgfsetstrokecolor{currentstroke}%
\pgfsetdash{}{0pt}%
\pgfpathmoveto{\pgfqpoint{0.545741in}{1.310285in}}%
\pgfpathlineto{\pgfqpoint{0.563005in}{1.310285in}}%
\pgfpathlineto{\pgfqpoint{0.580269in}{1.255466in}}%
\pgfpathlineto{\pgfqpoint{0.597533in}{1.218920in}}%
\pgfpathlineto{\pgfqpoint{0.614797in}{1.185919in}}%
\pgfpathlineto{\pgfqpoint{0.632062in}{1.153192in}}%
\pgfpathlineto{\pgfqpoint{0.649326in}{1.117191in}}%
\pgfpathlineto{\pgfqpoint{0.666590in}{1.084464in}}%
\pgfpathlineto{\pgfqpoint{0.683854in}{1.057736in}}%
\pgfpathlineto{\pgfqpoint{0.701118in}{1.024735in}}%
\pgfpathlineto{\pgfqpoint{0.718382in}{0.993644in}}%
\pgfpathlineto{\pgfqpoint{0.735647in}{0.965553in}}%
\pgfpathlineto{\pgfqpoint{0.752911in}{0.938007in}}%
\pgfpathlineto{\pgfqpoint{0.770175in}{0.911552in}}%
\pgfpathlineto{\pgfqpoint{0.787439in}{0.892188in}}%
\pgfpathlineto{\pgfqpoint{0.804703in}{0.867915in}}%
\pgfpathlineto{\pgfqpoint{0.821967in}{0.846097in}}%
\pgfpathlineto{\pgfqpoint{0.839231in}{0.823460in}}%
\pgfpathlineto{\pgfqpoint{0.856496in}{0.801914in}}%
\pgfpathlineto{\pgfqpoint{0.873760in}{0.776278in}}%
\pgfpathlineto{\pgfqpoint{0.891024in}{0.753914in}}%
\pgfpathlineto{\pgfqpoint{0.908288in}{0.739186in}}%
\pgfpathlineto{\pgfqpoint{0.925552in}{0.720368in}}%
\pgfpathlineto{\pgfqpoint{0.942816in}{0.702640in}}%
\pgfpathlineto{\pgfqpoint{0.960081in}{0.686004in}}%
\pgfpathlineto{\pgfqpoint{0.977345in}{0.669367in}}%
\pgfpathlineto{\pgfqpoint{0.994609in}{0.654912in}}%
\pgfpathlineto{\pgfqpoint{1.011873in}{0.644003in}}%
\pgfpathlineto{\pgfqpoint{1.029137in}{0.630094in}}%
\pgfpathlineto{\pgfqpoint{1.046401in}{0.618912in}}%
\pgfpathlineto{\pgfqpoint{1.063665in}{0.608003in}}%
\pgfpathlineto{\pgfqpoint{1.080930in}{0.602275in}}%
\pgfpathlineto{\pgfqpoint{1.098194in}{0.593821in}}%
\pgfpathlineto{\pgfqpoint{1.115458in}{0.585911in}}%
\pgfpathlineto{\pgfqpoint{1.132722in}{0.574457in}}%
\pgfpathlineto{\pgfqpoint{1.149986in}{0.566547in}}%
\pgfpathlineto{\pgfqpoint{1.167250in}{0.558911in}}%
\pgfpathlineto{\pgfqpoint{1.184515in}{0.554547in}}%
\pgfpathlineto{\pgfqpoint{1.201779in}{0.549638in}}%
\pgfpathlineto{\pgfqpoint{1.219043in}{0.544729in}}%
\pgfpathlineto{\pgfqpoint{1.236307in}{0.541729in}}%
\pgfpathlineto{\pgfqpoint{1.253571in}{0.538456in}}%
\pgfpathlineto{\pgfqpoint{1.270835in}{0.534911in}}%
\pgfpathlineto{\pgfqpoint{1.288099in}{0.533547in}}%
\pgfpathlineto{\pgfqpoint{1.305364in}{0.531092in}}%
\pgfpathlineto{\pgfqpoint{1.322628in}{0.528365in}}%
\pgfpathlineto{\pgfqpoint{1.339892in}{0.525638in}}%
\pgfpathlineto{\pgfqpoint{1.357156in}{0.524820in}}%
\pgfpathlineto{\pgfqpoint{1.374420in}{0.522365in}}%
\pgfpathlineto{\pgfqpoint{1.391684in}{0.521547in}}%
\pgfpathlineto{\pgfqpoint{1.408948in}{0.519910in}}%
\pgfpathlineto{\pgfqpoint{1.426213in}{0.518001in}}%
\pgfpathlineto{\pgfqpoint{1.443477in}{0.516910in}}%
\pgfpathlineto{\pgfqpoint{1.460741in}{0.515820in}}%
\pgfpathlineto{\pgfqpoint{1.478005in}{0.515001in}}%
\pgfpathlineto{\pgfqpoint{1.495269in}{0.514456in}}%
\pgfpathlineto{\pgfqpoint{1.512533in}{0.512547in}}%
\pgfpathlineto{\pgfqpoint{1.529798in}{0.511183in}}%
\pgfpathlineto{\pgfqpoint{1.547062in}{0.509819in}}%
\pgfpathlineto{\pgfqpoint{1.564326in}{0.509274in}}%
\pgfpathlineto{\pgfqpoint{1.581590in}{0.508456in}}%
\pgfpathlineto{\pgfqpoint{1.598854in}{0.507910in}}%
\pgfpathlineto{\pgfqpoint{1.616118in}{0.507092in}}%
\pgfpathlineto{\pgfqpoint{1.633382in}{0.506547in}}%
\pgfpathlineto{\pgfqpoint{1.650647in}{0.506001in}}%
\pgfpathlineto{\pgfqpoint{1.667911in}{0.505456in}}%
\pgfpathlineto{\pgfqpoint{1.685175in}{0.504638in}}%
\pgfpathlineto{\pgfqpoint{1.702439in}{0.504092in}}%
\pgfpathlineto{\pgfqpoint{1.719703in}{0.504092in}}%
\pgfpathlineto{\pgfqpoint{1.736967in}{0.503819in}}%
\pgfpathlineto{\pgfqpoint{1.754232in}{0.503819in}}%
\pgfpathlineto{\pgfqpoint{1.771496in}{0.503001in}}%
\pgfpathlineto{\pgfqpoint{1.788760in}{0.502183in}}%
\pgfpathlineto{\pgfqpoint{1.806024in}{0.501638in}}%
\pgfpathlineto{\pgfqpoint{1.823288in}{0.501365in}}%
\pgfpathlineto{\pgfqpoint{1.840552in}{0.500547in}}%
\pgfpathlineto{\pgfqpoint{1.857816in}{0.500547in}}%
\pgfpathlineto{\pgfqpoint{1.875081in}{0.500274in}}%
\pgfpathlineto{\pgfqpoint{1.892345in}{0.499728in}}%
\pgfpathlineto{\pgfqpoint{1.909609in}{0.499456in}}%
\pgfpathlineto{\pgfqpoint{1.926873in}{0.498910in}}%
\pgfpathlineto{\pgfqpoint{1.944137in}{0.498910in}}%
\pgfpathlineto{\pgfqpoint{1.961401in}{0.498910in}}%
\pgfpathlineto{\pgfqpoint{1.978666in}{0.498365in}}%
\pgfpathlineto{\pgfqpoint{1.995930in}{0.497547in}}%
\pgfpathlineto{\pgfqpoint{2.013194in}{0.497001in}}%
\pgfpathlineto{\pgfqpoint{2.030458in}{0.497001in}}%
\pgfpathlineto{\pgfqpoint{2.047722in}{0.497001in}}%
\pgfpathlineto{\pgfqpoint{2.064986in}{0.496183in}}%
\pgfpathlineto{\pgfqpoint{2.082250in}{0.496183in}}%
\pgfpathlineto{\pgfqpoint{2.099515in}{0.495637in}}%
\pgfpathlineto{\pgfqpoint{2.116779in}{0.495092in}}%
\pgfpathlineto{\pgfqpoint{2.134043in}{0.494819in}}%
\pgfpathlineto{\pgfqpoint{2.151307in}{0.494819in}}%
\pgfpathlineto{\pgfqpoint{2.168571in}{0.494274in}}%
\pgfpathlineto{\pgfqpoint{2.185835in}{0.494001in}}%
\pgfpathlineto{\pgfqpoint{2.203100in}{0.493728in}}%
\pgfpathlineto{\pgfqpoint{2.220364in}{0.493728in}}%
\pgfpathlineto{\pgfqpoint{2.237628in}{0.493728in}}%
\pgfusepath{stroke}%
\end{pgfscope}%
\begin{pgfscope}%
\pgfpathrectangle{\pgfqpoint{0.461147in}{0.451389in}}{\pgfqpoint{1.861076in}{0.925467in}}%
\pgfusepath{clip}%
\pgfsetroundcap%
\pgfsetroundjoin%
\pgfsetlinewidth{1.505625pt}%
\definecolor{currentstroke}{rgb}{0.866667,0.517647,0.321569}%
\pgfsetstrokecolor{currentstroke}%
\pgfsetdash{}{0pt}%
\pgfpathmoveto{\pgfqpoint{0.545741in}{1.310285in}}%
\pgfpathlineto{\pgfqpoint{0.563005in}{1.310285in}}%
\pgfpathlineto{\pgfqpoint{0.580269in}{1.310285in}}%
\pgfpathlineto{\pgfqpoint{0.597533in}{1.256829in}}%
\pgfpathlineto{\pgfqpoint{0.614797in}{1.225465in}}%
\pgfpathlineto{\pgfqpoint{0.632062in}{1.197647in}}%
\pgfpathlineto{\pgfqpoint{0.649326in}{1.174737in}}%
\pgfpathlineto{\pgfqpoint{0.666590in}{1.149919in}}%
\pgfpathlineto{\pgfqpoint{0.683854in}{1.121555in}}%
\pgfpathlineto{\pgfqpoint{0.701118in}{1.097009in}}%
\pgfpathlineto{\pgfqpoint{0.718382in}{1.074645in}}%
\pgfpathlineto{\pgfqpoint{0.735647in}{1.050918in}}%
\pgfpathlineto{\pgfqpoint{0.752911in}{1.029099in}}%
\pgfpathlineto{\pgfqpoint{0.770175in}{1.006735in}}%
\pgfpathlineto{\pgfqpoint{0.787439in}{0.987644in}}%
\pgfpathlineto{\pgfqpoint{0.804703in}{0.967735in}}%
\pgfpathlineto{\pgfqpoint{0.821967in}{0.950007in}}%
\pgfpathlineto{\pgfqpoint{0.839231in}{0.934734in}}%
\pgfpathlineto{\pgfqpoint{0.856496in}{0.915916in}}%
\pgfpathlineto{\pgfqpoint{0.873760in}{0.898734in}}%
\pgfpathlineto{\pgfqpoint{0.891024in}{0.882915in}}%
\pgfpathlineto{\pgfqpoint{0.908288in}{0.871733in}}%
\pgfpathlineto{\pgfqpoint{0.925552in}{0.859188in}}%
\pgfpathlineto{\pgfqpoint{0.942816in}{0.846097in}}%
\pgfpathlineto{\pgfqpoint{0.960081in}{0.830006in}}%
\pgfpathlineto{\pgfqpoint{0.977345in}{0.814460in}}%
\pgfpathlineto{\pgfqpoint{0.994609in}{0.804369in}}%
\pgfpathlineto{\pgfqpoint{1.011873in}{0.790732in}}%
\pgfpathlineto{\pgfqpoint{1.029137in}{0.780096in}}%
\pgfpathlineto{\pgfqpoint{1.046401in}{0.765914in}}%
\pgfpathlineto{\pgfqpoint{1.063665in}{0.752550in}}%
\pgfpathlineto{\pgfqpoint{1.080930in}{0.743004in}}%
\pgfpathlineto{\pgfqpoint{1.098194in}{0.731550in}}%
\pgfpathlineto{\pgfqpoint{1.115458in}{0.722550in}}%
\pgfpathlineto{\pgfqpoint{1.132722in}{0.715186in}}%
\pgfpathlineto{\pgfqpoint{1.149986in}{0.709458in}}%
\pgfpathlineto{\pgfqpoint{1.167250in}{0.699367in}}%
\pgfpathlineto{\pgfqpoint{1.184515in}{0.693913in}}%
\pgfpathlineto{\pgfqpoint{1.201779in}{0.686549in}}%
\pgfpathlineto{\pgfqpoint{1.219043in}{0.677276in}}%
\pgfpathlineto{\pgfqpoint{1.236307in}{0.670731in}}%
\pgfpathlineto{\pgfqpoint{1.253571in}{0.666367in}}%
\pgfpathlineto{\pgfqpoint{1.270835in}{0.661731in}}%
\pgfpathlineto{\pgfqpoint{1.288099in}{0.655458in}}%
\pgfpathlineto{\pgfqpoint{1.305364in}{0.649185in}}%
\pgfpathlineto{\pgfqpoint{1.322628in}{0.645367in}}%
\pgfpathlineto{\pgfqpoint{1.339892in}{0.640730in}}%
\pgfpathlineto{\pgfqpoint{1.357156in}{0.636912in}}%
\pgfpathlineto{\pgfqpoint{1.374420in}{0.632276in}}%
\pgfpathlineto{\pgfqpoint{1.391684in}{0.627366in}}%
\pgfpathlineto{\pgfqpoint{1.408948in}{0.623276in}}%
\pgfpathlineto{\pgfqpoint{1.426213in}{0.619730in}}%
\pgfpathlineto{\pgfqpoint{1.443477in}{0.614548in}}%
\pgfpathlineto{\pgfqpoint{1.460741in}{0.611003in}}%
\pgfpathlineto{\pgfqpoint{1.478005in}{0.606639in}}%
\pgfpathlineto{\pgfqpoint{1.495269in}{0.605275in}}%
\pgfpathlineto{\pgfqpoint{1.512533in}{0.602003in}}%
\pgfpathlineto{\pgfqpoint{1.529798in}{0.599821in}}%
\pgfpathlineto{\pgfqpoint{1.547062in}{0.596275in}}%
\pgfpathlineto{\pgfqpoint{1.564326in}{0.592730in}}%
\pgfpathlineto{\pgfqpoint{1.581590in}{0.589730in}}%
\pgfpathlineto{\pgfqpoint{1.598854in}{0.584820in}}%
\pgfpathlineto{\pgfqpoint{1.616118in}{0.582366in}}%
\pgfpathlineto{\pgfqpoint{1.633382in}{0.579911in}}%
\pgfpathlineto{\pgfqpoint{1.650647in}{0.577729in}}%
\pgfpathlineto{\pgfqpoint{1.667911in}{0.575820in}}%
\pgfpathlineto{\pgfqpoint{1.685175in}{0.572548in}}%
\pgfpathlineto{\pgfqpoint{1.702439in}{0.570911in}}%
\pgfpathlineto{\pgfqpoint{1.719703in}{0.567366in}}%
\pgfpathlineto{\pgfqpoint{1.736967in}{0.566820in}}%
\pgfpathlineto{\pgfqpoint{1.754232in}{0.564366in}}%
\pgfpathlineto{\pgfqpoint{1.771496in}{0.561911in}}%
\pgfpathlineto{\pgfqpoint{1.788760in}{0.560547in}}%
\pgfpathlineto{\pgfqpoint{1.806024in}{0.557275in}}%
\pgfpathlineto{\pgfqpoint{1.823288in}{0.555911in}}%
\pgfpathlineto{\pgfqpoint{1.840552in}{0.552093in}}%
\pgfpathlineto{\pgfqpoint{1.857816in}{0.551275in}}%
\pgfpathlineto{\pgfqpoint{1.875081in}{0.549911in}}%
\pgfpathlineto{\pgfqpoint{1.892345in}{0.548547in}}%
\pgfpathlineto{\pgfqpoint{1.909609in}{0.548002in}}%
\pgfpathlineto{\pgfqpoint{1.926873in}{0.546638in}}%
\pgfpathlineto{\pgfqpoint{1.944137in}{0.545820in}}%
\pgfpathlineto{\pgfqpoint{1.961401in}{0.545002in}}%
\pgfpathlineto{\pgfqpoint{1.978666in}{0.544456in}}%
\pgfpathlineto{\pgfqpoint{1.995930in}{0.544184in}}%
\pgfpathlineto{\pgfqpoint{2.013194in}{0.542820in}}%
\pgfpathlineto{\pgfqpoint{2.030458in}{0.541456in}}%
\pgfpathlineto{\pgfqpoint{2.047722in}{0.540365in}}%
\pgfpathlineto{\pgfqpoint{2.064986in}{0.539547in}}%
\pgfpathlineto{\pgfqpoint{2.082250in}{0.539274in}}%
\pgfpathlineto{\pgfqpoint{2.099515in}{0.538729in}}%
\pgfpathlineto{\pgfqpoint{2.116779in}{0.538183in}}%
\pgfpathlineto{\pgfqpoint{2.134043in}{0.537093in}}%
\pgfpathlineto{\pgfqpoint{2.151307in}{0.536547in}}%
\pgfpathlineto{\pgfqpoint{2.168571in}{0.535729in}}%
\pgfpathlineto{\pgfqpoint{2.185835in}{0.535183in}}%
\pgfpathlineto{\pgfqpoint{2.203100in}{0.534638in}}%
\pgfpathlineto{\pgfqpoint{2.220364in}{0.533274in}}%
\pgfpathlineto{\pgfqpoint{2.237628in}{0.533002in}}%
\pgfusepath{stroke}%
\end{pgfscope}%
\begin{pgfscope}%
\pgfsetrectcap%
\pgfsetmiterjoin%
\pgfsetlinewidth{1.003750pt}%
\definecolor{currentstroke}{rgb}{0.700000,0.700000,0.700000}%
\pgfsetstrokecolor{currentstroke}%
\pgfsetdash{}{0pt}%
\pgfpathmoveto{\pgfqpoint{0.461147in}{0.451389in}}%
\pgfpathlineto{\pgfqpoint{0.461147in}{1.376856in}}%
\pgfusepath{stroke}%
\end{pgfscope}%
\begin{pgfscope}%
\pgfsetrectcap%
\pgfsetmiterjoin%
\pgfsetlinewidth{1.003750pt}%
\definecolor{currentstroke}{rgb}{0.700000,0.700000,0.700000}%
\pgfsetstrokecolor{currentstroke}%
\pgfsetdash{}{0pt}%
\pgfpathmoveto{\pgfqpoint{2.322222in}{0.451389in}}%
\pgfpathlineto{\pgfqpoint{2.322222in}{1.376856in}}%
\pgfusepath{stroke}%
\end{pgfscope}%
\begin{pgfscope}%
\pgfsetrectcap%
\pgfsetmiterjoin%
\pgfsetlinewidth{1.003750pt}%
\definecolor{currentstroke}{rgb}{0.700000,0.700000,0.700000}%
\pgfsetstrokecolor{currentstroke}%
\pgfsetdash{}{0pt}%
\pgfpathmoveto{\pgfqpoint{0.461147in}{0.451389in}}%
\pgfpathlineto{\pgfqpoint{2.322222in}{0.451389in}}%
\pgfusepath{stroke}%
\end{pgfscope}%
\begin{pgfscope}%
\pgfsetrectcap%
\pgfsetmiterjoin%
\pgfsetlinewidth{1.003750pt}%
\definecolor{currentstroke}{rgb}{0.700000,0.700000,0.700000}%
\pgfsetstrokecolor{currentstroke}%
\pgfsetdash{}{0pt}%
\pgfpathmoveto{\pgfqpoint{0.461147in}{1.376856in}}%
\pgfpathlineto{\pgfqpoint{2.322222in}{1.376856in}}%
\pgfusepath{stroke}%
\end{pgfscope}%
\begin{pgfscope}%
\pgfsetbuttcap%
\pgfsetmiterjoin%
\definecolor{currentfill}{rgb}{1.000000,1.000000,1.000000}%
\pgfsetfillcolor{currentfill}%
\pgfsetlinewidth{1.003750pt}%
\definecolor{currentstroke}{rgb}{0.800000,0.800000,0.800000}%
\pgfsetstrokecolor{currentstroke}%
\pgfsetdash{}{0pt}%
\pgfpathmoveto{\pgfqpoint{1.426881in}{0.831638in}}%
\pgfpathlineto{\pgfqpoint{2.254167in}{0.831638in}}%
\pgfpathquadraticcurveto{\pgfqpoint{2.273611in}{0.831638in}}{\pgfqpoint{2.273611in}{0.851083in}}%
\pgfpathlineto{\pgfqpoint{2.273611in}{1.308801in}}%
\pgfpathquadraticcurveto{\pgfqpoint{2.273611in}{1.328245in}}{\pgfqpoint{2.254167in}{1.328245in}}%
\pgfpathlineto{\pgfqpoint{1.426881in}{1.328245in}}%
\pgfpathquadraticcurveto{\pgfqpoint{1.407437in}{1.328245in}}{\pgfqpoint{1.407437in}{1.308801in}}%
\pgfpathlineto{\pgfqpoint{1.407437in}{0.851083in}}%
\pgfpathquadraticcurveto{\pgfqpoint{1.407437in}{0.831638in}}{\pgfqpoint{1.426881in}{0.831638in}}%
\pgfpathclose%
\pgfusepath{stroke,fill}%
\end{pgfscope}%
\begin{pgfscope}%
\pgfsetbuttcap%
\pgfsetmiterjoin%
\definecolor{currentfill}{rgb}{0.121569,0.466667,0.705882}%
\pgfsetfillcolor{currentfill}%
\pgfsetfillopacity{0.200000}%
\pgfsetlinewidth{1.003750pt}%
\definecolor{currentstroke}{rgb}{1.000000,1.000000,1.000000}%
\pgfsetstrokecolor{currentstroke}%
\pgfsetstrokeopacity{0.200000}%
\pgfsetdash{}{0pt}%
\pgfpathmoveto{\pgfqpoint{1.446326in}{1.163322in}}%
\pgfpathlineto{\pgfqpoint{1.640770in}{1.163322in}}%
\pgfpathlineto{\pgfqpoint{1.640770in}{1.231378in}}%
\pgfpathlineto{\pgfqpoint{1.446326in}{1.231378in}}%
\pgfpathclose%
\pgfusepath{stroke,fill}%
\end{pgfscope}%
\begin{pgfscope}%
\definecolor{textcolor}{rgb}{0.150000,0.150000,0.150000}%
\pgfsetstrokecolor{textcolor}%
\pgfsetfillcolor{textcolor}%
\pgftext[x=1.718548in,y=1.222397in,left,base]{\color{textcolor}\rmfamily\fontsize{7.000000}{8.400000}\selectfont Certifiably}%
\end{pgfscope}%
\begin{pgfscope}%
\definecolor{textcolor}{rgb}{0.150000,0.150000,0.150000}%
\pgfsetstrokecolor{textcolor}%
\pgfsetfillcolor{textcolor}%
\pgftext[x=1.718548in,y=1.123147in,left,base]{\color{textcolor}\rmfamily\fontsize{7.000000}{8.400000}\selectfont robust}%
\end{pgfscope}%
\begin{pgfscope}%
\pgfsetbuttcap%
\pgfsetmiterjoin%
\definecolor{currentfill}{rgb}{1.000000,0.498039,0.054902}%
\pgfsetfillcolor{currentfill}%
\pgfsetfillopacity{0.200000}%
\pgfsetlinewidth{1.003750pt}%
\definecolor{currentstroke}{rgb}{1.000000,1.000000,1.000000}%
\pgfsetstrokecolor{currentstroke}%
\pgfsetstrokeopacity{0.200000}%
\pgfsetdash{}{0pt}%
\pgfpathmoveto{\pgfqpoint{1.446326in}{0.929602in}}%
\pgfpathlineto{\pgfqpoint{1.640770in}{0.929602in}}%
\pgfpathlineto{\pgfqpoint{1.640770in}{0.997658in}}%
\pgfpathlineto{\pgfqpoint{1.446326in}{0.997658in}}%
\pgfpathclose%
\pgfusepath{stroke,fill}%
\end{pgfscope}%
\begin{pgfscope}%
\definecolor{textcolor}{rgb}{0.150000,0.150000,0.150000}%
\pgfsetstrokecolor{textcolor}%
\pgfsetfillcolor{textcolor}%
\pgftext[x=1.718548in,y=0.988677in,left,base]{\color{textcolor}\rmfamily\fontsize{7.000000}{8.400000}\selectfont Certifiably}%
\end{pgfscope}%
\begin{pgfscope}%
\definecolor{textcolor}{rgb}{0.150000,0.150000,0.150000}%
\pgfsetstrokecolor{textcolor}%
\pgfsetfillcolor{textcolor}%
\pgftext[x=1.718548in,y=0.889427in,left,base]{\color{textcolor}\rmfamily\fontsize{7.000000}{8.400000}\selectfont nonrobust}%
\end{pgfscope}%
\end{pgfpicture}%
\makeatother%
\endgroup%

%% file: sections/experiments.tex
\section{Experimental Evaluation}
Our experimental contributions are twofold. (i) We evaluate the robustness of traditionally trained GNNs using, and thus analyzing, our certification method. (ii) We show that our robust training procedure can dramatically improve GNNs' robustness while sacrificing only minimal accuracy on the unlabeled nodes.

We evaluate our method on the widely used and publicly available datasets \textsc{Cora-ML} (N=2,995, E=8,416, D=2,879, K=7) \cite{mccallum2000automating}, \textsc{Citeseer} (N=3,312, E=4,715, D=3,703, K=6) \cite{sen2008collective}, and \textsc{PubMed} (N=19,717, E=44,324, D=500, K=3) \cite{sen2008collective}. For every dataset, we allow \emph{local} (i.e. per-node) changes to the node attributes amounting to 1\% of the attribute dimension, i.e. $q=0.01\numdim$. $Q$ is analyzed in detail in the experiments reflecting different perturbation spaces.

We refer to the traditional training of GNNs as \emph{Cross Entropy} (short \emph{CE}), to the robust variant of cross entropy as \emph{Robust Cross Entropy (RCE)}, and to our hinge loss variants as 
\emph{Robust Hinge Loss (RH)} and \emph{Robust Hinge Loss with Unlabeled (RH-U)}, where the latter enforces a margin loss also on the unlabeled nodes. We set $\margin_1$, i.e. the margin on the training nodes to $\log(0.9/0.1)$ and $\margin_2$ to $\log(0.6/0.4)$ for the unlabeled nodes (\emph{RH-U} only). This means that we train the GNN to (correctly) classify the labeled nodes with output probability of 90\% \emph{in the worst case}, and the unlabeled nodes with 60\%, reflecting that we do not want our model to be overconfident on the unlabeled nodes.
Please note that we do not need to compare against graph adversarial attack models such as \cite{zugner2018adversarial} since our method gives provable guarantees on the robustness.

While our method can be used for any GNN of the form in Eq.~\eqref{eq:GNN}, we study the well-established GCN \cite{kipf2016semi}, which has shown to outperform many more complicated models. Following \cite{kipf2016semi}, we consider GCNs with one hidden layer (i.e. $\numlayers=3$), and choose a latent dimensionality of 32. We split the datasets into 10\% labeled and 90\% unlabeled nodes. See the appendix for further details.

\begin{figure*}[!t!]
	\vspace*{-2mm}
	\begin{minipage}[t]{0.33\textwidth}
		\centering
		\captionsetup{width=.95\linewidth}
		\resizebox{1\textwidth}{!}{\input{figures/cora_ml/robust_nonrobust_Q.pgf}}
		\vspace*{-10mm}
		\caption{Robust training (\textsc{Cora-ML}). Dashed lines are w/o robust training.}\label{fig:core-robust-q-q}
	\end{minipage}%
	\begin{minipage}[t]{0.33\textwidth}
		\centering
		\captionsetup{width=.9\linewidth}
		\resizebox{1\textwidth}{!}{\input{figures/citeseer/robust_nonrobust_Q.pgf}}
		\vspace*{-10mm}
		\caption{Robust training (\textsc{Citeseer}). Dashed lines are w/o robust training.}\label{fig:citeseer-robust-q-q}
		\label{fig:citeseer_certifiable}
	\end{minipage}%
	\begin{minipage}[t]{0.33\textwidth}
		\centering
		\captionsetup{width=.92\linewidth}		
		\resizebox{1\textwidth}{!}{\input{figures/cora_ml/rob_ceritifed_methods.pgf}}
		\vspace*{-10mm}
		\caption{RH-U is most successful for robustness at $Q=12$ (\textsc{Cora-ML}).}\label{fig:q-q-trainings}
		\label{fig:cora_certifiable}
	\end{minipage}%
\end{figure*}

\subsection{Certificates: Robustness of GNNs}
We first start to investigate our (non-)robustness certificates by analyzing GNNs trained using standard cross entropy training. Figure~\ref{fig:Cora-q-Q-plot} shows the main result: for varying $Q$ we report the percentage of nodes (train+test) which are certifiable robust/non-robust on \textsc{Cora-ML}. We can make two important observations: 
(i) Our certificates are often very tight. That is, the white area (nodes for which we cannot give any -- robustness or non-robustness -- certificate) is rather small. Indeed, for any given $Q$, \emph{at most} 30\% of the nodes cannot be certified across all datasets and despite no robust training, highlighting the tightness of our bounds and relaxations and the effectiveness of our certification method. 
(ii) GNNs trained traditionally are only certifiably robust up to very small perturbations. At $Q=12$, less than 55\% of the nodes are certifiably robust on \textsc{Cora-ML}. In case of \textsc{Citeseer} even less than 20\% (Table~\ref{fig:table}; training: CE). Even worse, at this point already \emph{two thirds} (for \textsc{Citeseer}) and a quarter (\textsc{Cora-ML}) of the nodes are certifiably non-robust (i.e.\ we can find adversarial examples), confirming the issues reported in \cite{zugner2018adversarial}. \textsc{PubMed} behaves similarly (as we will see later, e.g., in Table \ref{fig:table}).
In our experiments, the labeled nodes are on average more robust than the unlabeled nodes, which is not surprising given that the classifier was not trained using the labels of the latter.

We also investigate what contributes to certain nodes being more robust than others. In Figure \ref{fig:corapurity} we see that neighborhood purity (i.e.\ the share of nodes in a respective node's two-hop neighborhood that is assigned the same class by the classifier) plays an important role. On \textsc{Cora-ML}, almost \emph{all} nodes that are certifiably robust above $Q\geq 50$ have a neighborhood purity of at least 80\%. When analyzing the degree (Figure \ref{fig:coradegree}), it seems that nodes with a medium degree are most robust. While counterintuitive at first, having many neighbors also means a large surface for adversarial attacks. Nodes with low degree, in contrast, might be affected more strongly since each node in its neighborhood has a larger influence.

 	\begin{wrapfigure}[11]{r}{0.19 \textwidth}
	\vspace*{-4mm}
	\hspace*{-0mm}
	\includegraphics[scale=.95]{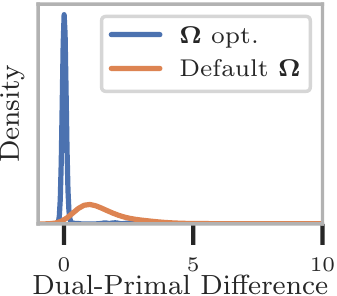}
	\vspace*{-1mm}
	\caption{Difference of Primal and Dual Bound.}\label{fig:dualprimal}
	\vspace{-30mm}
\end{wrapfigure}
\textbf{\textit{Tightness of lower bounds:}}
Next, we aim to analyze how tight our dual lower bounds are, which we needed to obtain efficient certification. For this, we analyze (i) the value of $g_{q,Q}(\cdot)$ we obtain from our dual solution (either when optimizing over $\alphavec$ are using the default value), compared to (ii) the value of the primal solution we obtain using our construction from Sec. \ref{sec:certificate}. The smaller the difference, the better. As seen in Figure \ref{fig:dualprimal}, when optimizing over $\alphavec$, for most of the nodes the gap is $0$. Thus, indeed we can often find the \textit{exact} minimum of the primal via the dual. As expected, when using the default value for $\alphavec$ the difference between dual and primal is larger. Still, for most nodes the difference is small.
Indeed, and more importantly, when considering the actual certificates (where we only need to verify whether the dual is positive; its actual value is not important), the difference between optimizing $\alphavec$ and its default value become negligible: on \textsc{Cora-ML}, the average maximal $Q$ for which we can certify robustness drops by 0.54; \textsc{Citeseer} 0.18; \textsc{PubMed} 2.3. This highlights that we can use the default values of $\alphavec$ to very efficiently certify many or even all nodes in a GNN. In all remaining experiments we, thus, only operate with this default choice.

\newcommand{\STAB}[1]{\begin{tabular}{@{}c@{}}#1\end{tabular}}

\begin{figure*}[!t!]
	\vspace{5mm}
	\begin{minipage}[t]{0.65\textwidth}
		\centering
		\captionsetup{width=.95\linewidth}
		\resizebox{0.75\textwidth}{!}{
			\begin{tabular}{clrrrr}
				\toprule
				Dataset & Training                     &  \makecell{Avg. Max\\Q robust} &  \makecell{\% Robust\\$Q=12$} &  \makecell{Acc.\\(labeled)} &  \makecell{Acc.\\(unlabeled)} \\
				\midrule
				\multirow{4}{*}{\STAB{\rotatebox[origin=c]{90}{ \textsc{Citeseer}}}} & CE &           6.77 &            0.17 &                1.00 &                  0.67 \\
				& RCE &              18.62 &            0.58 &                0.99 &                  0.69 \\
				& RH &              15.51 &            0.54 &                0.99 &                  0.68 \\
				& RH-U &              18.48 &            0.76 &                0.99 &                  0.68 \\
				\cline{1-6}
				\multirow{4}{*}{\STAB{\rotatebox[origin=c]{90}{ \textsc{Cora-ML}}}} & CE &          16.36 &            0.54 &                1.00 &                  0.83 \\
				& RCE &              38.58 &            0.77 &                1.00 &                  0.83 \\
				& RH &              32.49 &            0.74 &                1.00 &                  0.83 \\
				& RH-U &              35.58 &            0.91 &                1.00 &                  0.83 \\
				\cline{1-6}
				\multirow{4}{*}{\STAB{\rotatebox[origin=c]{90}{ \textsc{PubMed}}}} & CE &         5.82 &            0.15 &                0.99 &                  0.86 \\
				& RCE &              50.68 &            0.62 &                0.88 &                  0.84 \\
				& RH &              48.56 &            0.62 &                0.90 &                  0.85 \\
				& RH-U &              47.56 &            0.63 &                0.90 &                  0.86 \\
				
				\bottomrule
		\end{tabular}}
		\captionof{table}{Robust training results. Our robust training methods significantly improve the robustness of GNNs while not sacrificing accuracy. Robust training was done for $Q=12$. Results are averaged over five random data splits.}\label{fig:table}
	\end{minipage}%
\begin{minipage}[t]{0.32\textwidth}
\vspace{-33mm}
\begin{minipage}[t]{1\textwidth}
	\centering
	\captionsetup{width=.9\linewidth}
	
	\resizebox{0.95\textwidth}{!}{ \includegraphics[scale=1]{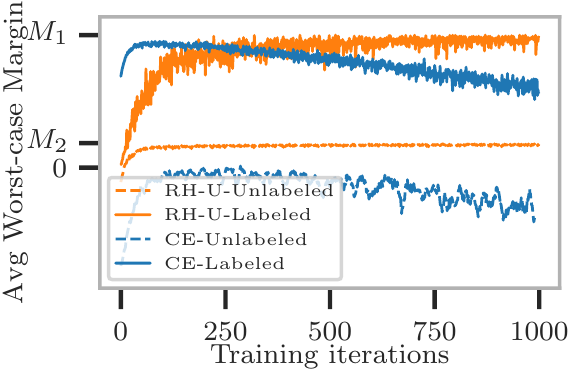}}
	\vspace*{-4mm}
	\caption{Training dynamics.}\label{fig:dynamics}
\end{minipage}%

\begin{minipage}[t]{1\textwidth}
	\centering
	\captionsetup{width=.9\linewidth}
	\vspace*{2mm}
	\resizebox{0.9\textwidth}{!}{ \includegraphics[scale=1]{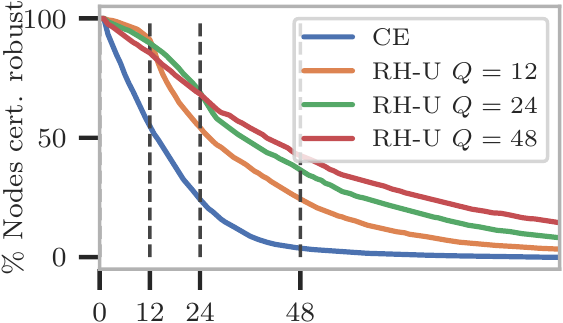}}
	\vspace*{-4mm}
	\caption{Robust training, diff. $Q$}	\label{fig:cora_different_Q}
\end{minipage}%
\end{minipage}%
\vspace*{-5mm}
\end{figure*}

\vspace*{-3mm}
\subsection{Robust Training of GNNs}
Next, we analyze our robust training procedure. If not mentioned otherwise, we use our robust hinge-loss including the unlabeled nodes \emph{RH-U} and we robustify the models with $Q=12$ since for this value more than 50\% of nodes across our datasets were not certifiably robust (when using standard training).

Figure \ref{fig:core-robust-q-q} and \ref{fig:citeseer-robust-q-q} show again the percentage of certified nodes w.r.t.\ a certain $Q$ -- now when using a robustly trained GCN. With dotted lines, we have plotted the curves one obtains for the standard (non-robust) training -- e.g.\ the dotted lines in Fig. \ref{fig:core-robust-q-q} are the ones already seen in Fig. \ref{fig:Cora-q-Q-plot}. As it becomes clear, with robust training, we can dramatically increase the number of nodes which are robust. Almost every node is robust when considering the $Q$ for which the model has been trained for. E.g. for \textsc{Citeseer}, our method is able to quadruple the number of certifiable nodes for $Q=12$. Put simply: When performing an adversarial attack with $Q\leq 12$ on this model, it cannot do any harm!
Moreover the share of nodes that can be certified for any given $Q$ has increased significantly (even though we have not trained the model for $Q>12$). Most remarkably, nodes for which we certified non-robustness before become now certifiably robust (the blue region above the gray lines).

\textbf{\textit{Accuracy:}} The increased robustness comes at almost \textit{no loss in classification accuracy} as Table~\ref{fig:table} shows. There we report the results for all datasets and all training principles. The last two columns show the accuracy obtained for node classification (for train and test nodes separately).  In some cases, our robust classifiers even outperform the non-robust one on the unlabeled nodes. Interestingly, for \textsc{PubMed} we see that the accuracy on the labeled nodes drops to the accuracy on the unlabeled nodes. This indicates that our method can even improve generalization.

\textbf{\textit{Training principles:}} 
Comparing the different robust training procedures (also given in more detail in Figure \ref{fig:q-q-trainings}), we see that RH-U achieves significantly higher robustness when considering $Q=12$. This is shown by the third-last column in the table, where the percentage of nodes which are certifiably robust for $Q=12$ (i.e. the $Q$ the models have been robustified for) is shown.  The third column shows the largest $Q$ for which a node is still certifiably robust (averaged over all nodes). As shown, for all training principles the average exceeds the value of $12$.

\textbf{\textit{Effect of training with $Q$:}}
If we strongly increase the $Q$ for which the classifier is trained for, we only observe a small drop in the classification accuracy. E.g., training accuracy drops from 99\% to 87\% when going from $Q=12$ to 48, while test accuracy stays almost unchanged (68\% vs.\ 66\%) on \textsc{Citeseer}. We attribute this to the fact that the GNN still uses the \emph{normal} CE loss in addition to our robust hinge loss during training. Figure \ref{fig:cora_different_Q} shows the results for Cora where we trained three models with different $Q$. To clarify: We have to distinguish between the $Q$ used for training a model (mentioned in the legend) and the $Q$ we are computing certificates for (the x-axis). We see: (i) Clearly, all trainings lead to significantly more robust models. Though, the larger $Q$, the harder it gets. (ii) Importantly, each model is the `winner in robustness' when considering the $Q$ for which the model has been trained for.

\textbf{\textit{Training Dynamics:}} Lastly, we analyze the behavior when training a GCN using either standard training or robust training with \emph{RH-U}. In Figure \ref{fig:dynamics} we monitor the worst-case margin (averaged over a minibatch of nodes; separately for the labeled and unlabeled nodes) obtained in each training iteration. As seen,  with \emph{RH-U} the worst-case margin \textit{increases} to the specified values $M_1$/$M_2$ -- i.e. making them robust. In contrast, for standard training the worst-case margin \textit{decreases}. Specifically the unlabeled nodes (which account to 90\% of all nodes) are not robust.

Overall, all experiments show that our robust training is highly effective: robustness is increased while the accuracy is still high.

%% file: figures/cora_ml/rob_ceritifed_methods.pgf
\begingroup%
\makeatletter%
\begin{pgfpicture}%
\pgfpathrectangle{\pgfpointorigin}{\pgfqpoint{2.475000in}{1.529634in}}%
\pgfusepath{use as bounding box, clip}%
\begin{pgfscope}%
\pgfsetbuttcap%
\pgfsetmiterjoin%
\definecolor{currentfill}{rgb}{1.000000,1.000000,1.000000}%
\pgfsetfillcolor{currentfill}%
\pgfsetlinewidth{0.000000pt}%
\definecolor{currentstroke}{rgb}{1.000000,1.000000,1.000000}%
\pgfsetstrokecolor{currentstroke}%
\pgfsetdash{}{0pt}%
\pgfpathmoveto{\pgfqpoint{0.000000in}{0.000000in}}%
\pgfpathlineto{\pgfqpoint{2.475000in}{0.000000in}}%
\pgfpathlineto{\pgfqpoint{2.475000in}{1.529634in}}%
\pgfpathlineto{\pgfqpoint{0.000000in}{1.529634in}}%
\pgfpathclose%
\pgfusepath{fill}%
\end{pgfscope}%
\begin{pgfscope}%
\pgfsetbuttcap%
\pgfsetmiterjoin%
\definecolor{currentfill}{rgb}{1.000000,1.000000,1.000000}%
\pgfsetfillcolor{currentfill}%
\pgfsetlinewidth{0.000000pt}%
\definecolor{currentstroke}{rgb}{0.000000,0.000000,0.000000}%
\pgfsetstrokecolor{currentstroke}%
\pgfsetstrokeopacity{0.000000}%
\pgfsetdash{}{0pt}%
\pgfpathmoveto{\pgfqpoint{0.495869in}{0.430556in}}%
\pgfpathlineto{\pgfqpoint{2.322222in}{0.430556in}}%
\pgfpathlineto{\pgfqpoint{2.322222in}{1.376856in}}%
\pgfpathlineto{\pgfqpoint{0.495869in}{1.376856in}}%
\pgfpathclose%
\pgfusepath{fill}%
\end{pgfscope}%
\begin{pgfscope}%
\pgfsetbuttcap%
\pgfsetroundjoin%
\definecolor{currentfill}{rgb}{0.150000,0.150000,0.150000}%
\pgfsetfillcolor{currentfill}%
\pgfsetlinewidth{1.254687pt}%
\definecolor{currentstroke}{rgb}{0.150000,0.150000,0.150000}%
\pgfsetstrokecolor{currentstroke}%
\pgfsetdash{}{0pt}%
\pgfsys@defobject{currentmarker}{\pgfqpoint{0.000000in}{-0.083333in}}{\pgfqpoint{0.000000in}{0.000000in}}{%
\pgfpathmoveto{\pgfqpoint{0.000000in}{0.000000in}}%
\pgfpathlineto{\pgfqpoint{0.000000in}{-0.083333in}}%
\pgfusepath{stroke,fill}%
}%
\begin{pgfscope}%
\pgfsys@transformshift{0.575275in}{0.430556in}%
\pgfsys@useobject{currentmarker}{}%
\end{pgfscope}%
\end{pgfscope}%
\begin{pgfscope}%
\definecolor{textcolor}{rgb}{0.150000,0.150000,0.150000}%
\pgfsetstrokecolor{textcolor}%
\pgfsetfillcolor{textcolor}%
\pgftext[x=0.575275in,y=0.298611in,,top]{\color{textcolor}\rmfamily\fontsize{8.000000}{9.600000}\selectfont \(\displaystyle 0\)}%
\end{pgfscope}%
\begin{pgfscope}%
\pgfsetbuttcap%
\pgfsetroundjoin%
\definecolor{currentfill}{rgb}{0.150000,0.150000,0.150000}%
\pgfsetfillcolor{currentfill}%
\pgfsetlinewidth{1.254687pt}%
\definecolor{currentstroke}{rgb}{0.150000,0.150000,0.150000}%
\pgfsetstrokecolor{currentstroke}%
\pgfsetdash{}{0pt}%
\pgfsys@defobject{currentmarker}{\pgfqpoint{0.000000in}{-0.083333in}}{\pgfqpoint{0.000000in}{0.000000in}}{%
\pgfpathmoveto{\pgfqpoint{0.000000in}{0.000000in}}%
\pgfpathlineto{\pgfqpoint{0.000000in}{-0.083333in}}%
\pgfusepath{stroke,fill}%
}%
\begin{pgfscope}%
\pgfsys@transformshift{1.369342in}{0.430556in}%
\pgfsys@useobject{currentmarker}{}%
\end{pgfscope}%
\end{pgfscope}%
\begin{pgfscope}%
\definecolor{textcolor}{rgb}{0.150000,0.150000,0.150000}%
\pgfsetstrokecolor{textcolor}%
\pgfsetfillcolor{textcolor}%
\pgftext[x=1.369342in,y=0.298611in,,top]{\color{textcolor}\rmfamily\fontsize{8.000000}{9.600000}\selectfont \(\displaystyle 50\)}%
\end{pgfscope}%
\begin{pgfscope}%
\pgfsetbuttcap%
\pgfsetroundjoin%
\definecolor{currentfill}{rgb}{0.150000,0.150000,0.150000}%
\pgfsetfillcolor{currentfill}%
\pgfsetlinewidth{1.254687pt}%
\definecolor{currentstroke}{rgb}{0.150000,0.150000,0.150000}%
\pgfsetstrokecolor{currentstroke}%
\pgfsetdash{}{0pt}%
\pgfsys@defobject{currentmarker}{\pgfqpoint{0.000000in}{-0.083333in}}{\pgfqpoint{0.000000in}{0.000000in}}{%
\pgfpathmoveto{\pgfqpoint{0.000000in}{0.000000in}}%
\pgfpathlineto{\pgfqpoint{0.000000in}{-0.083333in}}%
\pgfusepath{stroke,fill}%
}%
\begin{pgfscope}%
\pgfsys@transformshift{2.163409in}{0.430556in}%
\pgfsys@useobject{currentmarker}{}%
\end{pgfscope}%
\end{pgfscope}%
\begin{pgfscope}%
\definecolor{textcolor}{rgb}{0.150000,0.150000,0.150000}%
\pgfsetstrokecolor{textcolor}%
\pgfsetfillcolor{textcolor}%
\pgftext[x=2.163409in,y=0.298611in,,top]{\color{textcolor}\rmfamily\fontsize{8.000000}{9.600000}\selectfont \(\displaystyle 100\)}%
\end{pgfscope}%
\begin{pgfscope}%
\pgfsetbuttcap%
\pgfsetroundjoin%
\definecolor{currentfill}{rgb}{0.150000,0.150000,0.150000}%
\pgfsetfillcolor{currentfill}%
\pgfsetlinewidth{1.254687pt}%
\definecolor{currentstroke}{rgb}{0.150000,0.150000,0.150000}%
\pgfsetstrokecolor{currentstroke}%
\pgfsetdash{}{0pt}%
\pgfsys@defobject{currentmarker}{\pgfqpoint{0.000000in}{-0.083333in}}{\pgfqpoint{0.000000in}{0.000000in}}{%
\pgfpathmoveto{\pgfqpoint{0.000000in}{0.000000in}}%
\pgfpathlineto{\pgfqpoint{0.000000in}{-0.083333in}}%
\pgfusepath{stroke,fill}%
}%
\begin{pgfscope}%
\pgfsys@transformshift{0.765851in}{0.430556in}%
\pgfsys@useobject{currentmarker}{}%
\end{pgfscope}%
\end{pgfscope}%
\begin{pgfscope}%
\definecolor{textcolor}{rgb}{0.150000,0.150000,0.150000}%
\pgfsetstrokecolor{textcolor}%
\pgfsetfillcolor{textcolor}%
\pgftext[x=0.765851in,y=0.298611in,,top]{\color{textcolor}\rmfamily\fontsize{8.000000}{9.600000}\selectfont \(\displaystyle 12\)}%
\end{pgfscope}%
\begin{pgfscope}%
\definecolor{textcolor}{rgb}{0.150000,0.150000,0.150000}%
\pgfsetstrokecolor{textcolor}%
\pgfsetfillcolor{textcolor}%
\pgftext[x=1.409045in,y=0.221320in,,top]{\color{textcolor}\rmfamily\fontsize{8.000000}{9.600000}\selectfont Certificate w.r.t \(\displaystyle Q\)}%
\end{pgfscope}%
\begin{pgfscope}%
\pgfsetbuttcap%
\pgfsetroundjoin%
\definecolor{currentfill}{rgb}{0.150000,0.150000,0.150000}%
\pgfsetfillcolor{currentfill}%
\pgfsetlinewidth{1.254687pt}%
\definecolor{currentstroke}{rgb}{0.150000,0.150000,0.150000}%
\pgfsetstrokecolor{currentstroke}%
\pgfsetdash{}{0pt}%
\pgfsys@defobject{currentmarker}{\pgfqpoint{-0.083333in}{0.000000in}}{\pgfqpoint{0.000000in}{0.000000in}}{%
\pgfpathmoveto{\pgfqpoint{0.000000in}{0.000000in}}%
\pgfpathlineto{\pgfqpoint{-0.083333in}{0.000000in}}%
\pgfusepath{stroke,fill}%
}%
\begin{pgfscope}%
\pgfsys@transformshift{0.495869in}{0.473569in}%
\pgfsys@useobject{currentmarker}{}%
\end{pgfscope}%
\end{pgfscope}%
\begin{pgfscope}%
\definecolor{textcolor}{rgb}{0.150000,0.150000,0.150000}%
\pgfsetstrokecolor{textcolor}%
\pgfsetfillcolor{textcolor}%
\pgftext[x=0.304896in,y=0.435307in,left,base]{\color{textcolor}\rmfamily\fontsize{8.000000}{9.600000}\selectfont \(\displaystyle 0\)}%
\end{pgfscope}%
\begin{pgfscope}%
\pgfsetbuttcap%
\pgfsetroundjoin%
\definecolor{currentfill}{rgb}{0.150000,0.150000,0.150000}%
\pgfsetfillcolor{currentfill}%
\pgfsetlinewidth{1.254687pt}%
\definecolor{currentstroke}{rgb}{0.150000,0.150000,0.150000}%
\pgfsetstrokecolor{currentstroke}%
\pgfsetdash{}{0pt}%
\pgfsys@defobject{currentmarker}{\pgfqpoint{-0.083333in}{0.000000in}}{\pgfqpoint{0.000000in}{0.000000in}}{%
\pgfpathmoveto{\pgfqpoint{0.000000in}{0.000000in}}%
\pgfpathlineto{\pgfqpoint{-0.083333in}{0.000000in}}%
\pgfusepath{stroke,fill}%
}%
\begin{pgfscope}%
\pgfsys@transformshift{0.495869in}{0.903706in}%
\pgfsys@useobject{currentmarker}{}%
\end{pgfscope}%
\end{pgfscope}%
\begin{pgfscope}%
\definecolor{textcolor}{rgb}{0.150000,0.150000,0.150000}%
\pgfsetstrokecolor{textcolor}%
\pgfsetfillcolor{textcolor}%
\pgftext[x=0.245867in,y=0.865444in,left,base]{\color{textcolor}\rmfamily\fontsize{8.000000}{9.600000}\selectfont \(\displaystyle 50\)}%
\end{pgfscope}%
\begin{pgfscope}%
\pgfsetbuttcap%
\pgfsetroundjoin%
\definecolor{currentfill}{rgb}{0.150000,0.150000,0.150000}%
\pgfsetfillcolor{currentfill}%
\pgfsetlinewidth{1.254687pt}%
\definecolor{currentstroke}{rgb}{0.150000,0.150000,0.150000}%
\pgfsetstrokecolor{currentstroke}%
\pgfsetdash{}{0pt}%
\pgfsys@defobject{currentmarker}{\pgfqpoint{-0.083333in}{0.000000in}}{\pgfqpoint{0.000000in}{0.000000in}}{%
\pgfpathmoveto{\pgfqpoint{0.000000in}{0.000000in}}%
\pgfpathlineto{\pgfqpoint{-0.083333in}{0.000000in}}%
\pgfusepath{stroke,fill}%
}%
\begin{pgfscope}%
\pgfsys@transformshift{0.495869in}{1.333843in}%
\pgfsys@useobject{currentmarker}{}%
\end{pgfscope}%
\end{pgfscope}%
\begin{pgfscope}%
\definecolor{textcolor}{rgb}{0.150000,0.150000,0.150000}%
\pgfsetstrokecolor{textcolor}%
\pgfsetfillcolor{textcolor}%
\pgftext[x=0.186838in,y=1.295580in,left,base]{\color{textcolor}\rmfamily\fontsize{8.000000}{9.600000}\selectfont \(\displaystyle 100\)}%
\end{pgfscope}%
\begin{pgfscope}%
\definecolor{textcolor}{rgb}{0.150000,0.150000,0.150000}%
\pgfsetstrokecolor{textcolor}%
\pgfsetfillcolor{textcolor}%
\pgftext[x=0.193783in,y=0.903706in,,bottom,rotate=90.000000]{\color{textcolor}\rmfamily\fontsize{8.000000}{9.600000}\selectfont \% Nodes cert. robust}%
\end{pgfscope}%
\begin{pgfscope}%
\pgfpathrectangle{\pgfqpoint{0.495869in}{0.430556in}}{\pgfqpoint{1.826353in}{0.946301in}}%
\pgfusepath{clip}%
\pgfsetroundcap%
\pgfsetroundjoin%
\pgfsetlinewidth{1.505625pt}%
\definecolor{currentstroke}{rgb}{1.000000,0.498039,0.054902}%
\pgfsetstrokecolor{currentstroke}%
\pgfsetdash{}{0pt}%
\pgfpathmoveto{\pgfqpoint{0.575275in}{1.333843in}}%
\pgfpathlineto{\pgfqpoint{0.591157in}{1.333843in}}%
\pgfpathlineto{\pgfqpoint{0.607038in}{1.328385in}}%
\pgfpathlineto{\pgfqpoint{0.622919in}{1.326374in}}%
\pgfpathlineto{\pgfqpoint{0.638801in}{1.322353in}}%
\pgfpathlineto{\pgfqpoint{0.654682in}{1.316608in}}%
\pgfpathlineto{\pgfqpoint{0.670563in}{1.312300in}}%
\pgfpathlineto{\pgfqpoint{0.686445in}{1.305693in}}%
\pgfpathlineto{\pgfqpoint{0.702326in}{1.299949in}}%
\pgfpathlineto{\pgfqpoint{0.718207in}{1.294778in}}%
\pgfpathlineto{\pgfqpoint{0.734089in}{1.283002in}}%
\pgfpathlineto{\pgfqpoint{0.749970in}{1.271512in}}%
\pgfpathlineto{\pgfqpoint{0.765851in}{1.257150in}}%
\pgfpathlineto{\pgfqpoint{0.781733in}{1.220384in}}%
\pgfpathlineto{\pgfqpoint{0.797614in}{1.172990in}}%
\pgfpathlineto{\pgfqpoint{0.813495in}{1.135362in}}%
\pgfpathlineto{\pgfqpoint{0.829377in}{1.105490in}}%
\pgfpathlineto{\pgfqpoint{0.845258in}{1.079064in}}%
\pgfpathlineto{\pgfqpoint{0.861139in}{1.056085in}}%
\pgfpathlineto{\pgfqpoint{0.877021in}{1.030234in}}%
\pgfpathlineto{\pgfqpoint{0.892902in}{1.010989in}}%
\pgfpathlineto{\pgfqpoint{0.908783in}{0.993467in}}%
\pgfpathlineto{\pgfqpoint{0.924665in}{0.974797in}}%
\pgfpathlineto{\pgfqpoint{0.940546in}{0.957276in}}%
\pgfpathlineto{\pgfqpoint{0.956427in}{0.941190in}}%
\pgfpathlineto{\pgfqpoint{0.972309in}{0.922233in}}%
\pgfpathlineto{\pgfqpoint{0.988190in}{0.906722in}}%
\pgfpathlineto{\pgfqpoint{1.004071in}{0.892073in}}%
\pgfpathlineto{\pgfqpoint{1.019953in}{0.878573in}}%
\pgfpathlineto{\pgfqpoint{1.035834in}{0.869956in}}%
\pgfpathlineto{\pgfqpoint{1.051715in}{0.857892in}}%
\pgfpathlineto{\pgfqpoint{1.067597in}{0.846689in}}%
\pgfpathlineto{\pgfqpoint{1.083478in}{0.833477in}}%
\pgfpathlineto{\pgfqpoint{1.099359in}{0.825434in}}%
\pgfpathlineto{\pgfqpoint{1.131122in}{0.807051in}}%
\pgfpathlineto{\pgfqpoint{1.147003in}{0.795561in}}%
\pgfpathlineto{\pgfqpoint{1.162885in}{0.783785in}}%
\pgfpathlineto{\pgfqpoint{1.178766in}{0.776029in}}%
\pgfpathlineto{\pgfqpoint{1.194647in}{0.765114in}}%
\pgfpathlineto{\pgfqpoint{1.210529in}{0.757646in}}%
\pgfpathlineto{\pgfqpoint{1.226410in}{0.746157in}}%
\pgfpathlineto{\pgfqpoint{1.242291in}{0.736678in}}%
\pgfpathlineto{\pgfqpoint{1.258173in}{0.726337in}}%
\pgfpathlineto{\pgfqpoint{1.274054in}{0.718869in}}%
\pgfpathlineto{\pgfqpoint{1.289935in}{0.709678in}}%
\pgfpathlineto{\pgfqpoint{1.305817in}{0.700199in}}%
\pgfpathlineto{\pgfqpoint{1.321698in}{0.691295in}}%
\pgfpathlineto{\pgfqpoint{1.337579in}{0.682677in}}%
\pgfpathlineto{\pgfqpoint{1.353461in}{0.674922in}}%
\pgfpathlineto{\pgfqpoint{1.369342in}{0.666879in}}%
\pgfpathlineto{\pgfqpoint{1.385223in}{0.660273in}}%
\pgfpathlineto{\pgfqpoint{1.401105in}{0.651369in}}%
\pgfpathlineto{\pgfqpoint{1.416986in}{0.645911in}}%
\pgfpathlineto{\pgfqpoint{1.432867in}{0.639879in}}%
\pgfpathlineto{\pgfqpoint{1.448749in}{0.634709in}}%
\pgfpathlineto{\pgfqpoint{1.480511in}{0.622071in}}%
\pgfpathlineto{\pgfqpoint{1.496393in}{0.618624in}}%
\pgfpathlineto{\pgfqpoint{1.512274in}{0.612592in}}%
\pgfpathlineto{\pgfqpoint{1.528155in}{0.608858in}}%
\pgfpathlineto{\pgfqpoint{1.544037in}{0.604836in}}%
\pgfpathlineto{\pgfqpoint{1.559918in}{0.599379in}}%
\pgfpathlineto{\pgfqpoint{1.575799in}{0.592772in}}%
\pgfpathlineto{\pgfqpoint{1.591681in}{0.588751in}}%
\pgfpathlineto{\pgfqpoint{1.607562in}{0.585304in}}%
\pgfpathlineto{\pgfqpoint{1.623443in}{0.581570in}}%
\pgfpathlineto{\pgfqpoint{1.639325in}{0.577262in}}%
\pgfpathlineto{\pgfqpoint{1.655206in}{0.574389in}}%
\pgfpathlineto{\pgfqpoint{1.671087in}{0.571230in}}%
\pgfpathlineto{\pgfqpoint{1.686969in}{0.568357in}}%
\pgfpathlineto{\pgfqpoint{1.702850in}{0.565198in}}%
\pgfpathlineto{\pgfqpoint{1.718731in}{0.562900in}}%
\pgfpathlineto{\pgfqpoint{1.734613in}{0.561176in}}%
\pgfpathlineto{\pgfqpoint{1.750494in}{0.556293in}}%
\pgfpathlineto{\pgfqpoint{1.766375in}{0.554283in}}%
\pgfpathlineto{\pgfqpoint{1.782257in}{0.551123in}}%
\pgfpathlineto{\pgfqpoint{1.798138in}{0.549687in}}%
\pgfpathlineto{\pgfqpoint{1.814019in}{0.547676in}}%
\pgfpathlineto{\pgfqpoint{1.829901in}{0.544804in}}%
\pgfpathlineto{\pgfqpoint{1.845782in}{0.541644in}}%
\pgfpathlineto{\pgfqpoint{1.861663in}{0.539634in}}%
\pgfpathlineto{\pgfqpoint{1.909307in}{0.531878in}}%
\pgfpathlineto{\pgfqpoint{1.925189in}{0.528719in}}%
\pgfpathlineto{\pgfqpoint{1.941070in}{0.527570in}}%
\pgfpathlineto{\pgfqpoint{1.956951in}{0.526134in}}%
\pgfpathlineto{\pgfqpoint{1.988714in}{0.524410in}}%
\pgfpathlineto{\pgfqpoint{2.004595in}{0.521538in}}%
\pgfpathlineto{\pgfqpoint{2.020477in}{0.521250in}}%
\pgfpathlineto{\pgfqpoint{2.036358in}{0.519240in}}%
\pgfpathlineto{\pgfqpoint{2.052239in}{0.517804in}}%
\pgfpathlineto{\pgfqpoint{2.084002in}{0.515506in}}%
\pgfpathlineto{\pgfqpoint{2.099883in}{0.514931in}}%
\pgfpathlineto{\pgfqpoint{2.115765in}{0.514644in}}%
\pgfpathlineto{\pgfqpoint{2.131646in}{0.513495in}}%
\pgfpathlineto{\pgfqpoint{2.147527in}{0.512059in}}%
\pgfpathlineto{\pgfqpoint{2.163409in}{0.511197in}}%
\pgfpathlineto{\pgfqpoint{2.179290in}{0.510048in}}%
\pgfpathlineto{\pgfqpoint{2.195171in}{0.509474in}}%
\pgfpathlineto{\pgfqpoint{2.211053in}{0.508038in}}%
\pgfpathlineto{\pgfqpoint{2.226934in}{0.506314in}}%
\pgfpathlineto{\pgfqpoint{2.258697in}{0.505165in}}%
\pgfpathlineto{\pgfqpoint{2.274578in}{0.505165in}}%
\pgfpathlineto{\pgfqpoint{2.290459in}{0.504304in}}%
\pgfpathlineto{\pgfqpoint{2.306341in}{0.503729in}}%
\pgfpathlineto{\pgfqpoint{2.323889in}{0.502837in}}%
\pgfpathlineto{\pgfqpoint{2.323889in}{0.502837in}}%
\pgfusepath{stroke}%
\end{pgfscope}%
\begin{pgfscope}%
\pgfpathrectangle{\pgfqpoint{0.495869in}{0.430556in}}{\pgfqpoint{1.826353in}{0.946301in}}%
\pgfusepath{clip}%
\pgfsetroundcap%
\pgfsetroundjoin%
\pgfsetlinewidth{1.505625pt}%
\definecolor{currentstroke}{rgb}{0.172549,0.627451,0.172549}%
\pgfsetstrokecolor{currentstroke}%
\pgfsetdash{}{0pt}%
\pgfpathmoveto{\pgfqpoint{0.575275in}{1.333843in}}%
\pgfpathlineto{\pgfqpoint{0.591157in}{1.333843in}}%
\pgfpathlineto{\pgfqpoint{0.607038in}{1.300035in}}%
\pgfpathlineto{\pgfqpoint{0.622919in}{1.280008in}}%
\pgfpathlineto{\pgfqpoint{0.638801in}{1.261489in}}%
\pgfpathlineto{\pgfqpoint{0.654682in}{1.246846in}}%
\pgfpathlineto{\pgfqpoint{0.670563in}{1.227035in}}%
\pgfpathlineto{\pgfqpoint{0.686445in}{1.209593in}}%
\pgfpathlineto{\pgfqpoint{0.702326in}{1.189351in}}%
\pgfpathlineto{\pgfqpoint{0.718207in}{1.178369in}}%
\pgfpathlineto{\pgfqpoint{0.734089in}{1.164372in}}%
\pgfpathlineto{\pgfqpoint{0.749970in}{1.149083in}}%
\pgfpathlineto{\pgfqpoint{0.765851in}{1.134655in}}%
\pgfpathlineto{\pgfqpoint{0.781733in}{1.118936in}}%
\pgfpathlineto{\pgfqpoint{0.797614in}{1.102355in}}%
\pgfpathlineto{\pgfqpoint{0.829377in}{1.067685in}}%
\pgfpathlineto{\pgfqpoint{0.861139in}{1.036892in}}%
\pgfpathlineto{\pgfqpoint{0.877021in}{1.024833in}}%
\pgfpathlineto{\pgfqpoint{0.892902in}{1.012344in}}%
\pgfpathlineto{\pgfqpoint{0.908783in}{0.997701in}}%
\pgfpathlineto{\pgfqpoint{0.924665in}{0.980904in}}%
\pgfpathlineto{\pgfqpoint{0.940546in}{0.967123in}}%
\pgfpathlineto{\pgfqpoint{0.956427in}{0.952695in}}%
\pgfpathlineto{\pgfqpoint{0.972309in}{0.937406in}}%
\pgfpathlineto{\pgfqpoint{0.988190in}{0.923194in}}%
\pgfpathlineto{\pgfqpoint{1.004071in}{0.910920in}}%
\pgfpathlineto{\pgfqpoint{1.035834in}{0.887879in}}%
\pgfpathlineto{\pgfqpoint{1.051715in}{0.877542in}}%
\pgfpathlineto{\pgfqpoint{1.067597in}{0.868714in}}%
\pgfpathlineto{\pgfqpoint{1.083478in}{0.860531in}}%
\pgfpathlineto{\pgfqpoint{1.099359in}{0.850625in}}%
\pgfpathlineto{\pgfqpoint{1.115241in}{0.843088in}}%
\pgfpathlineto{\pgfqpoint{1.131122in}{0.834906in}}%
\pgfpathlineto{\pgfqpoint{1.147003in}{0.826077in}}%
\pgfpathlineto{\pgfqpoint{1.162885in}{0.817894in}}%
\pgfpathlineto{\pgfqpoint{1.178766in}{0.808204in}}%
\pgfpathlineto{\pgfqpoint{1.194647in}{0.799806in}}%
\pgfpathlineto{\pgfqpoint{1.210529in}{0.794422in}}%
\pgfpathlineto{\pgfqpoint{1.242291in}{0.778918in}}%
\pgfpathlineto{\pgfqpoint{1.258173in}{0.771596in}}%
\pgfpathlineto{\pgfqpoint{1.274054in}{0.762983in}}%
\pgfpathlineto{\pgfqpoint{1.289935in}{0.755877in}}%
\pgfpathlineto{\pgfqpoint{1.305817in}{0.750278in}}%
\pgfpathlineto{\pgfqpoint{1.321698in}{0.743172in}}%
\pgfpathlineto{\pgfqpoint{1.337579in}{0.735635in}}%
\pgfpathlineto{\pgfqpoint{1.353461in}{0.729606in}}%
\pgfpathlineto{\pgfqpoint{1.369342in}{0.724222in}}%
\pgfpathlineto{\pgfqpoint{1.385223in}{0.716685in}}%
\pgfpathlineto{\pgfqpoint{1.416986in}{0.701181in}}%
\pgfpathlineto{\pgfqpoint{1.432867in}{0.693644in}}%
\pgfpathlineto{\pgfqpoint{1.464630in}{0.683739in}}%
\pgfpathlineto{\pgfqpoint{1.480511in}{0.675556in}}%
\pgfpathlineto{\pgfqpoint{1.496393in}{0.671249in}}%
\pgfpathlineto{\pgfqpoint{1.512274in}{0.665220in}}%
\pgfpathlineto{\pgfqpoint{1.528155in}{0.660913in}}%
\pgfpathlineto{\pgfqpoint{1.544037in}{0.654453in}}%
\pgfpathlineto{\pgfqpoint{1.559918in}{0.648854in}}%
\pgfpathlineto{\pgfqpoint{1.575799in}{0.644332in}}%
\pgfpathlineto{\pgfqpoint{1.591681in}{0.637656in}}%
\pgfpathlineto{\pgfqpoint{1.607562in}{0.633134in}}%
\pgfpathlineto{\pgfqpoint{1.623443in}{0.628397in}}%
\pgfpathlineto{\pgfqpoint{1.639325in}{0.625598in}}%
\pgfpathlineto{\pgfqpoint{1.655206in}{0.621506in}}%
\pgfpathlineto{\pgfqpoint{1.671087in}{0.616984in}}%
\pgfpathlineto{\pgfqpoint{1.686969in}{0.615046in}}%
\pgfpathlineto{\pgfqpoint{1.702850in}{0.611385in}}%
\pgfpathlineto{\pgfqpoint{1.718731in}{0.607940in}}%
\pgfpathlineto{\pgfqpoint{1.734613in}{0.603633in}}%
\pgfpathlineto{\pgfqpoint{1.750494in}{0.600834in}}%
\pgfpathlineto{\pgfqpoint{1.766375in}{0.597388in}}%
\pgfpathlineto{\pgfqpoint{1.782257in}{0.593297in}}%
\pgfpathlineto{\pgfqpoint{1.798138in}{0.590067in}}%
\pgfpathlineto{\pgfqpoint{1.829901in}{0.586621in}}%
\pgfpathlineto{\pgfqpoint{1.845782in}{0.582745in}}%
\pgfpathlineto{\pgfqpoint{1.877545in}{0.575424in}}%
\pgfpathlineto{\pgfqpoint{1.909307in}{0.570686in}}%
\pgfpathlineto{\pgfqpoint{1.925189in}{0.568748in}}%
\pgfpathlineto{\pgfqpoint{1.941070in}{0.566164in}}%
\pgfpathlineto{\pgfqpoint{1.956951in}{0.562504in}}%
\pgfpathlineto{\pgfqpoint{1.972833in}{0.559058in}}%
\pgfpathlineto{\pgfqpoint{1.988714in}{0.556259in}}%
\pgfpathlineto{\pgfqpoint{2.004595in}{0.554967in}}%
\pgfpathlineto{\pgfqpoint{2.020477in}{0.553029in}}%
\pgfpathlineto{\pgfqpoint{2.036358in}{0.551306in}}%
\pgfpathlineto{\pgfqpoint{2.068121in}{0.546999in}}%
\pgfpathlineto{\pgfqpoint{2.084002in}{0.545923in}}%
\pgfpathlineto{\pgfqpoint{2.099883in}{0.544631in}}%
\pgfpathlineto{\pgfqpoint{2.115765in}{0.543554in}}%
\pgfpathlineto{\pgfqpoint{2.131646in}{0.541831in}}%
\pgfpathlineto{\pgfqpoint{2.147527in}{0.540970in}}%
\pgfpathlineto{\pgfqpoint{2.163409in}{0.539678in}}%
\pgfpathlineto{\pgfqpoint{2.179290in}{0.537740in}}%
\pgfpathlineto{\pgfqpoint{2.195171in}{0.536232in}}%
\pgfpathlineto{\pgfqpoint{2.211053in}{0.534079in}}%
\pgfpathlineto{\pgfqpoint{2.226934in}{0.533218in}}%
\pgfpathlineto{\pgfqpoint{2.242815in}{0.532141in}}%
\pgfpathlineto{\pgfqpoint{2.258697in}{0.530418in}}%
\pgfpathlineto{\pgfqpoint{2.274578in}{0.529126in}}%
\pgfpathlineto{\pgfqpoint{2.290459in}{0.527404in}}%
\pgfpathlineto{\pgfqpoint{2.306341in}{0.526542in}}%
\pgfpathlineto{\pgfqpoint{2.323889in}{0.525999in}}%
\pgfpathlineto{\pgfqpoint{2.323889in}{0.525999in}}%
\pgfusepath{stroke}%
\end{pgfscope}%
\begin{pgfscope}%
\pgfpathrectangle{\pgfqpoint{0.495869in}{0.430556in}}{\pgfqpoint{1.826353in}{0.946301in}}%
\pgfusepath{clip}%
\pgfsetroundcap%
\pgfsetroundjoin%
\pgfsetlinewidth{1.505625pt}%
\definecolor{currentstroke}{rgb}{0.839216,0.152941,0.156863}%
\pgfsetstrokecolor{currentstroke}%
\pgfsetdash{}{0pt}%
\pgfpathmoveto{\pgfqpoint{0.575275in}{1.333843in}}%
\pgfpathlineto{\pgfqpoint{0.591157in}{1.333843in}}%
\pgfpathlineto{\pgfqpoint{0.607038in}{1.302403in}}%
\pgfpathlineto{\pgfqpoint{0.622919in}{1.278286in}}%
\pgfpathlineto{\pgfqpoint{0.638801in}{1.257613in}}%
\pgfpathlineto{\pgfqpoint{0.654682in}{1.235218in}}%
\pgfpathlineto{\pgfqpoint{0.670563in}{1.217560in}}%
\pgfpathlineto{\pgfqpoint{0.686445in}{1.197965in}}%
\pgfpathlineto{\pgfqpoint{0.702326in}{1.177723in}}%
\pgfpathlineto{\pgfqpoint{0.718207in}{1.159204in}}%
\pgfpathlineto{\pgfqpoint{0.734089in}{1.142838in}}%
\pgfpathlineto{\pgfqpoint{0.749970in}{1.127334in}}%
\pgfpathlineto{\pgfqpoint{0.765851in}{1.105154in}}%
\pgfpathlineto{\pgfqpoint{0.781733in}{1.086204in}}%
\pgfpathlineto{\pgfqpoint{0.797614in}{1.071777in}}%
\pgfpathlineto{\pgfqpoint{0.813495in}{1.051966in}}%
\pgfpathlineto{\pgfqpoint{0.829377in}{1.034954in}}%
\pgfpathlineto{\pgfqpoint{0.845258in}{1.017296in}}%
\pgfpathlineto{\pgfqpoint{0.861139in}{0.997485in}}%
\pgfpathlineto{\pgfqpoint{0.877021in}{0.979612in}}%
\pgfpathlineto{\pgfqpoint{0.908783in}{0.947096in}}%
\pgfpathlineto{\pgfqpoint{0.924665in}{0.932023in}}%
\pgfpathlineto{\pgfqpoint{0.940546in}{0.919749in}}%
\pgfpathlineto{\pgfqpoint{0.956427in}{0.906398in}}%
\pgfpathlineto{\pgfqpoint{0.972309in}{0.890032in}}%
\pgfpathlineto{\pgfqpoint{0.988190in}{0.875604in}}%
\pgfpathlineto{\pgfqpoint{1.004071in}{0.862899in}}%
\pgfpathlineto{\pgfqpoint{1.019953in}{0.848687in}}%
\pgfpathlineto{\pgfqpoint{1.035834in}{0.834044in}}%
\pgfpathlineto{\pgfqpoint{1.051715in}{0.823277in}}%
\pgfpathlineto{\pgfqpoint{1.083478in}{0.803466in}}%
\pgfpathlineto{\pgfqpoint{1.099359in}{0.795499in}}%
\pgfpathlineto{\pgfqpoint{1.115241in}{0.783871in}}%
\pgfpathlineto{\pgfqpoint{1.131122in}{0.774826in}}%
\pgfpathlineto{\pgfqpoint{1.147003in}{0.765352in}}%
\pgfpathlineto{\pgfqpoint{1.162885in}{0.756738in}}%
\pgfpathlineto{\pgfqpoint{1.178766in}{0.748555in}}%
\pgfpathlineto{\pgfqpoint{1.194647in}{0.739080in}}%
\pgfpathlineto{\pgfqpoint{1.210529in}{0.727883in}}%
\pgfpathlineto{\pgfqpoint{1.226410in}{0.720777in}}%
\pgfpathlineto{\pgfqpoint{1.242291in}{0.711948in}}%
\pgfpathlineto{\pgfqpoint{1.258173in}{0.701612in}}%
\pgfpathlineto{\pgfqpoint{1.274054in}{0.693429in}}%
\pgfpathlineto{\pgfqpoint{1.305817in}{0.677925in}}%
\pgfpathlineto{\pgfqpoint{1.321698in}{0.671249in}}%
\pgfpathlineto{\pgfqpoint{1.337579in}{0.663928in}}%
\pgfpathlineto{\pgfqpoint{1.353461in}{0.657467in}}%
\pgfpathlineto{\pgfqpoint{1.369342in}{0.650577in}}%
\pgfpathlineto{\pgfqpoint{1.385223in}{0.641317in}}%
\pgfpathlineto{\pgfqpoint{1.416986in}{0.631842in}}%
\pgfpathlineto{\pgfqpoint{1.432867in}{0.626028in}}%
\pgfpathlineto{\pgfqpoint{1.448749in}{0.618491in}}%
\pgfpathlineto{\pgfqpoint{1.464630in}{0.611601in}}%
\pgfpathlineto{\pgfqpoint{1.480511in}{0.605571in}}%
\pgfpathlineto{\pgfqpoint{1.496393in}{0.600618in}}%
\pgfpathlineto{\pgfqpoint{1.512274in}{0.595450in}}%
\pgfpathlineto{\pgfqpoint{1.528155in}{0.591574in}}%
\pgfpathlineto{\pgfqpoint{1.544037in}{0.587267in}}%
\pgfpathlineto{\pgfqpoint{1.559918in}{0.582530in}}%
\pgfpathlineto{\pgfqpoint{1.575799in}{0.579946in}}%
\pgfpathlineto{\pgfqpoint{1.591681in}{0.576716in}}%
\pgfpathlineto{\pgfqpoint{1.607562in}{0.573917in}}%
\pgfpathlineto{\pgfqpoint{1.623443in}{0.568318in}}%
\pgfpathlineto{\pgfqpoint{1.639325in}{0.563365in}}%
\pgfpathlineto{\pgfqpoint{1.655206in}{0.557982in}}%
\pgfpathlineto{\pgfqpoint{1.671087in}{0.554105in}}%
\pgfpathlineto{\pgfqpoint{1.686969in}{0.551091in}}%
\pgfpathlineto{\pgfqpoint{1.702850in}{0.548722in}}%
\pgfpathlineto{\pgfqpoint{1.718731in}{0.545492in}}%
\pgfpathlineto{\pgfqpoint{1.734613in}{0.543769in}}%
\pgfpathlineto{\pgfqpoint{1.750494in}{0.541185in}}%
\pgfpathlineto{\pgfqpoint{1.766375in}{0.539893in}}%
\pgfpathlineto{\pgfqpoint{1.782257in}{0.538170in}}%
\pgfpathlineto{\pgfqpoint{1.798138in}{0.535156in}}%
\pgfpathlineto{\pgfqpoint{1.814019in}{0.532787in}}%
\pgfpathlineto{\pgfqpoint{1.829901in}{0.531064in}}%
\pgfpathlineto{\pgfqpoint{1.861663in}{0.527188in}}%
\pgfpathlineto{\pgfqpoint{1.877545in}{0.524604in}}%
\pgfpathlineto{\pgfqpoint{1.893426in}{0.523312in}}%
\pgfpathlineto{\pgfqpoint{1.909307in}{0.521589in}}%
\pgfpathlineto{\pgfqpoint{1.925189in}{0.520082in}}%
\pgfpathlineto{\pgfqpoint{1.941070in}{0.519651in}}%
\pgfpathlineto{\pgfqpoint{1.956951in}{0.518144in}}%
\pgfpathlineto{\pgfqpoint{1.972833in}{0.516852in}}%
\pgfpathlineto{\pgfqpoint{1.988714in}{0.515775in}}%
\pgfpathlineto{\pgfqpoint{2.004595in}{0.514914in}}%
\pgfpathlineto{\pgfqpoint{2.020477in}{0.513837in}}%
\pgfpathlineto{\pgfqpoint{2.036358in}{0.513407in}}%
\pgfpathlineto{\pgfqpoint{2.052239in}{0.513191in}}%
\pgfpathlineto{\pgfqpoint{2.068121in}{0.512761in}}%
\pgfpathlineto{\pgfqpoint{2.084002in}{0.512761in}}%
\pgfpathlineto{\pgfqpoint{2.115765in}{0.511038in}}%
\pgfpathlineto{\pgfqpoint{2.131646in}{0.510607in}}%
\pgfpathlineto{\pgfqpoint{2.147527in}{0.509315in}}%
\pgfpathlineto{\pgfqpoint{2.163409in}{0.508669in}}%
\pgfpathlineto{\pgfqpoint{2.179290in}{0.506947in}}%
\pgfpathlineto{\pgfqpoint{2.195171in}{0.506301in}}%
\pgfpathlineto{\pgfqpoint{2.211053in}{0.505439in}}%
\pgfpathlineto{\pgfqpoint{2.242815in}{0.504147in}}%
\pgfpathlineto{\pgfqpoint{2.258697in}{0.504147in}}%
\pgfpathlineto{\pgfqpoint{2.274578in}{0.503932in}}%
\pgfpathlineto{\pgfqpoint{2.290459in}{0.502855in}}%
\pgfpathlineto{\pgfqpoint{2.306341in}{0.501994in}}%
\pgfpathlineto{\pgfqpoint{2.323889in}{0.500657in}}%
\pgfpathlineto{\pgfqpoint{2.323889in}{0.500657in}}%
\pgfusepath{stroke}%
\end{pgfscope}%
\begin{pgfscope}%
\pgfpathrectangle{\pgfqpoint{0.495869in}{0.430556in}}{\pgfqpoint{1.826353in}{0.946301in}}%
\pgfusepath{clip}%
\pgfsetroundcap%
\pgfsetroundjoin%
\pgfsetlinewidth{1.505625pt}%
\definecolor{currentstroke}{rgb}{0.121569,0.466667,0.705882}%
\pgfsetstrokecolor{currentstroke}%
\pgfsetdash{}{0pt}%
\pgfpathmoveto{\pgfqpoint{0.575275in}{1.333843in}}%
\pgfpathlineto{\pgfqpoint{0.591157in}{1.333843in}}%
\pgfpathlineto{\pgfqpoint{0.607038in}{1.276108in}}%
\pgfpathlineto{\pgfqpoint{0.622919in}{1.237618in}}%
\pgfpathlineto{\pgfqpoint{0.654682in}{1.168394in}}%
\pgfpathlineto{\pgfqpoint{0.670563in}{1.130479in}}%
\pgfpathlineto{\pgfqpoint{0.686445in}{1.096011in}}%
\pgfpathlineto{\pgfqpoint{0.702326in}{1.067862in}}%
\pgfpathlineto{\pgfqpoint{0.718207in}{1.033106in}}%
\pgfpathlineto{\pgfqpoint{0.734089in}{1.000361in}}%
\pgfpathlineto{\pgfqpoint{0.749970in}{0.970776in}}%
\pgfpathlineto{\pgfqpoint{0.765851in}{0.941765in}}%
\pgfpathlineto{\pgfqpoint{0.781733in}{0.913903in}}%
\pgfpathlineto{\pgfqpoint{0.797614in}{0.893509in}}%
\pgfpathlineto{\pgfqpoint{0.813495in}{0.867945in}}%
\pgfpathlineto{\pgfqpoint{0.829377in}{0.844966in}}%
\pgfpathlineto{\pgfqpoint{0.845258in}{0.821125in}}%
\pgfpathlineto{\pgfqpoint{0.861139in}{0.798434in}}%
\pgfpathlineto{\pgfqpoint{0.877021in}{0.771434in}}%
\pgfpathlineto{\pgfqpoint{0.892902in}{0.747880in}}%
\pgfpathlineto{\pgfqpoint{0.908783in}{0.732369in}}%
\pgfpathlineto{\pgfqpoint{0.924665in}{0.712550in}}%
\pgfpathlineto{\pgfqpoint{0.940546in}{0.693880in}}%
\pgfpathlineto{\pgfqpoint{0.972309in}{0.658837in}}%
\pgfpathlineto{\pgfqpoint{0.988190in}{0.643613in}}%
\pgfpathlineto{\pgfqpoint{1.004071in}{0.632124in}}%
\pgfpathlineto{\pgfqpoint{1.019953in}{0.617475in}}%
\pgfpathlineto{\pgfqpoint{1.035834in}{0.605698in}}%
\pgfpathlineto{\pgfqpoint{1.051715in}{0.594209in}}%
\pgfpathlineto{\pgfqpoint{1.067597in}{0.588177in}}%
\pgfpathlineto{\pgfqpoint{1.083478in}{0.579272in}}%
\pgfpathlineto{\pgfqpoint{1.099359in}{0.570942in}}%
\pgfpathlineto{\pgfqpoint{1.115241in}{0.558878in}}%
\pgfpathlineto{\pgfqpoint{1.131122in}{0.550549in}}%
\pgfpathlineto{\pgfqpoint{1.147003in}{0.542506in}}%
\pgfpathlineto{\pgfqpoint{1.162885in}{0.537910in}}%
\pgfpathlineto{\pgfqpoint{1.194647in}{0.527570in}}%
\pgfpathlineto{\pgfqpoint{1.210529in}{0.524410in}}%
\pgfpathlineto{\pgfqpoint{1.226410in}{0.520963in}}%
\pgfpathlineto{\pgfqpoint{1.242291in}{0.517229in}}%
\pgfpathlineto{\pgfqpoint{1.258173in}{0.515793in}}%
\pgfpathlineto{\pgfqpoint{1.274054in}{0.513208in}}%
\pgfpathlineto{\pgfqpoint{1.305817in}{0.507463in}}%
\pgfpathlineto{\pgfqpoint{1.321698in}{0.506601in}}%
\pgfpathlineto{\pgfqpoint{1.337579in}{0.504016in}}%
\pgfpathlineto{\pgfqpoint{1.353461in}{0.503155in}}%
\pgfpathlineto{\pgfqpoint{1.369342in}{0.501431in}}%
\pgfpathlineto{\pgfqpoint{1.385223in}{0.499421in}}%
\pgfpathlineto{\pgfqpoint{1.416986in}{0.497123in}}%
\pgfpathlineto{\pgfqpoint{1.432867in}{0.496261in}}%
\pgfpathlineto{\pgfqpoint{1.448749in}{0.495686in}}%
\pgfpathlineto{\pgfqpoint{1.464630in}{0.493676in}}%
\pgfpathlineto{\pgfqpoint{1.496393in}{0.490803in}}%
\pgfpathlineto{\pgfqpoint{1.512274in}{0.490229in}}%
\pgfpathlineto{\pgfqpoint{1.528155in}{0.489367in}}%
\pgfpathlineto{\pgfqpoint{1.544037in}{0.488793in}}%
\pgfpathlineto{\pgfqpoint{1.559918in}{0.487931in}}%
\pgfpathlineto{\pgfqpoint{1.607562in}{0.486208in}}%
\pgfpathlineto{\pgfqpoint{1.623443in}{0.485346in}}%
\pgfpathlineto{\pgfqpoint{1.639325in}{0.484771in}}%
\pgfpathlineto{\pgfqpoint{1.655206in}{0.484771in}}%
\pgfpathlineto{\pgfqpoint{1.671087in}{0.484484in}}%
\pgfpathlineto{\pgfqpoint{1.686969in}{0.484484in}}%
\pgfpathlineto{\pgfqpoint{1.718731in}{0.482761in}}%
\pgfpathlineto{\pgfqpoint{1.734613in}{0.482186in}}%
\pgfpathlineto{\pgfqpoint{1.750494in}{0.481899in}}%
\pgfpathlineto{\pgfqpoint{1.766375in}{0.481037in}}%
\pgfpathlineto{\pgfqpoint{1.782257in}{0.481037in}}%
\pgfpathlineto{\pgfqpoint{1.798138in}{0.480750in}}%
\pgfpathlineto{\pgfqpoint{1.814019in}{0.480176in}}%
\pgfpathlineto{\pgfqpoint{1.829901in}{0.479888in}}%
\pgfpathlineto{\pgfqpoint{1.845782in}{0.479314in}}%
\pgfpathlineto{\pgfqpoint{1.877545in}{0.479314in}}%
\pgfpathlineto{\pgfqpoint{1.893426in}{0.478739in}}%
\pgfpathlineto{\pgfqpoint{1.909307in}{0.477878in}}%
\pgfpathlineto{\pgfqpoint{1.925189in}{0.477303in}}%
\pgfpathlineto{\pgfqpoint{1.956951in}{0.477303in}}%
\pgfpathlineto{\pgfqpoint{1.972833in}{0.476442in}}%
\pgfpathlineto{\pgfqpoint{1.988714in}{0.476442in}}%
\pgfpathlineto{\pgfqpoint{2.020477in}{0.475293in}}%
\pgfpathlineto{\pgfqpoint{2.036358in}{0.475005in}}%
\pgfpathlineto{\pgfqpoint{2.052239in}{0.475005in}}%
\pgfpathlineto{\pgfqpoint{2.068121in}{0.474431in}}%
\pgfpathlineto{\pgfqpoint{2.099883in}{0.473856in}}%
\pgfpathlineto{\pgfqpoint{2.147527in}{0.473856in}}%
\pgfpathlineto{\pgfqpoint{2.163409in}{0.473569in}}%
\pgfpathlineto{\pgfqpoint{2.323889in}{0.473569in}}%
\pgfpathlineto{\pgfqpoint{2.323889in}{0.473569in}}%
\pgfusepath{stroke}%
\end{pgfscope}%
\begin{pgfscope}%
\pgfpathrectangle{\pgfqpoint{0.495869in}{0.430556in}}{\pgfqpoint{1.826353in}{0.946301in}}%
\pgfusepath{clip}%
\pgfsetbuttcap%
\pgfsetroundjoin%
\pgfsetlinewidth{1.003750pt}%
\definecolor{currentstroke}{rgb}{0.266667,0.266667,0.266667}%
\pgfsetstrokecolor{currentstroke}%
\pgfsetdash{{3.700000pt}{1.600000pt}}{0.000000pt}%
\pgfpathmoveto{\pgfqpoint{0.765851in}{0.473569in}}%
\pgfpathlineto{\pgfqpoint{0.765851in}{1.333843in}}%
\pgfusepath{stroke}%
\end{pgfscope}%
\begin{pgfscope}%
\pgfsetrectcap%
\pgfsetmiterjoin%
\pgfsetlinewidth{1.003750pt}%
\definecolor{currentstroke}{rgb}{0.700000,0.700000,0.700000}%
\pgfsetstrokecolor{currentstroke}%
\pgfsetdash{}{0pt}%
\pgfpathmoveto{\pgfqpoint{0.495869in}{0.430556in}}%
\pgfpathlineto{\pgfqpoint{0.495869in}{1.376856in}}%
\pgfusepath{stroke}%
\end{pgfscope}%
\begin{pgfscope}%
\pgfsetrectcap%
\pgfsetmiterjoin%
\pgfsetlinewidth{1.003750pt}%
\definecolor{currentstroke}{rgb}{0.700000,0.700000,0.700000}%
\pgfsetstrokecolor{currentstroke}%
\pgfsetdash{}{0pt}%
\pgfpathmoveto{\pgfqpoint{2.322222in}{0.430556in}}%
\pgfpathlineto{\pgfqpoint{2.322222in}{1.376856in}}%
\pgfusepath{stroke}%
\end{pgfscope}%
\begin{pgfscope}%
\pgfsetrectcap%
\pgfsetmiterjoin%
\pgfsetlinewidth{1.003750pt}%
\definecolor{currentstroke}{rgb}{0.700000,0.700000,0.700000}%
\pgfsetstrokecolor{currentstroke}%
\pgfsetdash{}{0pt}%
\pgfpathmoveto{\pgfqpoint{0.495869in}{0.430556in}}%
\pgfpathlineto{\pgfqpoint{2.322222in}{0.430556in}}%
\pgfusepath{stroke}%
\end{pgfscope}%
\begin{pgfscope}%
\pgfsetrectcap%
\pgfsetmiterjoin%
\pgfsetlinewidth{1.003750pt}%
\definecolor{currentstroke}{rgb}{0.700000,0.700000,0.700000}%
\pgfsetstrokecolor{currentstroke}%
\pgfsetdash{}{0pt}%
\pgfpathmoveto{\pgfqpoint{0.495869in}{1.376856in}}%
\pgfpathlineto{\pgfqpoint{2.322222in}{1.376856in}}%
\pgfusepath{stroke}%
\end{pgfscope}%
\begin{pgfscope}%
\pgfsetbuttcap%
\pgfsetmiterjoin%
\definecolor{currentfill}{rgb}{1.000000,1.000000,1.000000}%
\pgfsetfillcolor{currentfill}%
\pgfsetfillopacity{0.800000}%
\pgfsetlinewidth{1.003750pt}%
\definecolor{currentstroke}{rgb}{0.800000,0.800000,0.800000}%
\pgfsetstrokecolor{currentstroke}%
\pgfsetstrokeopacity{0.800000}%
\pgfsetdash{}{0pt}%
\pgfpathmoveto{\pgfqpoint{1.661103in}{0.756813in}}%
\pgfpathlineto{\pgfqpoint{2.254167in}{0.756813in}}%
\pgfpathquadraticcurveto{\pgfqpoint{2.273611in}{0.756813in}}{\pgfqpoint{2.273611in}{0.776257in}}%
\pgfpathlineto{\pgfqpoint{2.273611in}{1.308801in}}%
\pgfpathquadraticcurveto{\pgfqpoint{2.273611in}{1.328245in}}{\pgfqpoint{2.254167in}{1.328245in}}%
\pgfpathlineto{\pgfqpoint{1.661103in}{1.328245in}}%
\pgfpathquadraticcurveto{\pgfqpoint{1.641658in}{1.328245in}}{\pgfqpoint{1.641658in}{1.308801in}}%
\pgfpathlineto{\pgfqpoint{1.641658in}{0.776257in}}%
\pgfpathquadraticcurveto{\pgfqpoint{1.641658in}{0.756813in}}{\pgfqpoint{1.661103in}{0.756813in}}%
\pgfpathclose%
\pgfusepath{stroke,fill}%
\end{pgfscope}%
\begin{pgfscope}%
\pgfsetroundcap%
\pgfsetroundjoin%
\pgfsetlinewidth{1.505625pt}%
\definecolor{currentstroke}{rgb}{1.000000,0.498039,0.054902}%
\pgfsetstrokecolor{currentstroke}%
\pgfsetdash{}{0pt}%
\pgfpathmoveto{\pgfqpoint{1.680547in}{1.255329in}}%
\pgfpathlineto{\pgfqpoint{1.874992in}{1.255329in}}%
\pgfusepath{stroke}%
\end{pgfscope}%
\begin{pgfscope}%
\definecolor{textcolor}{rgb}{0.150000,0.150000,0.150000}%
\pgfsetstrokecolor{textcolor}%
\pgfsetfillcolor{textcolor}%
\pgftext[x=1.952769in,y=1.221301in,left,base]{\color{textcolor}\rmfamily\fontsize{7.000000}{8.400000}\selectfont RH-U}%
\end{pgfscope}%
\begin{pgfscope}%
\pgfsetroundcap%
\pgfsetroundjoin%
\pgfsetlinewidth{1.505625pt}%
\definecolor{currentstroke}{rgb}{0.172549,0.627451,0.172549}%
\pgfsetstrokecolor{currentstroke}%
\pgfsetdash{}{0pt}%
\pgfpathmoveto{\pgfqpoint{1.680547in}{1.119762in}}%
\pgfpathlineto{\pgfqpoint{1.874992in}{1.119762in}}%
\pgfusepath{stroke}%
\end{pgfscope}%
\begin{pgfscope}%
\definecolor{textcolor}{rgb}{0.150000,0.150000,0.150000}%
\pgfsetstrokecolor{textcolor}%
\pgfsetfillcolor{textcolor}%
\pgftext[x=1.952769in,y=1.085734in,left,base]{\color{textcolor}\rmfamily\fontsize{7.000000}{8.400000}\selectfont RCE}%
\end{pgfscope}%
\begin{pgfscope}%
\pgfsetroundcap%
\pgfsetroundjoin%
\pgfsetlinewidth{1.505625pt}%
\definecolor{currentstroke}{rgb}{0.839216,0.152941,0.156863}%
\pgfsetstrokecolor{currentstroke}%
\pgfsetdash{}{0pt}%
\pgfpathmoveto{\pgfqpoint{1.680547in}{0.984196in}}%
\pgfpathlineto{\pgfqpoint{1.874992in}{0.984196in}}%
\pgfusepath{stroke}%
\end{pgfscope}%
\begin{pgfscope}%
\definecolor{textcolor}{rgb}{0.150000,0.150000,0.150000}%
\pgfsetstrokecolor{textcolor}%
\pgfsetfillcolor{textcolor}%
\pgftext[x=1.952769in,y=0.950168in,left,base]{\color{textcolor}\rmfamily\fontsize{7.000000}{8.400000}\selectfont RH}%
\end{pgfscope}%
\begin{pgfscope}%
\pgfsetroundcap%
\pgfsetroundjoin%
\pgfsetlinewidth{1.505625pt}%
\definecolor{currentstroke}{rgb}{0.121569,0.466667,0.705882}%
\pgfsetstrokecolor{currentstroke}%
\pgfsetdash{}{0pt}%
\pgfpathmoveto{\pgfqpoint{1.680547in}{0.848629in}}%
\pgfpathlineto{\pgfqpoint{1.874992in}{0.848629in}}%
\pgfusepath{stroke}%
\end{pgfscope}%
\begin{pgfscope}%
\definecolor{textcolor}{rgb}{0.150000,0.150000,0.150000}%
\pgfsetstrokecolor{textcolor}%
\pgfsetfillcolor{textcolor}%
\pgftext[x=1.952769in,y=0.814602in,left,base]{\color{textcolor}\rmfamily\fontsize{7.000000}{8.400000}\selectfont CE}%
\end{pgfscope}%
\end{pgfpicture}%
\makeatother%
\endgroup%

%% file: sections/conclusion.tex
\section{Conclusion}
We proposed the first work on certifying robustness of GNNs, considering perturbations of the node attributes under a challenging $L_0$ perturbation budget and tackling the discrete data domain.
By relaxing the GNN and considering the dual, we realized an efficient computation of our certificates -- simultaneously our experiments have shown that our certificates are tight since for most nodes a certificate can be given.
We have shown that traditional training of GNNs leads to non-robust models that can easily be fooled. In contrast, using our novel (semi-supervised) robust training the resulting GNNs are shown to be much more robust. All this is achieved with only a minor effect on the classification accuracy. As future work we aim to consider perturbations of the graph structure.

%% file: sections/appendix-primal.tex
\section*{Acknowledgements}
This research was supported by the German Research Foundation, grant GU 1409/2-1.
\section{Appendix}
\textbf{Implementation Details:}
We perform the robust training using stochastic gradient descent with mini-batches and Adam Optimizer. For this we randomly sample in each iteration 20 nodes from the labeled nodes (for RH-U from all nodes) and compute the nodes' twohop neighbors. We then slice the adjacency and attribute matrices appropriately and compute the lower/upper activation bounds for all nodes in the batch. We use dropout of 0.5, $L_2$ regularization with strength $1e-5$, learning rate of $0.001$. We use Tensorflow 1.12 and train on NVIDIA GTX 1080 Ti.

\subsection{Proofs}
	We reformulate the problem in Eq.~\eqref{eq:linear_prog} as the linear program below.
\begin{equation}
	\underset{\Xadv, \hidden^{(\cdot)},\hiddenhat^{(\cdot)},\epshat}{\mathrm{minimize}}  \cvec^{\top} \hiddenhat^{(\numlayers)} \text{subject to} 
\end{equation}
\begin{align*}
	&\hiddenhat^{(\iterlayer+1)} = \Aslice^{(\iterlayer)} \hidden^{(\iterlayer)} \W^{(\iterlayer)} + \bvec^{(\iterlayer)}, \iterlayer=2,\ldots,\numlayers &&\Rightarrow \V^{(\iterlayer+1)} \in \mathbb{R}^{\mathcal{N}_{\numlayers-\iterlayer} \times \numhidden^{(\iterlayer)}} \\
	&\hidden^{(1)}_{nj} \leq 1 &&\Rightarrow \epsplus \in \mathbb{R}^{\mathcal{N}_{\numlayers-1} \times \numhidden^{(1)}} \\
	&\hidden^{(1)}_{nj} \geq 0 &&\Rightarrow \epsminus \in \mathbb{R}^{\mathcal{N}_{\numlayers-1} \times \numhidden^{(1)}} \\
	&\hidden^{(1)}_{nj} \leq \Xslice_{nj} + \epshat_{nj} &&\Rightarrow \gamplus \in \mathbb{R}^{\mathcal{N}_{\numlayers-1} \times \numhidden^{(1)}} \\
	&\hidden^{(1)}_{nj} \geq \Xslice_{nj} - \epshat_{nj} &&\Rightarrow \gamminus \in \mathbb{R}^{\mathcal{N}_{\numlayers-1} \times \numhidden^{(1)}} \\
	&\sum_{j} \epshat_{nj} \leq q \quad \forall n \in \mathcal{N}_{\numlayers-1}  &&\Rightarrow \Mlocal \in \mathbb{R}^{\mathcal{N}_{\numlayers-1}} \\
	&\sum_{n,j} \epshat_{nj} \leq Q  &&\Rightarrow \Mglobal \in \mathbb{R} \\
	&\hidden^{(\iterlayer)}_{nj} = 0, \iterlayer=2,\ldots,\numlayers-1, (n,j) \in \mathcal{I}^{(\iterlayer)}_{-}  \\
	&\hidden^{(\iterlayer)}_{nj} = \hiddenhat^{(\iterlayer)}_{nj}, \iterlayer=2,\ldots,\numlayers-1, (n,j) \in \mathcal{I}^{(\iterlayer)}_{+}  \\
	&\hidden^{(\iterlayer)}_{nj} \geq 0,  \iterlayer=2,\ldots,\numlayers-1, (n,j) \in \mathcal{I}^{(\iterlayer)} &&\Rightarrow \tauvar^{(\iterlayer)} \in \mathbb{R}^{\mathcal{N}_{\numlayers-\iterlayer} \times \numhidden^{(\iterlayer)}} \\
	&\hidden^{(\iterlayer)}_{nj} \geq \hiddenhat^{(\iterlayer)},  \iterlayer=2,\ldots,\numlayers-1, (n,j) \in \mathcal{I}^{(\iterlayer)} &&\Rightarrow \muvar^{(\iterlayer)} \in \mathbb{R}^{\mathcal{N}_{\numlayers-\iterlayer} \times \numhidden^{(\iterlayer)}} \\
	&\hidden_{nj}^{(\iterlayer)} \left(\upperb_{nj}^{(\iterlayer)} - \lowerb^{(\iterlayer)}_{nj} \right) \leq \upperb^{(\iterlayer)}_{nj} \left( \hiddenhat^{(\iterlayer)}_{nj} - \lowerb_{nj}^{(\iterlayer)} \right) , &&\Rightarrow \lambdavar^{(\iterlayer)} \in \mathbb{R}^{\mathcal{N}_{\numlayers-\iterlayer} \times \numhidden^{(\iterlayer)}} \\
	&\qquad \qquad \iterlayer=2,\ldots,\numlayers-1, (n,j) \in \mathcal{I}^{(\iterlayer)} 
\end{align*}
Note that $\Xadv=\hidden^{(1)}$; moreover the $\hidden^{(\iterlayer)}_{nj} = 0$ and $\hidden^{(\iterlayer)}_{nj} = \hiddenhat^{(\iterlayer)}$ can be simply eliminated from the optimization. $\lambdavar$, $\muvar$, and $\tauvar$ are only defined for $(n,j)\! \in\! \mathcal{I}^{(\iterlayer)}$; we keep the matrix notation for simplicity.

\begin{proof}[Proof of Theorem \ref{thm:dual_problem}]
Applying standard duality construction, the (non-simplified!) dual problem of the above linear program is 
\begin{align*}
	\max &\sum_{\iterlayer=2}^{\numlayers-1} \sum_{(n,j)\in \mathcal{I}^{(\iterlayer)}} \lambdavar^{(\iterlayer)}_{nj}\upperb^{(\iterlayer)}_{nj} \lowerb^{(\iterlayer)}_{nj}  
	- \sum_{\iterlayer=1}^{\numlayers-1}\bs{1}^\top \V^{(\iterlayer+1)}\bvec^{(\iterlayer)}   
	  \\
	&+\sum_{n,j}\Xslice_{nj}\left[\gamminus_{nj} - \gamplus_{nj} \right] - \epsplus  - q \sum_{n}\Mlocal_n - Q\Mglobal \quad \text{ subject to }
\end{align*}
\begin{align}
	\V^{(L)} = -\cvec \qquad  \V^{(\iterlayer)}_{nj} &= 0 && \text{ for }\iterlayer=2,\ldots \numlayers, (n,j) \in \mathcal{I}^{(\iterlayer)}_-  \notag\\
	\V^{(\iterlayer)} &= \Aslice^{(\iterlayer)\top} \V^{(\iterlayer+1)}\W^{(\iterlayer)\top}, && \text{ for }\iterlayer=2,\ldots \numlayers, (n,j) \in \mathcal{I}^{(\iterlayer)}_+  \notag \\
	\V_{nj}^{(\iterlayer)} &= \lambdavar_{nj}\upperb_{nj}^{(\iterlayer)} - \muvar_{nj}^{(\iterlayer)} && \text{ for }\iterlayer=2,\ldots \numlayers, (n,j) \in \mathcal{I}^{(\iterlayer)}  \notag \\
	\Aslice^{(\iterlayer)\top} \V^{(\iterlayer+1)}\W^{(\iterlayer)\top} &= \lambdavar^{(\iterlayer)} \odot \left[\upperb^{(\iterlayer)} - \lowerb^{(\iterlayer)}\right] \notag \\ 
	&- \tauvar^{(\iterlayer)} - \muvar^{(\iterlayer)} && \text{ for }\iterlayer=2,\ldots \numlayers, (n,j) \in \mathcal{I}^{(\iterlayer)} \notag \\
	\Aslice^{(1)\top}\V^{(2)}\W^{(1)\top} &= \epsplus - \epsminus + \gamplus - \gamminus \label{eq:gamma} \\
	\Mglobal + \Mlocal_{n} &\geq \gamplus_{nj} + \gamminus_{nj} \quad \forall n,j \label{eq:Mlocal_global}\\
	\lambdavar, \tauvar, \muvar, &\epsplus, \epsminus, \gamplus, \gamminus, \Mlocal, \Mglobal \geq 0 \notag
\end{align}
As done in \cite{kolterpolytope} we can exploit complementarity of the $\relu$ constraints corresponding to 
 $\hidden^{(\iterlayer)} \geq 0$ and $\hidden^{(\iterlayer)} \geq \hiddenhat^{(\iterlayer)}$ to eliminate $\tauvar$, $\muvar$, and $\lambdavar$  from the problem. For this we write
\begin{align*}
	\left[ \upperb^{(\iterlayer)} - \lowerb^{(\iterlayer)} \right] \odot \lambdavar^{(l)} &= \left[ \Aslice^{(\iterlayer)\top} \V^{(\iterlayer+1)}\W^{(\iterlayer)\top} \right]_+ = \left[ \Vhat^{(\iterlayer)} \right]_+\\
	\tauvar^{(\iterlayer)} + \muvar^{(\iterlayer)} &= \left[ \Aslice^{(\iterlayer)\top} \V^{(\iterlayer+1)}\W^{(\iterlayer)\top} \right]_- = \left[ \Vhat^{(\iterlayer)} \right]_-
\end{align*}
where we have defined $\Vhat^{(\iterlayer)}:=\Aslice^{(\iterlayer)\top} \V^{(\iterlayer+1)}\W^{(\iterlayer)\top}$. Given the non-negativity of the dual-variables, it becomes apparent that  $\tauvar^{(\iterlayer)}$ and $\muvar^{(\iterlayer)}$ ``share'' the negative part of $\Vhat^{(\iterlayer)}$. Thus, we define new variables $\alphavec^{(\iterlayer)}_{nj} \in [0,1]$ such that $\muvar^{(\iterlayer)}_{nj}=\alphavec^{(\iterlayer)}_{nj}\left[\Vhat^{(\iterlayer)}_{nj}\right]_-$. Combining this with the constraint $\V_{nj}^{(\iterlayer)} = \lambdavar_{nj}\upperb_{nj}^{(\iterlayer)} - \muvar_{nj}^{(\iterlayer)}$ we can rephrase to get
 \vspace*{-2mm}
\begin{align*} 
	\V^{(\iterlayer)}_{nj}&=\frac{\upperb_{nj}^{(\iterlayer)}}{\upperb_{nj}^{(\iterlayer)} - \lowerb_{nj}^{(\iterlayer)}}\left[  \Vhat^{(\iterlayer)}_{nj} \right]_{+}-\alphavec^{(\iterlayer)}_{nj}\left[ \Vhat^{(\iterlayer)}_{nj} \right]_{-} \\
	\lambdavar^{(\iterlayer)}_{nd} &= {\left[\Vhat^{(\iterlayer)}_{nd}\right]_+}\cdot {(\upperb_{nj}^{(\iterlayer)} - \lowerb_{nj}^{(\iterlayer)})}^{-1}
\end{align*}

Similarly, by complementarity of the constraints, we know that only one of $\epsplus_{nj}$ and $\epsminus_{nj}$ and only one of $\gamplus_{nj}$ and $\gamminus_{nj}$ can be positive. From Eq. \eqref{eq:gamma} we can therefore see that $\epsplus_{nj}$ and $\gamplus_{nj}$ need to ``share'' the positive part of the left hand side in Eq.~\eqref{eq:gamma} (since all variables are $\geq0$); similarly $\epsminus_{nj}$ and $\gamminus_{nj}$ share the negative part. We denote this (unknown) share by a new variable $\beta_{nj} \in [0,1]$ and get
\begin{align*}
	 \epsplus_{nj} &= \beta_{nj} \left [\Vhat^{(1)}_{nj} \right ]_{+}, &&\gamplus_{nj} = (1-\beta_{nj}) \left [\Vhat^{(1)}_{nj} \right ]_{+}\\
	 \epsminus_{nj} &= \beta_{nj} \left [\Vhat^{(1)}_{nj} \right ]_{-}, &&\gamminus_{nj} = (1-\beta_{nj}) \left [\Vhat^{(1)}_{nj} \right ]_{-}
\end{align*}
 Putting this into Eq.~\eqref{eq:Mlocal_global} we can now see that
\begin{align*}
	\Mlocal_n + \Mglobal &\geq (1-\beta_{nj})|\Vhat^{(1)}_{nj}| \quad \forall 1\leq n \leq \mathcal{N}_{\numlayers-1}, 1\leq j \leq \numhidden^{(1)},
\end{align*}
from which we can get 
\begin{align}
	\beta_{nj} \geq 1 - \frac{\Mglobal + \Mlocal_n}{|\Vhat^{(1)}_{nj}|}, \beta_{nj} \geq 0 \quad \forall n,j \label{eq:beta}
\end{align}
to replace the constraint in Eq.~\eqref{eq:Mlocal_global}. Now we can simplify the following term from the dual objective
\begin{align*}
	\Xslice_{nj}\left[\gamminus_{nj} - \gamplus_{nj} \right] - \epsplus_{nj} &= -\Vhat^{(1)}_{nj} \Xslice_{nj} + \beta_{nj} \Vhat^{(1)}_{nj} \Xslice_{nj} - \beta_{nj} \left[\Vhat^{(1)}_{nj} \right]_{+} \\
	&= -\Vhat^{(\iterlayer)}_{nj} \Xslice_{nj} - \bs{\Delta}_{nj}\beta_{nj} \text{ where } \\
	\bs{\Delta}_{nj} &:= \left[ \Vhat^{(1)}_{nj} \right]_+ \cdot(1-\Xslice_{nd}) + \left[ \Vhat^{(1)}_{nj} \right]_- \cdot \Xslice_{nd}.
\end{align*}
In the definition of $\bs{\Delta}_{nj}$ we essentially have a case distinction: if $\Vhat^{(1)}_{nj}$ is positive, we know that increasing the value of the corresponding primal variable $\Xadv_{nd}$ will improve the primal objective. If $\Xslice_{nd}$ is already 1, however, we set the value of $\bs{\Delta}_{nj}$ to zero by multiplying by $1-\Xslice_{nd}$ (similarly for the case when $\Vhat^{(1)}_{nj}$ is negative). This effectively enforces the $0 \leq \Xadv \leq 1$ constraint on the perturbations.

Plugging all terms defined above into our dual objective we get
\begin{align*}
\max &\sum_{\iterlayer=2}^{\numlayers-1} \sum_{(n,j) \in \mathcal{I}^{(\iterlayer)}} \frac{\upperb_{nj}^{(\iterlayer)} \lowerb_{nj}^{(\iterlayer)}}{\upperb_{nj}^{(\iterlayer)} - \lowerb_{nj}^{(\iterlayer)}} \left[\Vhat_{nj}^{(\iterlayer+1)} \right]_+  -\sum_{\iterlayer=1}^{\numlayers-1} \bs{1}^\top \V^{{(\iterlayer+1)}} \bvec^{(\iterlayer)}  \\ &  - \Tr \left[ \Xslice^\top \Vhat^{(1)} \right ] - \|\bs{\Delta} \odot \beta\|_1      - q\cdot \sum_{n}\Mlocal_n - Q\cdot \Mglobal
\end{align*}
Notice that $\Tr \left[ \Xslice^\top \Vhat^{(1)} \right ]=\sum_{n}\sum_j \Vhat_{nj} \Xslice_{nj}$. Since $\bs{\Delta} \geq 0$ and $\beta\geq 0$ we could safely write $\|\bs{\Delta} \odot \beta\|_1 $. By observing that $\bs{\Delta} \geq 0$ for all entries we see that to maximize the objective, we will set $\beta$ to a value as small as is admissible. This means we can replace Eq. \eqref{eq:beta} with $\beta = \max \left \{1- \frac{\Mglobal + \Mlocal_n}{|\Vhat^{(1)}_{nj}|}, 0 \right \}$. Thus, we have now eliminated all dual variables except $\alphavec$, $\Mglobal$, and $\Mlocal_n$. Finally, we define $\betavec_{nj}:= \bs{\Delta}_{nj} \beta_{nj} = \max\left\{ \bs{\Delta}_{nj} - (\Mlocal_n + \Mglobal),0 \right\}$, which finalizes the proof.
\end{proof}

%% file: sections/appendix2.tex
	\begin{proof}[Proof of Theorem \ref{thm:M_opt}]
	Given a fixed $\alphavec$, the dual function $g^t_{q,Q}$ reduces to
	$ - \|  \betavec\|_1 
	- q\cdot \sum_{n}\Mlocal_n - Q\cdot \Mglobal + const$ with the term
	$\betavec_{nd} =  \max\left\{\bs \Delta_{nd}  - (\Mlocal_n + \Mglobal), 0 \right\}$ and $\bs \Delta_{nd}$ constant.
	Noticing that $\betavec_{nd}\geq 0$, we see that maximizing the dual is equivalent to minimizing
	\begin{align*}
		\min_{\Mglobal,\Mlocal_n \geq 0} h(\Mglobal,\Mlocal_n)&=\sum_{n,d} \max\left\{\bs \Delta_{nd}  - (\Mlocal_n + \Mglobal), 0 \right\}
	+ q\cdot \sum_{n}\Mlocal_n + Q\cdot \Mglobal 
	\end{align*}

Observe that we can equivalently rephrase this as 
\begin{align*}
	\min_{\Mglobal, \Mlocal_n, \bs U_{nd} \geq 0} &h'(\Mglobal,\Mlocal_n,\bs U) =\sum_{n,d} \bs U_{nd} + q\cdot \sum_{n}\Mlocal_n + Q\cdot \Mglobal \\
& s.t.\; \bs U_{nd} \geq \bs \Delta_{nd} - \Mlocal_n - \Mglobal
\end{align*}
Here we have replaced $\max\{\bs \Delta_{nd} - (\Mlocal_n + \Mglobal),0\}$ in $h(\Mglobal,\Mlocal_n)$ with a new variable $\bs U_{nd}$ with the constraints $\bs U_{nd}\geq 0$ and $\bs U_{nd} \geq \bs \Delta_{nd} - \Mlocal_n - \Mglobal$ (for each $1 \leq n \leq \mathcal{N}_{\numlayers-1}, 1 \leq d \leq \numhidden^{(1)}$). Since we are minimizing, the optimal values w.r.t.\ $h'(\Mglobal,\Mlocal_n,\bs U)$ and $h(\Mglobal,\Mlocal_n)$ will be the same.

Finding the minimum of $h'(\cdot)$ is a linear program. Thus, we can again form its dual (denoting the dual variables as $\alpha_{nd}$):
\begin{align*}
	&\max_{\alpha_{nd} \geq 0} g'(\alpha_{nd}) = \sum_{n,d}\bs \Delta_{nd} \alpha_{nd} \\
	& s.t.\; \alpha_{nd} \leq 1, \qquad \sum_{n,d}\alpha_{nd} \leq Q, \qquad \sum_{d}\alpha_{nd} \leq q \; \forall n
\end{align*}
An optimal solution of this dual can be seen and computed easily. Since all $\bs \Delta_{nd}$ are nonnegative, we simply set those $\alpha_{nd}$ to 1 corresponding to the largest values of $\bs \Delta_{nd}$ -- additionally taking the two other constraints into account: The third constraint tells us that the row sums of the $\alpha$ matrix can be at most $q$, hence we can only set the $\alpha_{nd}$ corresponding to the $q$ largest $\bs \Delta_{nd}$ to 1 for each row to maximize the objective. The second constraint means that we can set at most $Q$ entries $\alpha_{nd}$ to 1. So among the set of all $q$ largest $\bs \Delta_{nd}$ of the rows we select again the $Q$ largest $\bs \Delta_{nd}$ and set their corresponding $\alpha_{nd}$ to 1\footnote{W.l.o.g. we assume $Q\leq  |\mathcal{N}_{\numlayers-1}| \cdot q$ here; otherwise we simply select all of the $q$ largest $\bs \Delta_{nd}$ per row (which is equivalent to choosing $Q=  |\mathcal{N}_{\numlayers-1}| \cdot q$)}. Observe that this is precisely the selection process described in the main text for Thm.~\ref{thm:M_opt}. That is, an optimal solution of the dual can be found by setting $\alpha_{nd}=1 \Leftrightarrow (n,d)\in S_Q$.

We now prove that the variables $\Mglobal$ and $\Mlocal_{n}$ as described in the main text (along with $\bs U_{nd}=\max \{\bs \Delta_{nd} - \Mlocal_{n} - \Mglobal,0\}$) correspond to an \emph{optimal} solution of their respective problem. For this we show that the Karush-Kuhn-Tucker (KKT) conditions hold -- using the above constructed solution for $\alpha_{nd}$. (1) Dual and primal feasibility hold by construction. (2) Next, we check  complementary slackness 
\begin{equation}
\alpha_{nd} (\bs \Delta_{nd} - \Mlocal_n - \Mglobal - \bs U_{nd}) = 0 \label{eq:complementary}
\end{equation} 
 If $\alpha_{nd} > 0$ it must hold that the second term is 0. In this case we know that $\bs \Delta_{nd} \geq \Mlocal_{n}+\Mglobal$ and thus $\bs U_{nd} = \Delta_{nd} - \Mlocal_{n}-\Mglobal$, which means the term is always 0. When the second term in Eq.~\eqref{eq:complementary} is nonzero, $\alpha_{nd}$ must be 0. This is given since when $\bs \Delta_{nd} < \Mlocal_n+\Mglobal+\bs U_{nd}$ it is smaller than the smallest $\bs \Delta_{nd}$ for which $\alpha_{nd}$ is set to 1 and therefore $\alpha_{nd}=0$. (3) Finally we show that $\nabla_{\theta}L(\theta, \alpha)=\nabla_{\theta} \sum_{n,d} \bs U_{nd} + q\cdot \sum_{n}\Mlocal_n + Q\cdot \Mglobal + \sum_{n,d} \alpha_{nd}(\bs \Delta_{nd} - \Mlocal_n - \Mglobal - \bs U_{nd}) = 0$ for $\theta = \{\bs U_{nd}, \Mlocal_n, \Mglobal\}$. Consider first $\Mglobal$: $\nabla_{\Mglobal}L(\theta, \alpha)=Q - \sum_{n,d}\alpha_{nd} = 0$ since we set exactly $Q$ many $\alpha_{nd}$ to 1 (and the rest to 0); $\Mlocal_{n}$ follows analogously. $\nabla_{\bs U_{nd}} L(\theta, \alpha)= \mathbb{I}_{\bs U_{nd} > 0}(1 - \alpha_{nd})=0$ holds since when $\alpha_{nd}=0$, $\bs \Delta_{nd} - \Mlocal_n - \Mglobal \leq 0$ which means that $\bs U_{nd}$ must be 0 because of its constraints. Thus, all KKT conditions hold. 
\end{proof}

\begin{proof}[Proof of Corollary \ref{corr:integral}]
	Assume we are given the optimal values for $\alphavec$. We can then compute the optimal values of $\Mglobal$ and $\Mlocal$ as described in Theorem~\ref{thm:M_opt}. Recall from the proof of Thm.~\ref{thm:dual_problem} that the $\bs \Delta_{nd}$ denote the improvement in the primal function objective when changing the attribute $\Xslice_{nd}$. As shown in the proof of Thm.~\ref{thm:M_opt} with the optimal $\alpha_{nd}$ we exactly choose the values $\bs \Delta_{nd}$ that lead to the largest improvement of the objective function -- and we trivially observed $\alpha_{nd} =1 $ for those elements. Thus, an optimal solution can be obtained by perturbing the attribute entries $\Xslice_{nd}$ from the set $P:=\{(n,d)| \alpha_{nd} =1, \bs \Delta_{nd}>0 \}$, i.e. setting them to $1-\Xslice_{nd}$. Thus, by construction we found an optimal solution which is integral, making the original linear program integral w.r.t. the attributes $\Xadv_{nd}$. By construction of $ \alpha_{nd}$ the set $P=\{(n,d)\in S_Q\mid \bs \Delta_{nd} > 0\}$.
\end{proof}

\begin{proof}[Proof of Corollary \ref{cor:bounds}]
Using Eq. \eqref{eq:Hslice}, the (un-perturbed)
$	\hiddenhat^{(2)}_{mj}$ is  $\Aslice^{(1)}_{m:} \Xslice\W^{(1)}_{:j}  + \bvec^{(1)}_j=\sum_n \sum_d \Aslice^{(1)}_{mn} \Xslice_{nd}\W^{(1)}_{dj}  + \bvec^{(1)}_j
$ which is simply a linear sum in $\Xslice_{nd}$. Clearly, for maximizing $\hiddenhat^{(2)}_{mj}$, one should only perturb $\Xslice_{nd}$ if ($\Aslice^{(1)}_{mn}\W^{(1)}_{dj}$ is positive and $\Xslice_{nd}=0$) or ($\Aslice^{(1)}_{mn}\W^{(1)}_{dj}$ is negative and $\Xslice_{nd}=1$). Thus, the maximal \textit{increase} of $\hiddenhat^{(2)}_{mj}$ based on $\Xslice_{nd}$ one can achieve is $\Aslice^{(1)}_{mn}\cdot\W^{(1)}_{dj}$ if the first condition holds, $-\Aslice^{(1)}_{mn}\cdot\W^{(1)}_{dj}$ if the second holds, and $0$ else. This can compactly be written as $\Aslice^{(1)}_{mn}\cdot ([\W^{(1)}_{dj}]_+\cdot (1-\Xslice_{nd})+[\W^{(1)}_{dj}]_-\cdot\Xslice_{nd})$ , which matches the terms in Eq.~\eqref{eq:upperbound}. To obtain the maximal \textit{overall} increase  in $	\hiddenhat^{(2)}_{mj}$, and, thus, an upper bound, one simply picks the largest elements that still adhere to the budget constraints (Q,q). Obviously, since this  is an admissible perturbation, this upper bound is tight. The proof for the lower bound is accordingly.
\end{proof}